%% file: main.tex
\pdfoutput=1

\documentclass[11pt]{article}
\usepackage[dvipsnames]{xcolor}

\usepackage[]{EMNLP2023}

\usepackage{times}
\usepackage{latexsym}

\usepackage[T1]{fontenc}

\usepackage[utf8]{inputenc}

\usepackage{microtype}

\usepackage{inconsolata}

\usepackage{amsmath}
\usepackage{bm}
\usepackage{graphicx}
\usepackage{dsfont}
\usepackage{amsmath,amsfonts,bm,mathtools}
\usepackage[inline]{enumitem}
\usepackage{booktabs}
\usepackage{array,multirow}
\usepackage{float}
\usepackage{comment}
\usepackage{tabularx}
\usepackage{cleveref}
\usepackage{dsfont}
\usepackage{caption}
\usepackage{subcaption}
\usepackage{url}
\usepackage[normalem]{ulem}

\definecolor{tan}{rgb}{0.69, 0.4, 0.0}
\definecolor{green}{rgb}{0.0, 0.5, 0.0}
\definecolor{bronze}{rgb}{0.8, 0.5, 0.2}

\definecolor{darkcerulean}{rgb}{0.03, 0.27, 0.49}
\definecolor{carolinablue}{rgb}{0.6, 0.73, 0.89}

\definecolor{darkgreen}{rgb}{0.12, 0.3, 0.17}
\newcommand{\green}[1]{\textcolor{darkcerulean}{#1}}
\newcommand{\yellow}[1]{\textcolor{bronze}{#1}} 
 
\newcommand{\bluelight}[1]{\textcolor{carolinablue}{#1}} 

\newcolumntype{H}{>{\setbox0=\hbox\bgroup}c<{\egroup}@{}}

\title{Attribution and Alignment:
Effects of Local Context Repetition on Utterance Production and Comprehension in Dialogue}

\newcommand{\amsterdam}[0]{$^{\triangleleft}$}
\newcommand{\aberdeen}[0]{$^{\diamond}$}

\author{Aron Molnar\aberdeen \ \ Jaap Jumelet\amsterdam \ \ Mario Giulianelli\amsterdam \ \ Arabella Sinclair\aberdeen\\
\aberdeen Department of Computing Science, University of Aberdeen\\ \amsterdam Institute for Logic, Language and Computation, University of Amsterdam\\
\url{a.molnar.19@abdn.ac.uk} \ \ \url{j.w.d.jumelet@uva.nl} \\ \url{m.giulianelli@uva.nl}\ \ \url{arabella.sinclair@abdn.ac.uk}
}

\begin{document}
\maketitle

\newcommand{\dgpt}{${\textsc{dgpt}}$}
\newcommand{\gpt}{${\textsc{gpt2}}$}
\newcommand{\opt}{${\textsc{opt}}$}

\input{sections/0_abstract}
\input{sections/introduction}
\input{sections/related}
\input{sections/experimental_setup}

\input{sections/results_repetition}
\input{sections/results_attribution}
\input{sections/discussion_conclusion}

\section*{Limitations}
Limitations of our work are that it is only conducted on English-spoken corpora, for two kinds types of dialogue context (conversational given a range of popular topics, and navigational task-oriented) and of that, native speakers of English only. Repetition patterns of dialogues in different conversational contexts, with language users of different cultures and in different languages may vary, and the patterns that models learn for these may also vary.

\section*{Acknowledgements}
We would like to thank the anonymous reviewers for their thoughtful and useful reviews and comments. We also wish to thank Ehud Reiter for his useful comments on this work at an early stage. MG is supported by the European Research Council (ERC) under the European Union's Horizon 2020 research and innovation programme (grant agreement No.\ 819455).

\bibliography{anthology,attribalign,exprepbib}
\bibliographystyle{acl_natbib}

\appendix

\input{sections/appendix_details}

\input{sections/appendix_models}

\end{document}

%% file: sections/0_abstract.tex
\begin{abstract}

Language models are often used as the backbone of modern dialogue systems. These models are pre-trained on large amounts of written \textit{fluent} language. 
Repetition is typically penalised when evaluating language model generations. However, it is a key component of dialogue.
Humans use \textit{local} and \textit{partner specific} repetitions; these are preferred by human users and lead to more successful communication in dialogue. 
In this study, we evaluate (a) whether language models produce human-like levels of repetition in dialogue, and (b) what are the processing mechanisms related to lexical re-use they use during comprehension.
We believe that such joint analysis of model production and comprehension behaviour can inform the development of cognitively inspired dialogue generation systems.

\end{abstract}

%% file: sections/introduction.tex
\section{Introduction}
\label{sec:introduction}

Human production in dialogue is influenced by many factors within the recent conversational history, leading speakers to repeat recently used lexical and structural elements of their own and their partners' language.
These factors can involve conceptual pacts speakers make in order to establish common ground~\citep{brennan1996conceptual}, priming of lexical or syntactic cues which influences their subsequent re-use~\citep{bock1986syntactic}, and other social, interpersonal, cognitive, or neural influences~\citep{pickering2005establishing,Danescu-Niculescu-Mizil+al:12a,hasson2012brain,fusaroli2014dialog}.\looseness-1 

Language models, which are often used as the backbone of modern dialogue systems, should learn to attend to such factors in order to successfully mimic human linguistic behaviour in interaction. 
The pre-training data of these models typically contains \textit{fluent} monologic language 
and little diverse dialogue data---and indeed one goal of building language generators is having them produce fluent language. A key aspect of achieving fluency is the avoidance of repetition: repetitions are typically 
thought of as evidence of degenerate production~\cite{li2016simple,li2016deep,welleck2019neural,Holtzman2019TheDegeneration}.

Recent advances in conversational language models, such as \textit{ChatGPT}, demonstrate neural models' impressive performance in producing human-like, proficient language. 
However, despite these advances, they are yet to display human-like 
communicative behaviour 
(i.e., adhering to Gricean maxims---the verbosity of such models can be high), and more nuanced, local, and partner-specific interactions. 
Humans in dialogue use specific communication strategies which rely on repetition, and, in particular, these are \textit{local} and \textit{partner-specific}~\cite{schlangen2004causes,pickering2005establishing,sinclair2023alignment}. We start from the desideratum that dialogue response generation models should also produce \textit{human-like} levels of repetition. 
While excessive levels of repetition, designed to mimic alignment, can hinder naturalness~\citep{isard2006individuality,foster2009evaluating},  humans generally prefer generated dialogue that contains higher levels of alignment~\citep{lopes2015rule,hu2016entrainment}, 
which also lead to more successful communication in human-human dialogue~\cite{xi2021taming,isard2006individuality}. 
Moreover, elements of alignment have been successfully incorporated in chat bots~\citep{hoegen2019end,gao2019structuring}.

Investigating and understanding the mechanisms which drive more human-like patterns of repetition is critical to creating more human-like natural language generation and dialogue systems. 
We therefore study whether models reproduce the repetition behaviour humans display in spoken dialogue, and the extent to which this repetition is  
affected by contextual cues.
In particular, we focus on locality effects, comparing repetition patterns of speakers with respect to their own, and their partner's language. 
We investigate language models' \textit{production} behaviour, via measuring the extent to which they 
generate similar local repetitions to humans, and their
\textit{comprehension} behaviour, through measuring the salience they assign to a given portion of the local dialogue context when comprehending an utterance.\looseness-1

%% file: sections/related.tex
\section{Background}
\label{sec:background}

\subsection{Human Repetition and Alignment}
Local repetition of shared language between speakers is one of many lower-level linguistic signals indicating the presence of interactive alignment between speakers~\cite{pickering_garrod_2004}. It is thought to contribute to more successful communication~\citep{pickering2005establishing} as it allows speakers to 
establish and maintain shared common ground~\citep{brennan1996conceptual,pickering2004toward}. 
Developing local routines---shared sequences of repeated language~\citep{pickering2005establishing,garrod2007alignment}---can also indicate 
mutual understanding between speakers~\citep{Gibbs1992,Gallotti2017}.
Producing repeated language in dialogue, either at a word level, or, in the case of routines, a construction level, is influenced by many factors in the local context. 
Speakers can be \textit{primed} by language they have been recently exposed to, which may, in addition to the coordination and alignment factors mentioned above, play a role in the choice to repeat language
\textit{locally}~\citep{tooley2010syntactic}. 
Priming effects can take place at multiple levels (from phonetic, lexical and syntactic to gesture, gaze and body posture), and are well attested in human dialogue~\citep{brennan1996conceptual,pardo2006phonetic,reitter-etal-2006-computational,holler2011co,rasenberg2020alignment}.\looseness-1 

Alignment and coordination between speakers in dialogue are often measured in terms of \textit{local} linguistic `alignment effects', i.e., whether adjacent utterances contain high linguistic overlap, and whether the incidence of repetitions decays with the distance between utterances~\cite{Reitter2006ComputationalDialogue,Xu2015AnMeasures,sinclair2018does,sinclair2021construction,giulianelli-etal-2022-construction}. 
Local shared construction use has been linked to more successful grounded communication
\citep{fusaroli2014dialog,reitter-moore-2007-predicting,REITTER201429,ward2007dialog,friedberg2012lexical,sinclair2021linguistic,norman2022studying}.
Local alignment is also affected by whether a speaker repeats their own or their partner's language, both in humans and in 
human-agent dialogue settings~\cite{Reitter2006ComputationalDialogue,sinclair2018does,duplessis2017automatic,sinclair2019tutorbot}.
We focus our attention on these short term, local repetition effects and structure our analyses accordingly.\looseness-1

\subsection{Understanding the Behaviour of Language Models}

Analysing model \textit{behaviour} is a key approach when investigating patterns of model repetition, for example, paradigms from psycholinguistics can be repurposed to this end~\cite[e.g.,][]{futrell-etal-2019-neural}.
During language comprehension, language models have been shown to be prone to structural priming effects, in a manner with parallels to findings in humans. In particular, recency of prime to target within the input context heavily influences the likelihood of the congruent structure~\cite{sinclair2022structural}. 
It is less clear, however, to what extent models are affected by priming and repetition during language \textit{production}, or generation, and what the mechanisms are that drive their \textit{comprehension} behaviour.
One method for \textit{explaining} model behaviour is to employ interpretability techniques such as attribution methods. 
Attribution methods \citep{DBLP:journals/jmlr/CovertLL21} allow for a high-level explanation of model behaviour that aligns strongly with how humans explain their decision-making, i.e., based on counterfactual examples \citep{yin-neubig-2022-interpreting}: \textit{how would the prediction have changed if a particular input feature was not present?} 
Attribution methods have been used to examine \textit{linguistic} patterns in model behaviour, and it has been argued they provide more comprehensive insights than attention heatmaps \citep{bastings-filippova-2020-elephant}, because attention only determines feature importance within a particular attention head, and not for model predictions as a whole \citep{DBLP:conf/naacl/JainW19}.
Linguistic phenomena investigated using attribution methods include co-reference, negation, and syntactic structure~\cite{DBLP:conf/conll/JumeletZH19,wu-etal-2021-polyjuice,nayak-timmapathini-2021-using, DBLP:journals/corr/abs-2306-12181}.
Within conversational NLP, feature attribution methods have been used to identify salient features in task-oriented dialogue modelling \citep{huang-etal-2020-generalizable}, dialogue response generation \citep{DBLP:conf/nips/TuanPCGW21}, and turn-taking prediction \citep{ekstedt2020turngpt}.
However, relatively little work involves these techniques used to analyse human alignment behaviour in dialogue, in terms of patterns 
of local repetition, which we make our focus.

%% file: sections/experimental_setup.tex
\section{Experimental Setup}

In this study, we investigate (a) to what extent repetition patterns in dialogue can be explained in terms of the re-use of lexical material in the local context; (b) whether LMs learn to generate repetitions with properties similar to those observed in human interaction 
and (c) how this relates to generation quality, as well as (d) whether LMs are influenced by the presence of repetitions in the local context when comprehending dialogue utterances. This section introduces the dialogue data and the language models used to study these four questions.\footnote{
    \href{https://github.com/the-context-lab/attribalign}{https://github.com/the-context-lab/attribalign}
}

\subsection{Corpora}
\label{sec:corpora}
We choose two high-quality, naturalistic dialogue corpora, transcribed from spoken human interactions, with different conversational dynamics and well attested local repetition patterns at a lexical and structural level \citep{reitter-etal-2006-computational,sinclair2021construction}. Although larger scale conversational corpora exist, often these
consist of more artificial interactions (e.g., very short or  highly closed-domain). 

\paragraph{\textit{Map Task}.} 
The Map Task corpus \cite{anderson1991hcrc} 
comprises 128 dialogues between speakers participating in a navigational task. Speakers have either an instruction giver or instruction-follower role: they either describe a route, or attempt to follow and mark the described route, on their map.\looseness-1

\paragraph{\textit{Switchboard}.}
The Switchboard corpus~\cite{godfrey1992} contains 1,155 dialogues between participants making conversation over the telephone about one of a pre-specified range of common conversational topics. Speakers in this setting have equal status, with no pre-defined roles. 

\paragraph{Extracting sample contexts.}
We are interested in evaluating the extent to which repetition occurs at a \textit{local} level, therefore we extract sample contexts of 10 utterances, using a sliding window approach. Of these, utterances 1-9 are the \textit{context}, and utterance 10 is the \textit{target} utterance which we investigate. Since we are interested in between- vs.\ within-speaker effects, we define utterances based on speech turns---i.e. each time a speaker changes, we consider this a new utterance. 
Details of the corpora and extracted samples are in Table~\ref{tab:datatset_stats}.

\begin{table}[ht]
    \centering 
    \small 
    \scalebox{0.9} { 
    \begin{tabular}{l r r} \toprule
         & \textbf{Switchboard} & \textbf{Map Task} \\ \midrule
        Full dialogues & 1,155 & 128 \\
        Number of utterances & 86.64$\pm$39.1 & 207.62$\pm$103.2 \\ 
        Unique vocabulary & 19,927 & 1,882 \\ 
        \midrule
        Samples \textit{(of 10 utterances)} & 8,705 & 2,395 \\ 
        Words per utterance & 14.6 $\pm$ 18.95 & 8.39 $\pm$ 9.21 \\ \bottomrule
        
    \end{tabular}
    }
\caption{Corpus statistics.} 
\label{tab:datatset_stats} 
\end{table}
\looseness-1

\subsection{Language Models}
\label{sec_lms}

We select three autoregressive neural language models for our analysis: DialoGPT~\cite[\dgpt;][]{zhang-etal-2020-dialogpt}, \gpt~\citep[][]{Radford2019LanguageMA}, and \opt~\cite{zhang2022opt}. We select \dgpt~as a model specifically designed for dialogue (yet still trained on written language, which differs significantly from our transcribed spoken language); \gpt~as its estimates are shown to be predictive of comprehension behaviour, even more so than larger LM variants \citep{shain2022large,oh2023does}; and \opt, which has demonstrated competitive performance across a range of benchmarks~\citep{paperno-etal-2016-lambada,Open-LLM-Leaderboard-Report-2023}. 
We fine-tune for 20 epochs, using an early stopping technique to save the best performing model based on perplexity.\footnote{More details of model sizes can be found in Appendix~\ref{app:model_sizes}.}

%% file: sections/results_repetition.tex
\section{Producing Repetitions}
\label{sec:results_repetition}

We expect human repetition patterns to be highly local, given prior results showing priming effects in the same corpora~\cite[e.g.,][]{reitter-moore-2007-predicting,sinclair2018does,sinclair2021construction}. 
We also expect repetition patterns to be modulated by which dialogue partner is being repeated. In particular, we expect between-speaker repetition patterns to be the strongest given that developing shared routines can signal alignment and coordination of speakers' mental models or interpersonal synergy~\citep{pickering2005establishing,pickering_garrod_2004,fusaroli2014dialog}. 
We firstly analyse locality and between- vs.\ within-speaker repetition in human-produced utterances, then investigate whether the same patterns occur in model generations.\looseness-1

\subsection{Methods}
\subsubsection{Measures of Repetition}

\newcommand{\vo}{${\it VO}$}
\newcommand{\co}{${\it CO}$}
\newcommand{\s}{$S$}
\newcommand{\pmi}{${\it PMI}$}

\newcommand{\sw}{Switchboard}
\newcommand{\mt}{Map Task}

To differentiate between routines vs.\ shared language, we compute two main measures of lexical repetition, at the word level, and in terms of shared word sequences (\textit{constructions}; see \Cref{sec:construction-extraction}), with which we hope to capture between-speaker routines. We measure repetition between utterance pairs, at varying distances from one another within a given context sample. 
We define additional measures to capture established human dialogue behaviours.\looseness-1

\paragraph{\textit{Vocabulary Overlap}.}
To compute vocabulary overlap, \vo, we exclude  punctuation, and calculate \vo\ as the proportion of words $w$ in the current turn $t_c$ that also appear in a previous turn $t_p$:

\begin{equation}
\small
    VO = \frac{
                | w_{t_c} \cap w_{t_p} | 
            }{
                | w_{t_c} |  
            }
    \label{eq:vo}
\end{equation}

\paragraph{\textit{Construction Repetition}.}

After extracting a shared inventory of constructions (\Cref{sec:construction-extraction}) for a dialogue, we measure the proportion of repetition of shared constructions $C$ as construction overlap \co\ as:
\begin{equation}
\small
    CO = \frac{
                | C_{t_c} \cap C_{t_p} |
            }{
                | w_{t_c} |
            }
    \label{eq:co}
\end{equation}

\paragraph{\textit{Between vs.\ Within-Speaker Repetition}.} This binary measure describes whether the producer of utterance $t_c$ and $t_p$ is the same (\textit{within}) or different (\textit{between}).

\paragraph{\textit{Locality}.} We measure locality as the distance in utterance index between $t_c$ and $t_p$. We take repetition decay, a negative effect of distance $d$ on the shared constructions between $t_c$ and $t_p$, as evidence of a local repetition effect.

\paragraph{\textit{Specificity}.} We calculate how sample-specific the extracted constructions are, and for each $t_c$, report average specificity of the repeated constructions. We measure specificity using pointwise mutual information (PMI), computed as follows:\looseness-1
\begin{equation}
\small
    PMI(c,s) = \log_2 \frac{P(c|s)}{P(c)}
    \label{eq:pmi}
\end{equation}
Higher PMI indicates a construction $c$ is more strongly associated with, or specific to, the sample $s$ it occurs within due to the frequency of occurrence in this context being higher relative to its general usage.

\subsubsection{Construction Extraction Procedure}
\label{sec:construction-extraction}
To extract repeated constructions we make use of \textit{dialign}, a framework for sequential pattern mining \cite{Guillaume2017}.\footnote{\href{https://github.com/GuillaumeDD/dialign}{https://github.com/GuillaumeDD/dialign}} 
We then discard repeated expressions with fewer than two alphanumeric tokens \cite[following][]{sinclair2021construction}. 
Repeated expressions consisting solely of punctuation or of more than half filled pauses are also excluded.
We further discard constructions which contain \textit{periods, commas and question marks}, to avoid constructions which include sentence boundaries: these do not contain the lexical elements we are interested in.
We define the resulting shared lexicon as \textit{constructions}. 
Table~\ref{tab:construction_statistics} provides details of their properties.
\footnote{
    Appendix~\ref{app:qual_examples} contains examples of constructions and how they are repeated, Appendix~\ref{app:filled_pauses} filled pauses.
}

\begin{table}[ht]
\small
\scalebox{0.85}{
    \setlength{\tabcolsep}{1.9pt}
    \begin{tabular}{@{}lcrrcrr@{}}
    \toprule
     & \multicolumn{3}{c}{\textbf{Switchboard}} & \multicolumn{3}{c}{\textbf{MapTask}} \\
            & {M}$\pm$Std & Med. & Max & M$\pm$Std & Med. & Max\\ \midrule
    \textit{Construction} & & & & & &\\
    \hspace{3 mm}Length    & 2.1 $\pm$ 0.4 & 2.0 & 5  & 2.4 $\pm$ 0.8 & 2.0 & 11   \\[2pt]      
    \hspace{3 mm}Frequency      & 3.0 $\pm$ 1.2 & 3.0 & 6  & 3.3 $\pm$ 1.1 & 3.0 & 6 \\[2pt]    
    \hspace{3 mm}Rep. Dist.   & 3.6 $\pm$ 2.7 & 3.0 & 8  & 3.3 $\pm$ 2.7 & 3.0 & 8 \\[2pt]      
    \hspace{3 mm}Incidence    & 1.6 $\pm$ 1.1 & 1.0 & 10  & 2.0 $\pm$ 1.1 & 2.0 & 8 \\[2pt]    
    \hspace{3 mm}PMI        & 6.8 $\pm$ 3.4 & 6.6 & 11.5 & 7.2 $\pm$ 2.2 & 7.6 & 9.6 \\[2pt]
    \textit{Utterance} & & & \\
    \hspace{3 mm} CO    & 0.004 $\pm$ 0.035 & 0.0 & 1.0   & 0.024 $\pm$ 0.13 & 0.0 & 2.8  \\[2pt]
    \hspace{3 mm} VO    & 0.13 $\pm$ 0.23 & 0.008 & 1.0  & 0.13 $\pm$ 0.24 & 0.0 & 1.0 \\[2pt]\bottomrule
    \end{tabular}
}
\caption{Construction properties. Repetition distance (\textit{Rep. Dist.}) measured in utterances.
}
\label{tab:construction_statistics}
\end{table}

\subsubsection{Generating Dialogue Utterances}
%

For each sample in our dataset of extracted dialogue excerpts, we precede each of the 9 utterances in the context with its speaker label, and append a final speaker label, corresponding to the upcoming target speaker, to the end. We then generate the target utterance using ancestral sampling \citep{bishop2006,koller2009probabilistic} to study an unbiased representation of the model's predictive distribution.
We set the maximum generation length to 64 tokens,
and take the presence of a newline to indicate the end of an utterance,  discarding any further generated text beyond this.\footnote{
        While the average token length for both datasets is relatively low, some utterances can be much longer. We analysed the distribution and select 64 as the maximum length since 95\% and 99\% of utterances fall below this length in Switchboard and in Map Task, respectively. 
 } The resulting text we refer to as the target. 
 To ensure that we take into account that a given context could support multiple targets---production variability is known to be high in dialogue \cite[see, e.g.,][]{giulianelli-etal-2023-what}---and to ensure our results are robust, we generate 5 utterances per context sample. 

\paragraph{Evaluating generation quality.}
\label{sec:generation_quality_stats}
\label{sec:generation_quality_metrics}

We measure the quality of a generated target utterance compared to the human reference in terms of their $n$-gram overlap \citep[BLEU;][]{Papineni2002Bleu:Translation} and semantic similarity \citep[BERTScore;][]{Zhang2019BERTScore:BERT}. We also evaluate generations using perplexity, as computed using independent models, both independently of ($PPL_{ii}$), and conditioned on the context ($PPL_{id}$); we choose GPT-2 for the same reasons highlighted in Section~\ref{sec_lms}, and Pythia (pythia-1.4b)~\citep{biderman2023pythia} for its open-source, highly performant properties. We additionally make use of MAUVE~\citep{Pillutla2021MAUVE:Frontiers} to capture higher-level distributional differences between human- vs.\ model-produced text.


\subsection{Analysis}
\subsubsection{Human vs.\ Model Repetitions} 
To analyse local production behaviour, we evaluate the extent to which human and model-produced utterances' \co\ is sensitive to 
between-speaker repetition, locality, and context-specificity.

\paragraph{\textit{The speaker being repeated affects \co\ and \vo\ in humans and models}.}

Dialogue partners differ in terms of what they repeat of their own vs.~their partner's language~\citep{reitter-etal-2006-computational,sinclair2018does}, thus we expect to find differences in our human data. We also expect that if speakers make use of local routines~\citep{pickering2005establishing}, then between-speaker \co~will be relatively higher.  
We observe that humans do indeed repeat constructions shared with their dialogue partner more so than they do those not shared (\co: \mt: $t=12.78$, $p<0.05$. \sw:  $t=17.74$, $p<0.05$ ). We observe the inverse effect for \vo, showing speakers repeat their own language relatively more so than they do their dialogue partner (\vo. \mt: $t=-13.64$, $p<0.05$. \sw:  $t=-26.66$, $p<0.05$). 
While models exhibit global human-like \co~and \vo~patterns to some degree, for example \gpt\ tuned is no different to human \co~ for within-speaker in \sw~($t=-0.18$, $p=0.86$), and between-speaker in \mt~($t=-1.86$, $p=0.06$), these effects are not consistent across models or corpora. 
Figure~\ref{fig:repstats} illustrates these results, details of statistical differences in Appendix~\ref{app:repetitionstats}. 

\begin{figure}[ht]
    \centering 
    \includegraphics[height=2.4cm]{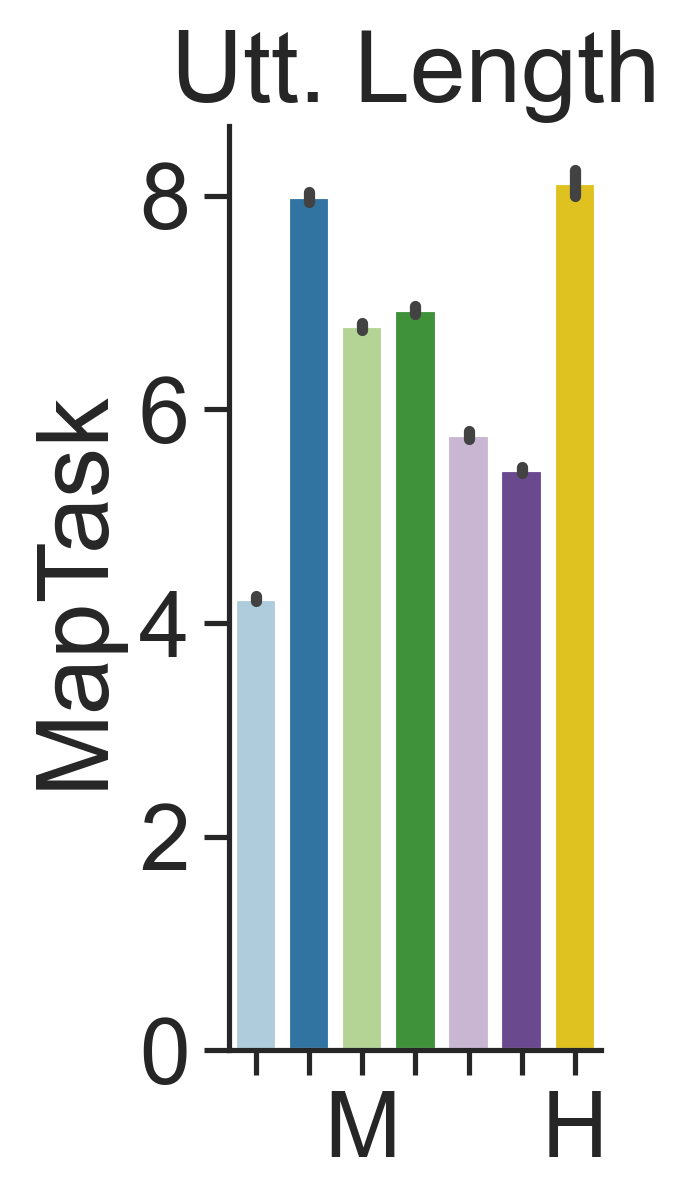}
    \includegraphics[height=2.4cm]{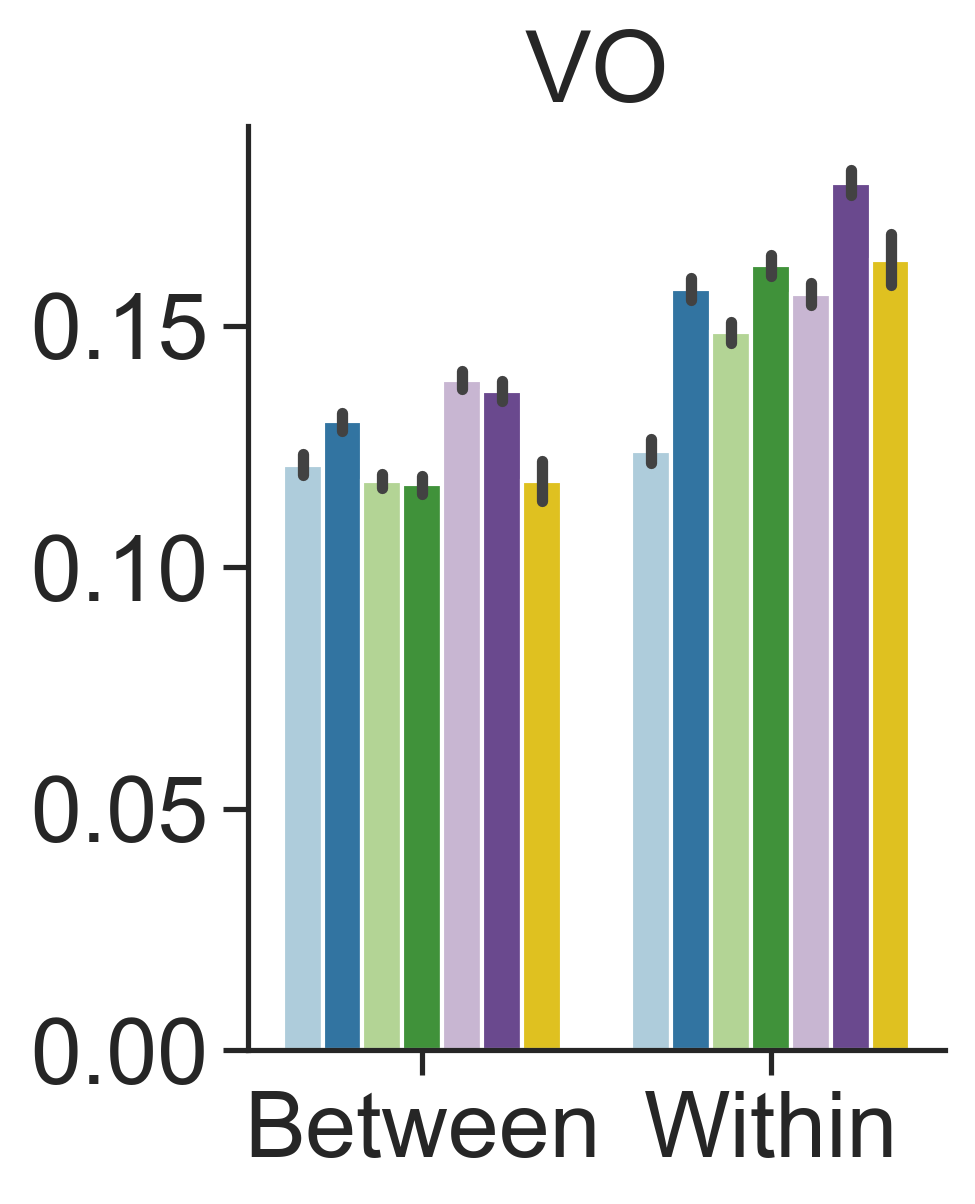}
    \includegraphics[height=2.4cm]{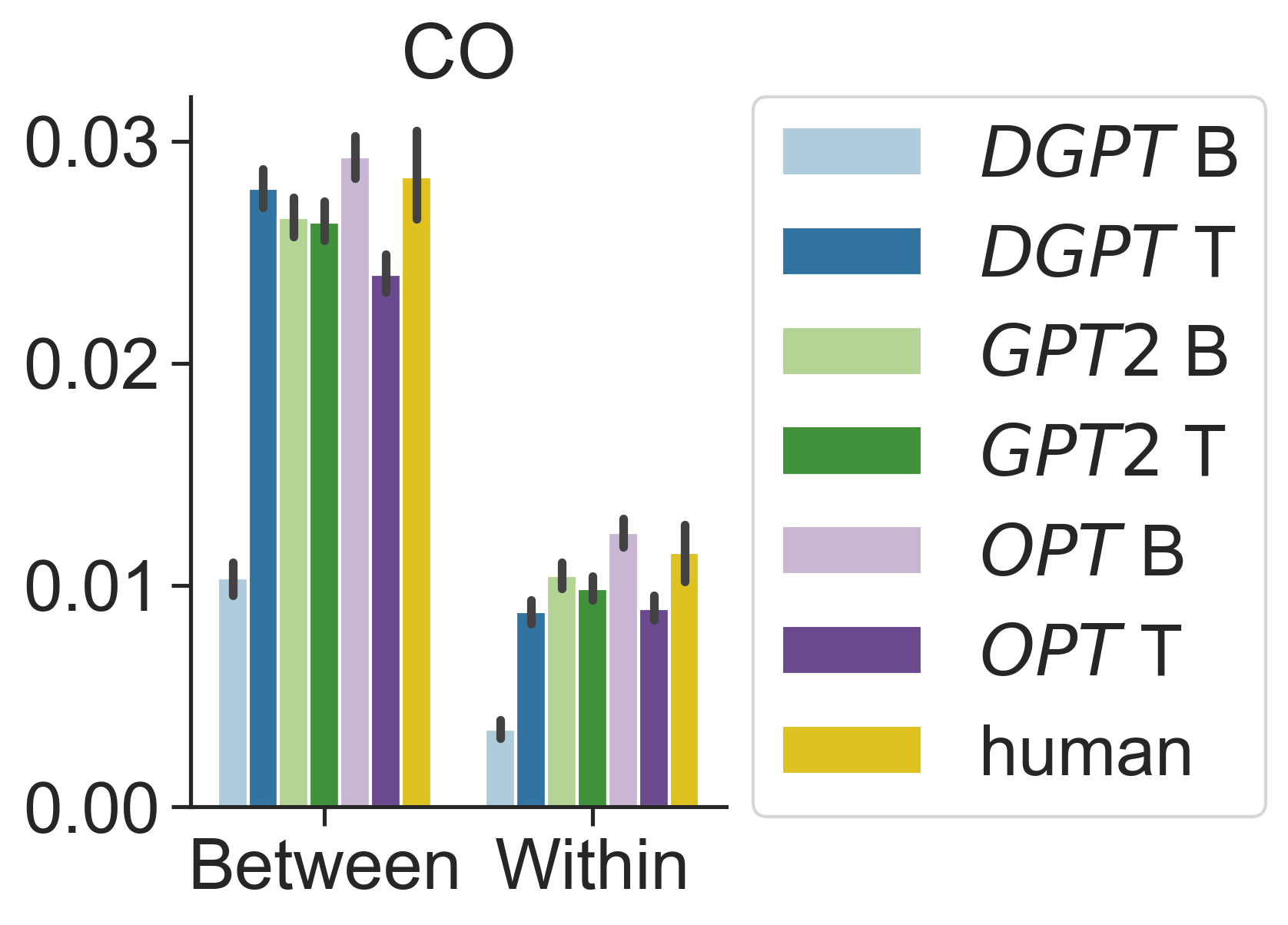}
    
    \includegraphics[height=2.4cm]{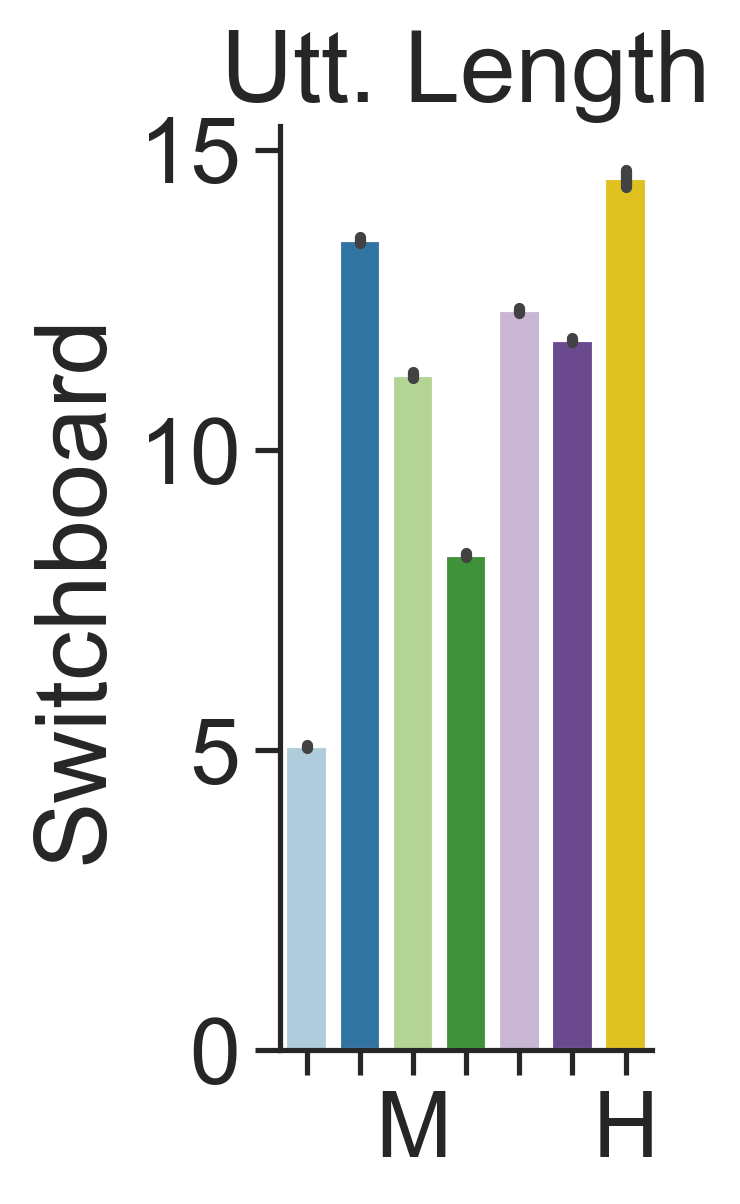}
    \includegraphics[height=2.4cm]{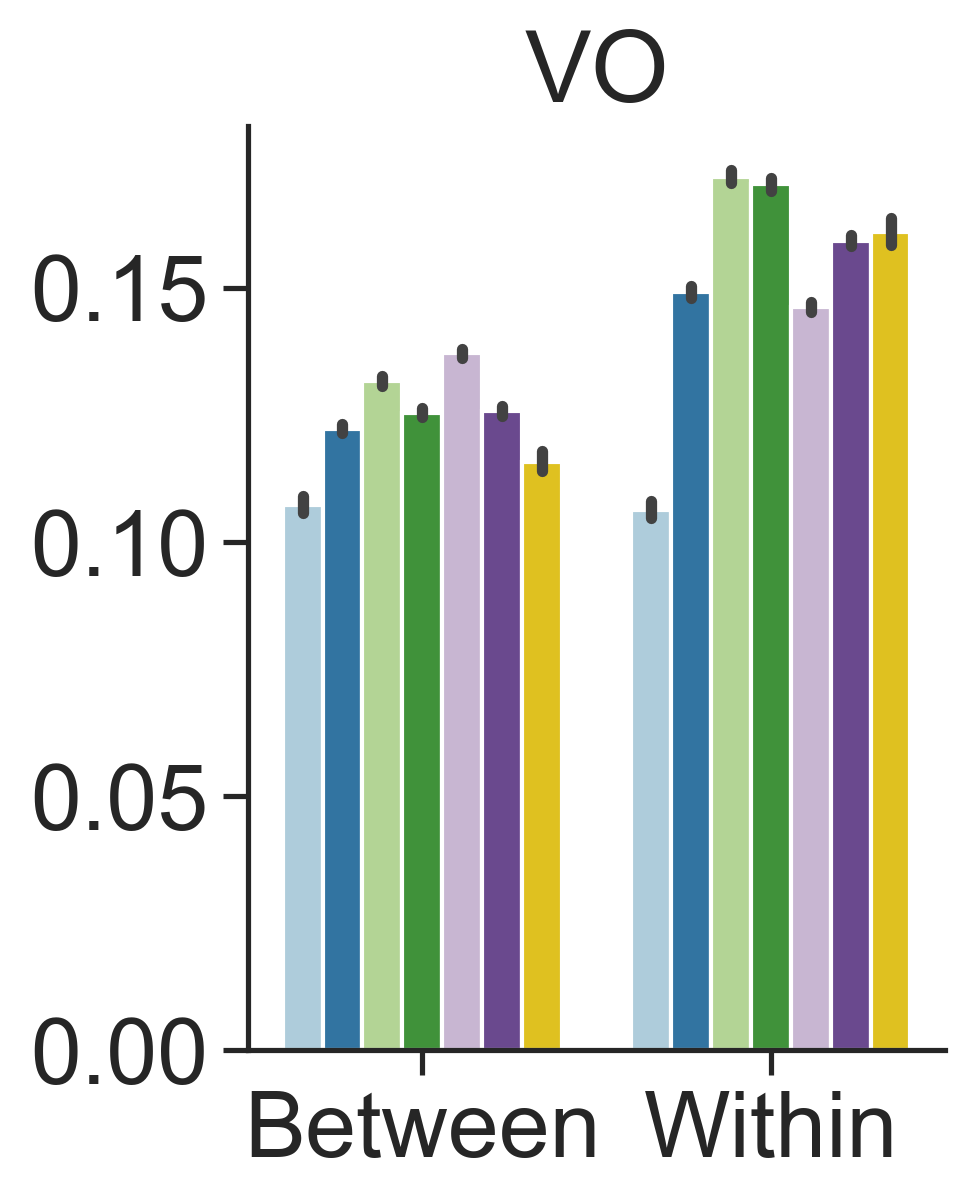}
    \includegraphics[height=2.4cm]{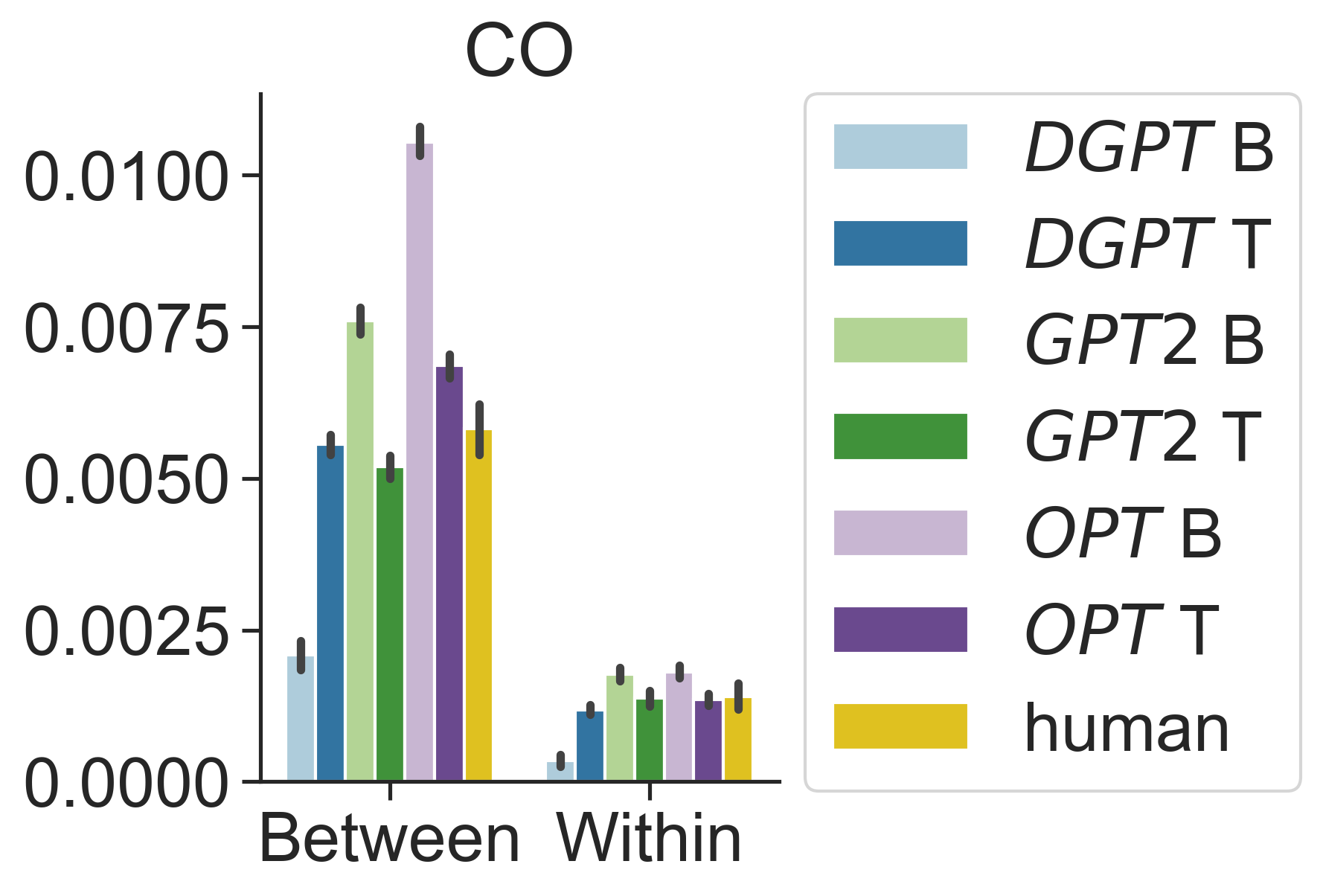}
    \caption{
    Human and model repetition properties. \textit{B} indicates base models, \textit{T} tuned models. 
    }
    \label{fig:repstats}
\end{figure}

\begin{figure*}
     \centering
     \begin{subfigure}[b]{0.23\textwidth}
        \centering 
            \includegraphics[height=2.3cm]{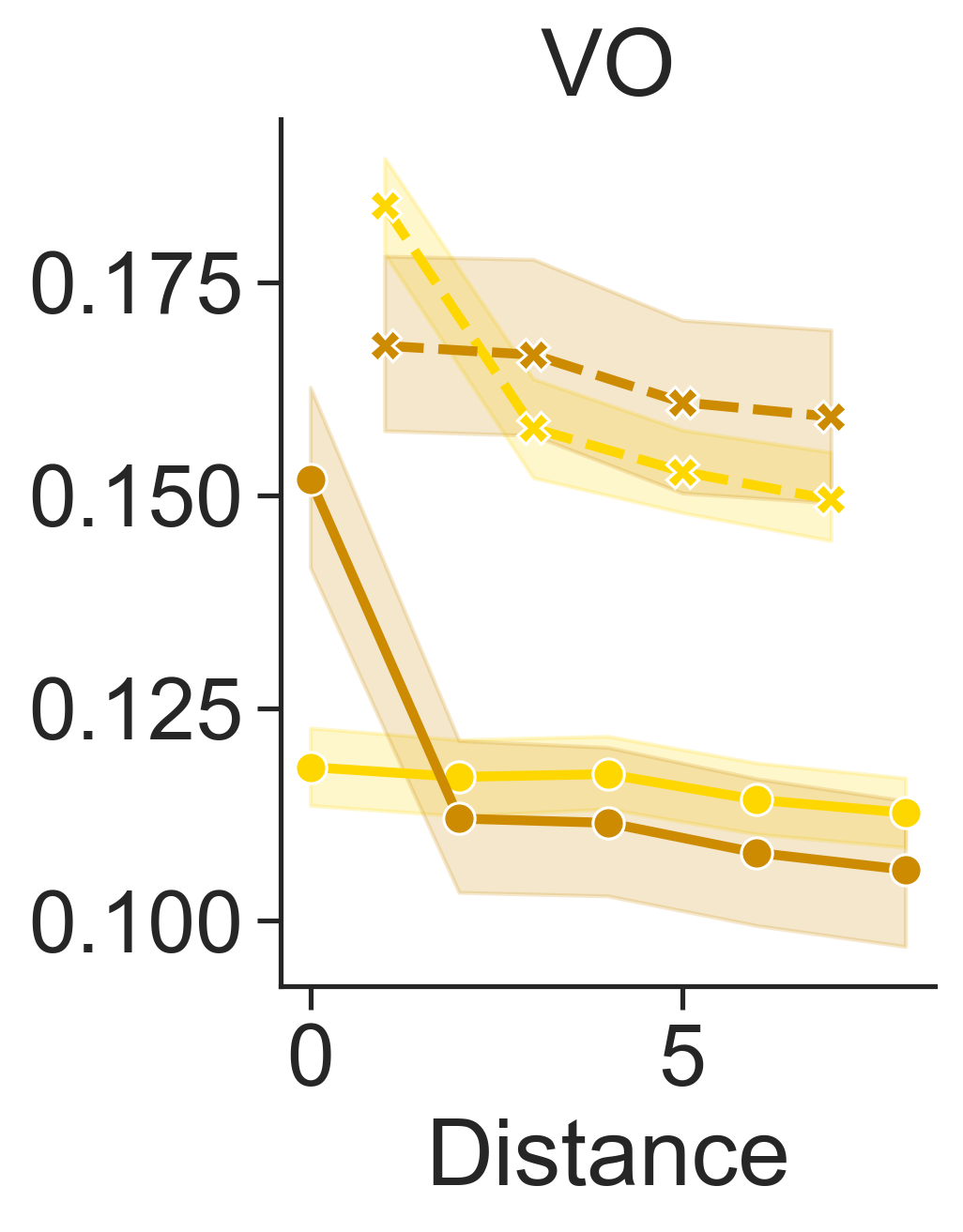}
            \includegraphics[height=2.3cm]{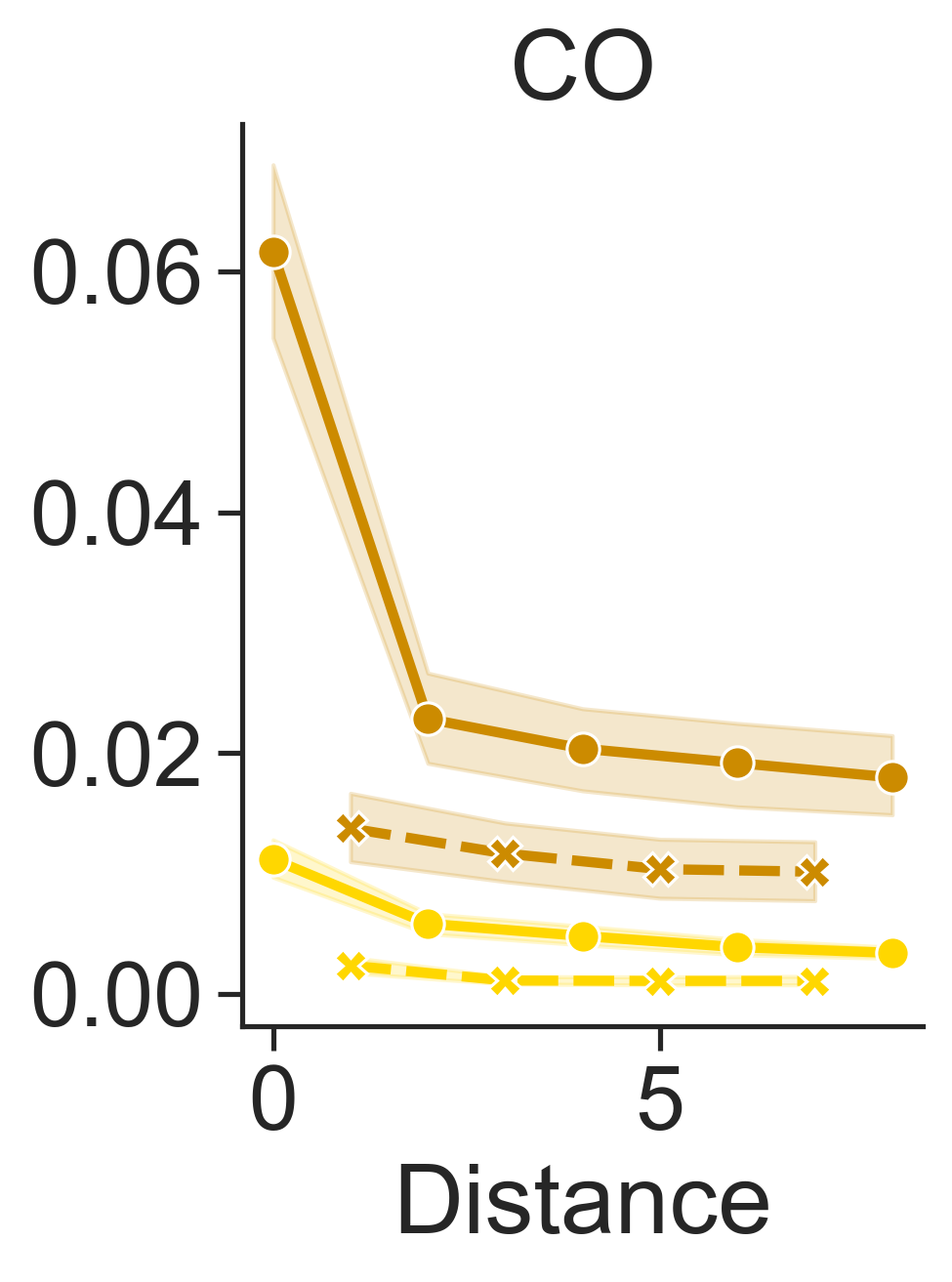}
            \includegraphics[height=2.3cm]{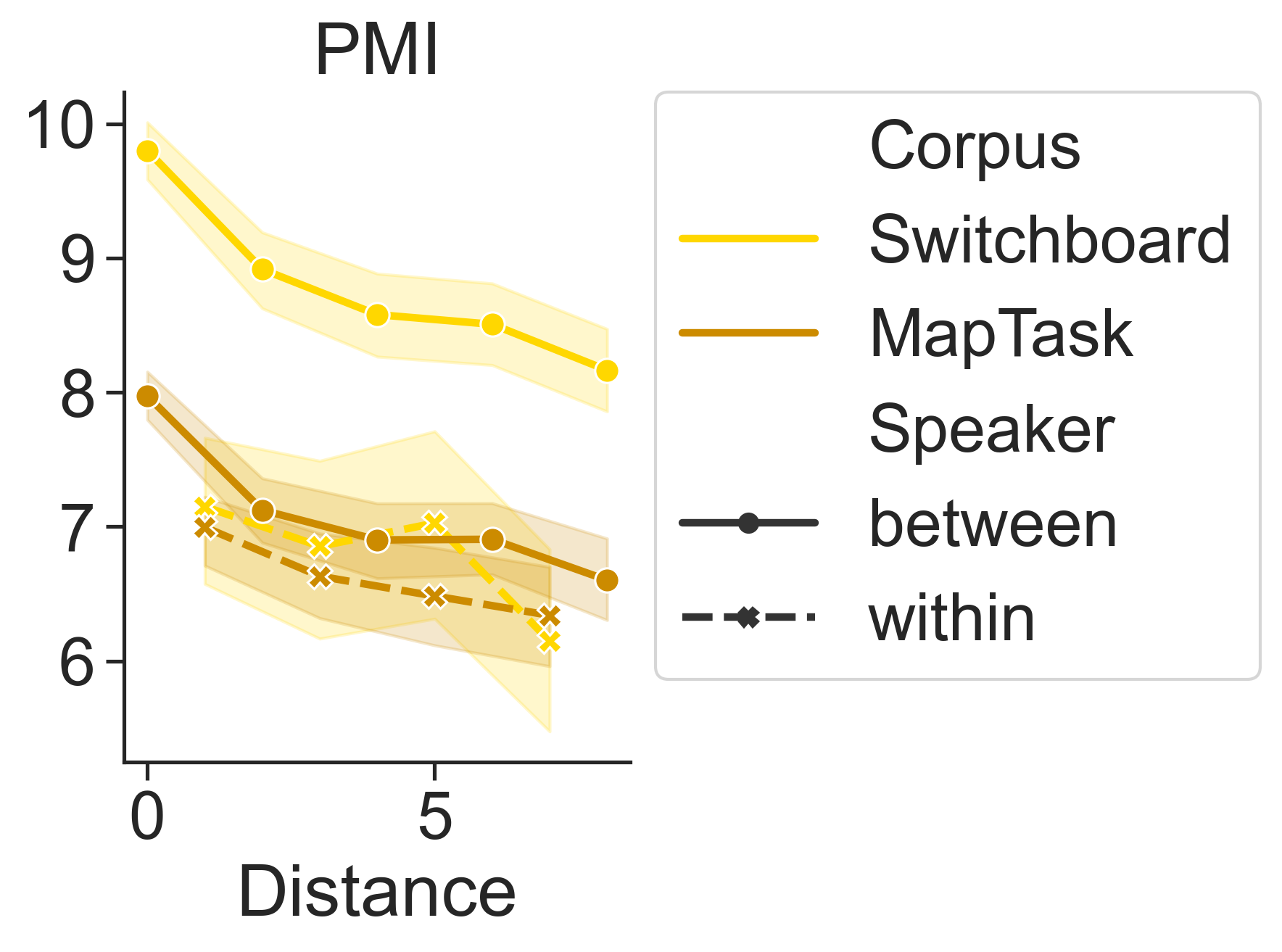}
        \caption{Human \co, \vo\ \& \pmi}
        \label{fig:vocohuman}
     \end{subfigure}
     \hfill
     \begin{subfigure}[b]{0.31\textwidth}
         \centering 
            \includegraphics[height=2.3cm]{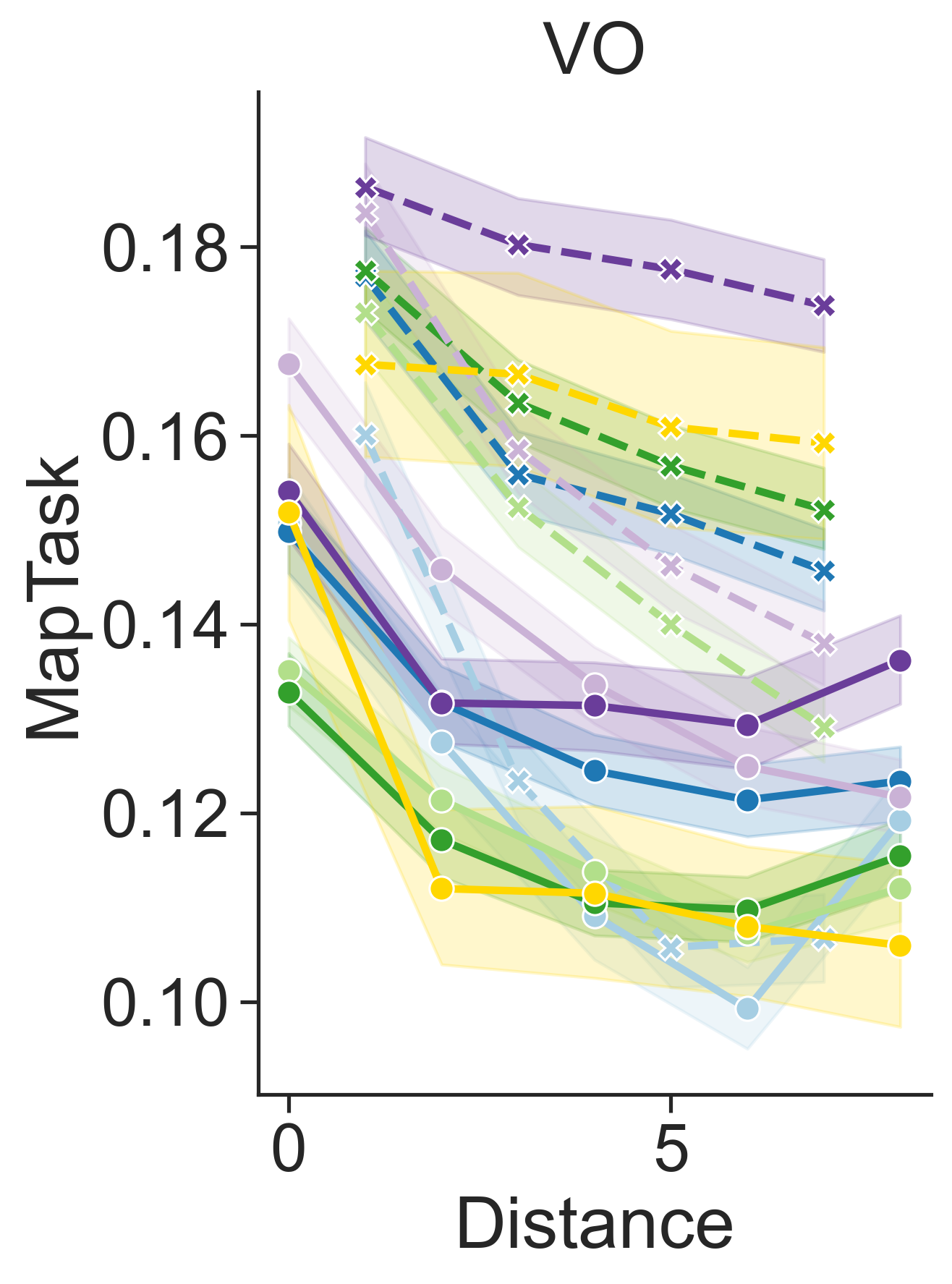}
            \includegraphics[height=2.3cm]{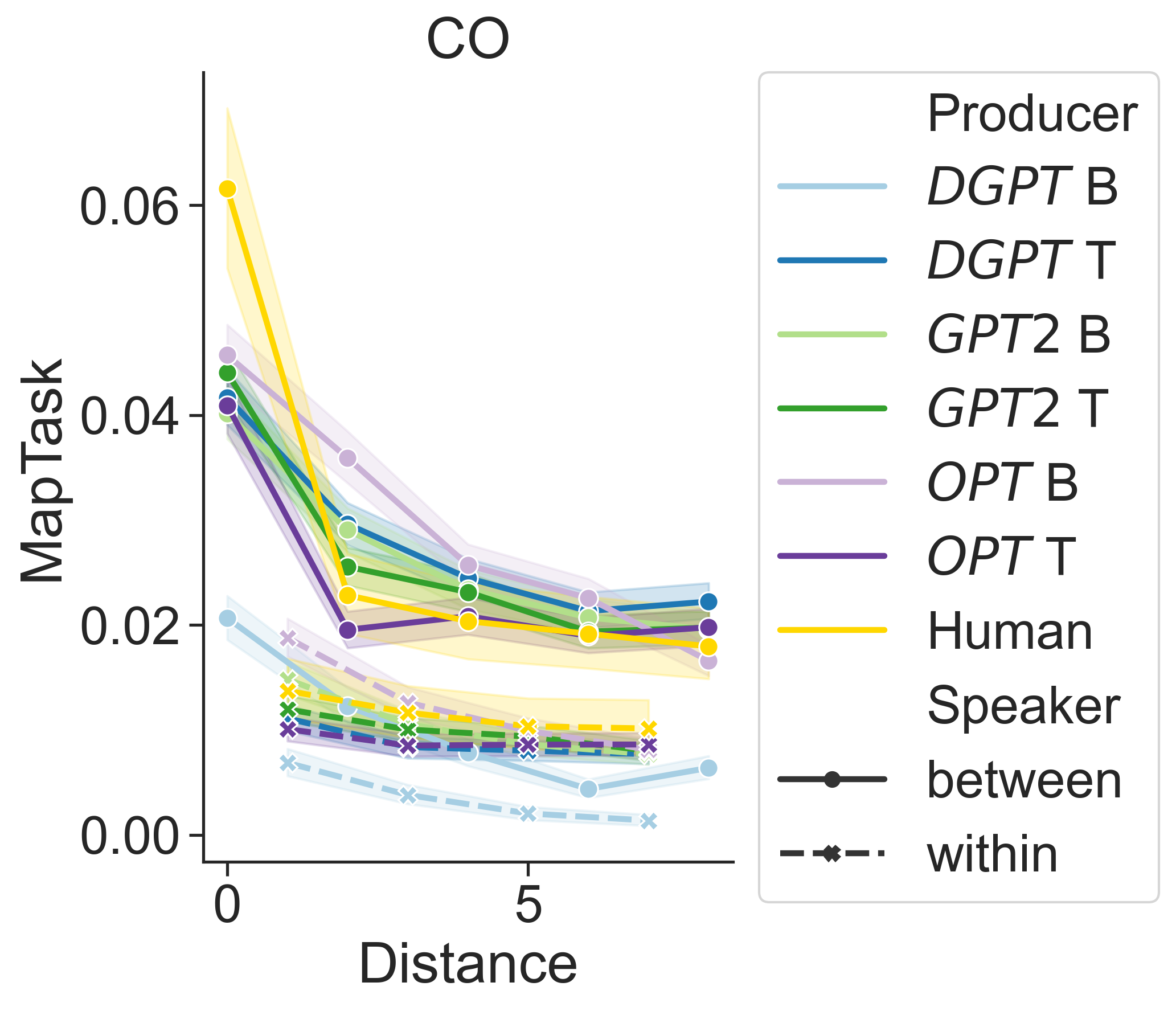}
            \includegraphics[height=2.3cm]{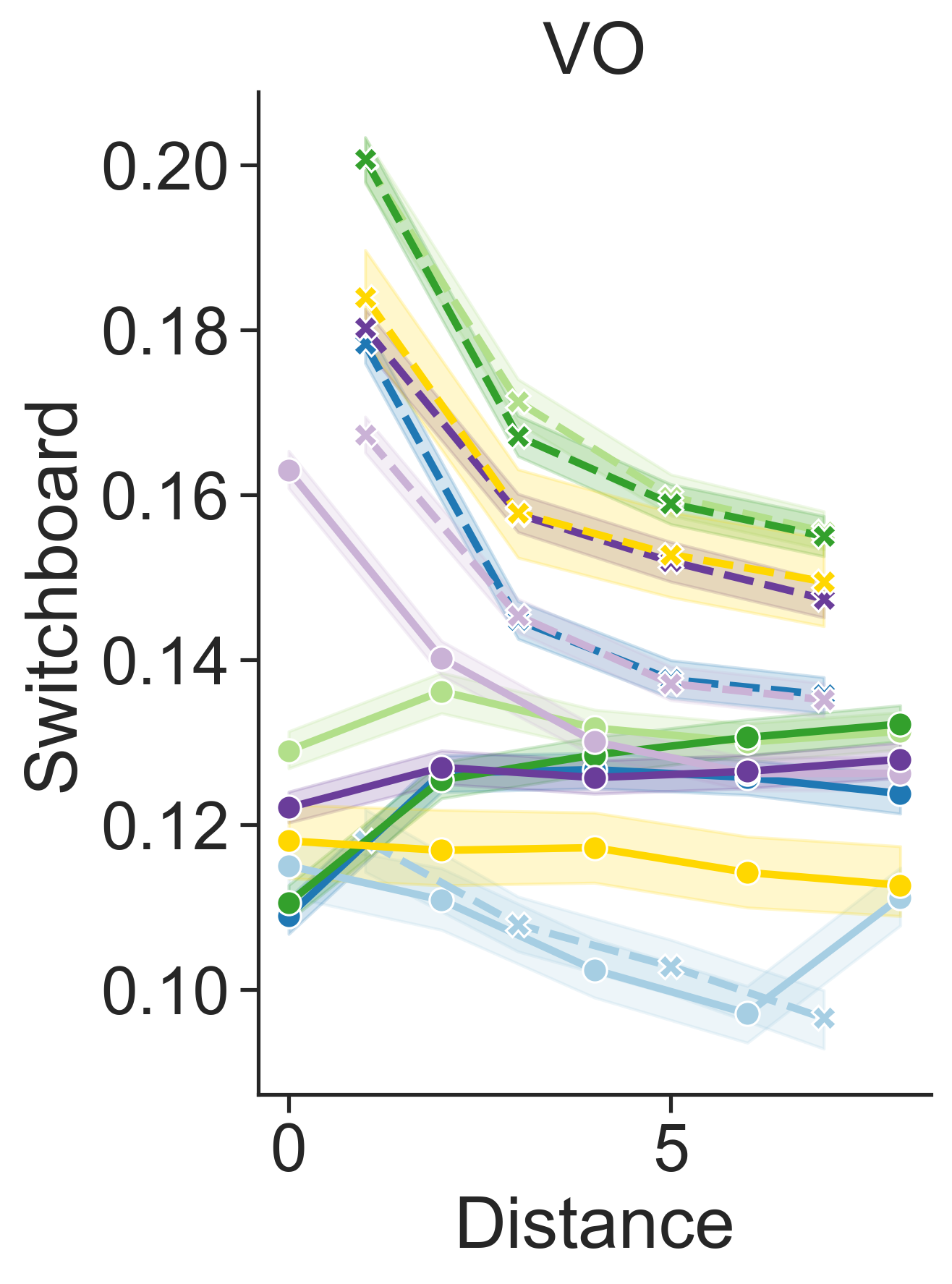}
            \includegraphics[height=2.3cm]{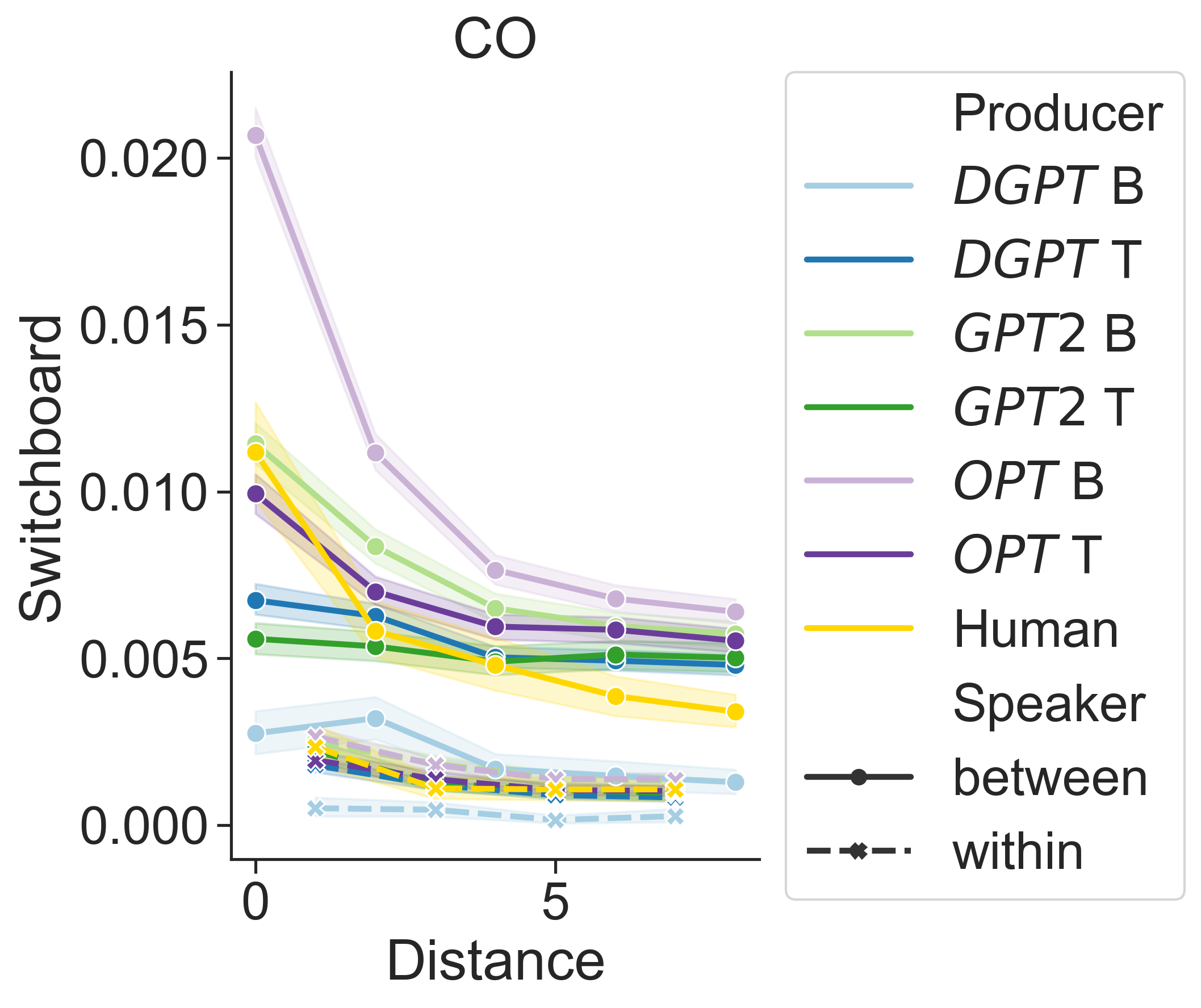}
        \caption{Human vs. Model \co\ \& \vo.}
        \label{fig:human_model_voco}
     \end{subfigure}
     \hfill
     \begin{subfigure}[b]{0.42\textwidth}
        \centering 
            \includegraphics[height=2.3cm]{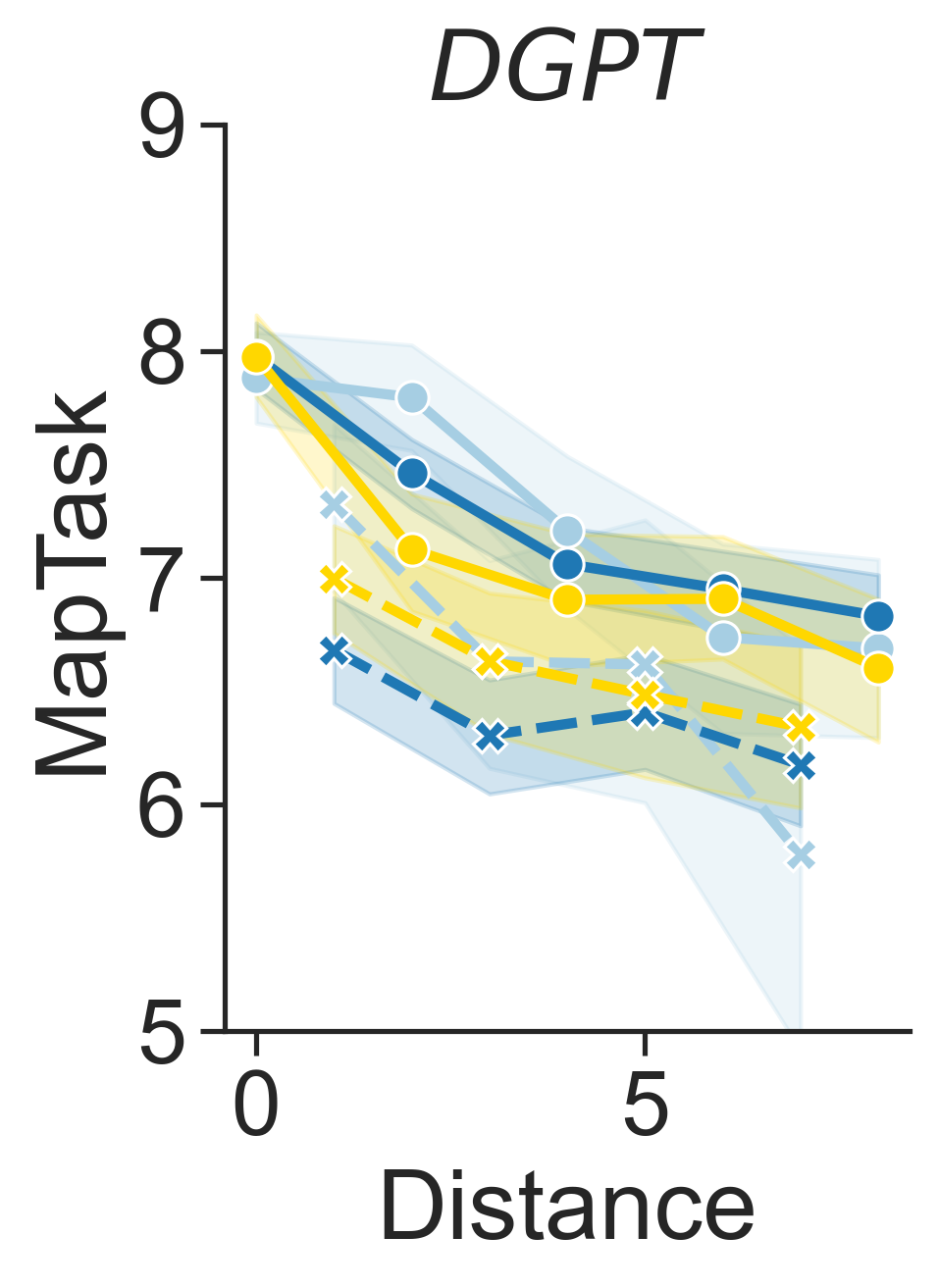}
            \includegraphics[height=2.3cm]{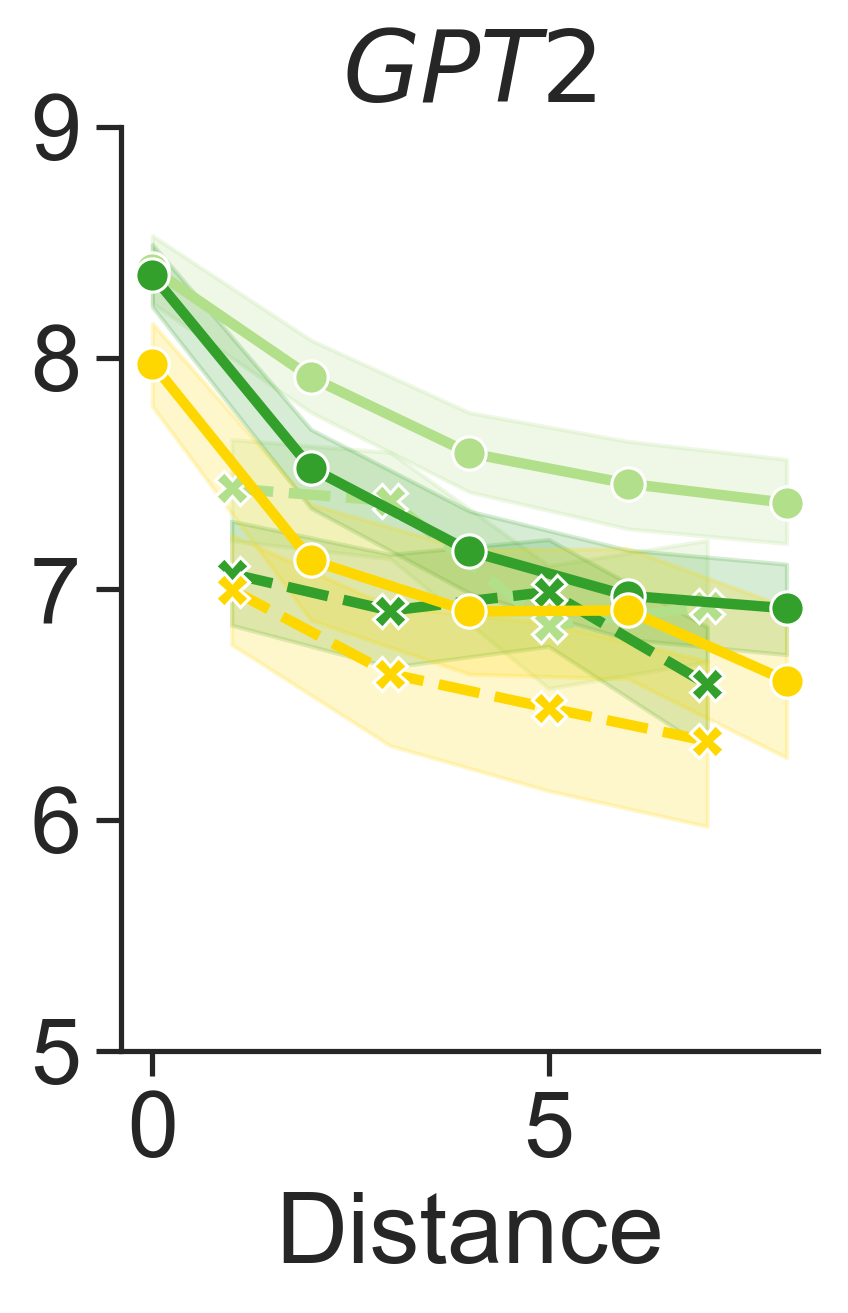}
            \includegraphics[height=2.3cm]{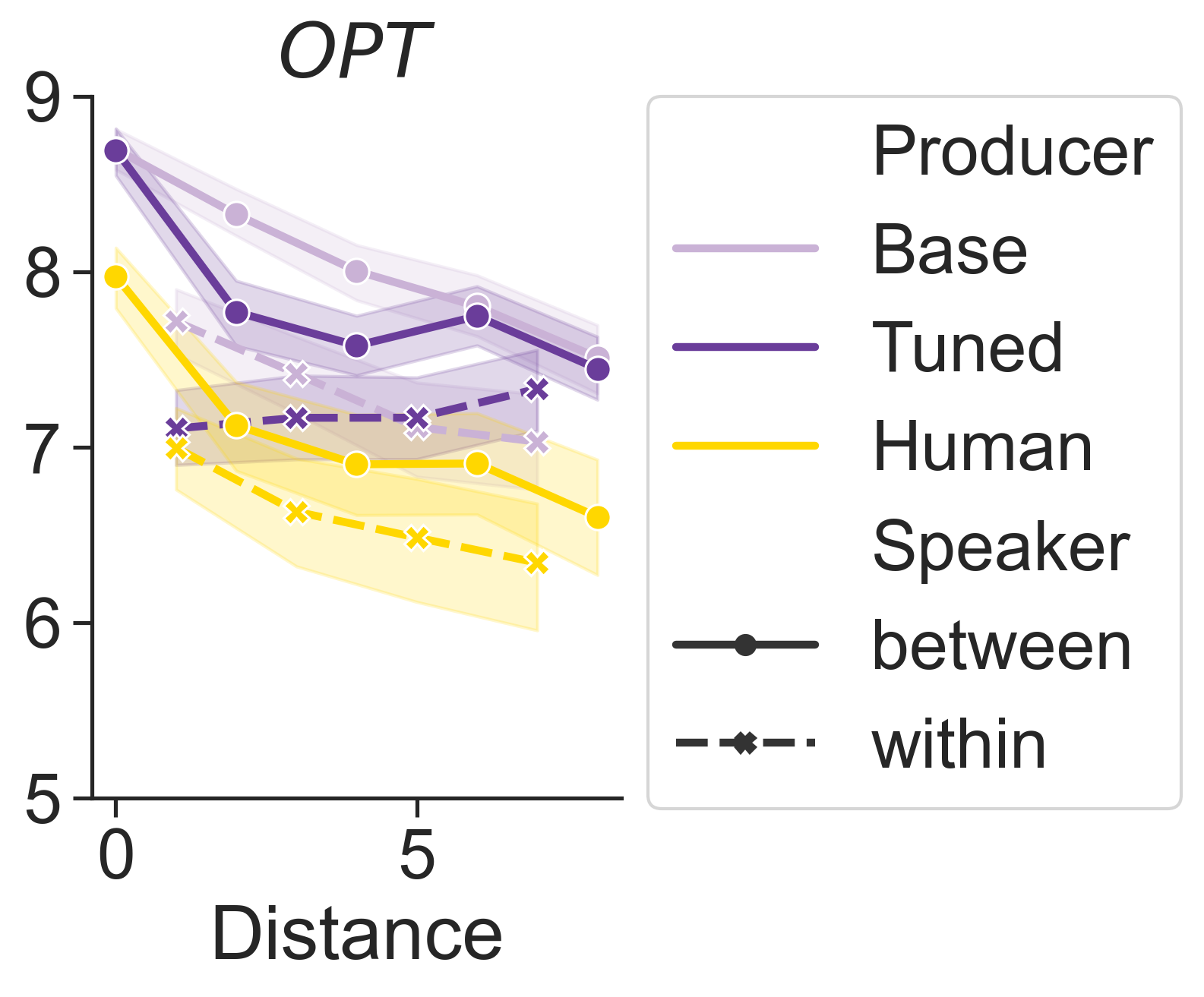}
            \includegraphics[height=2.3cm]{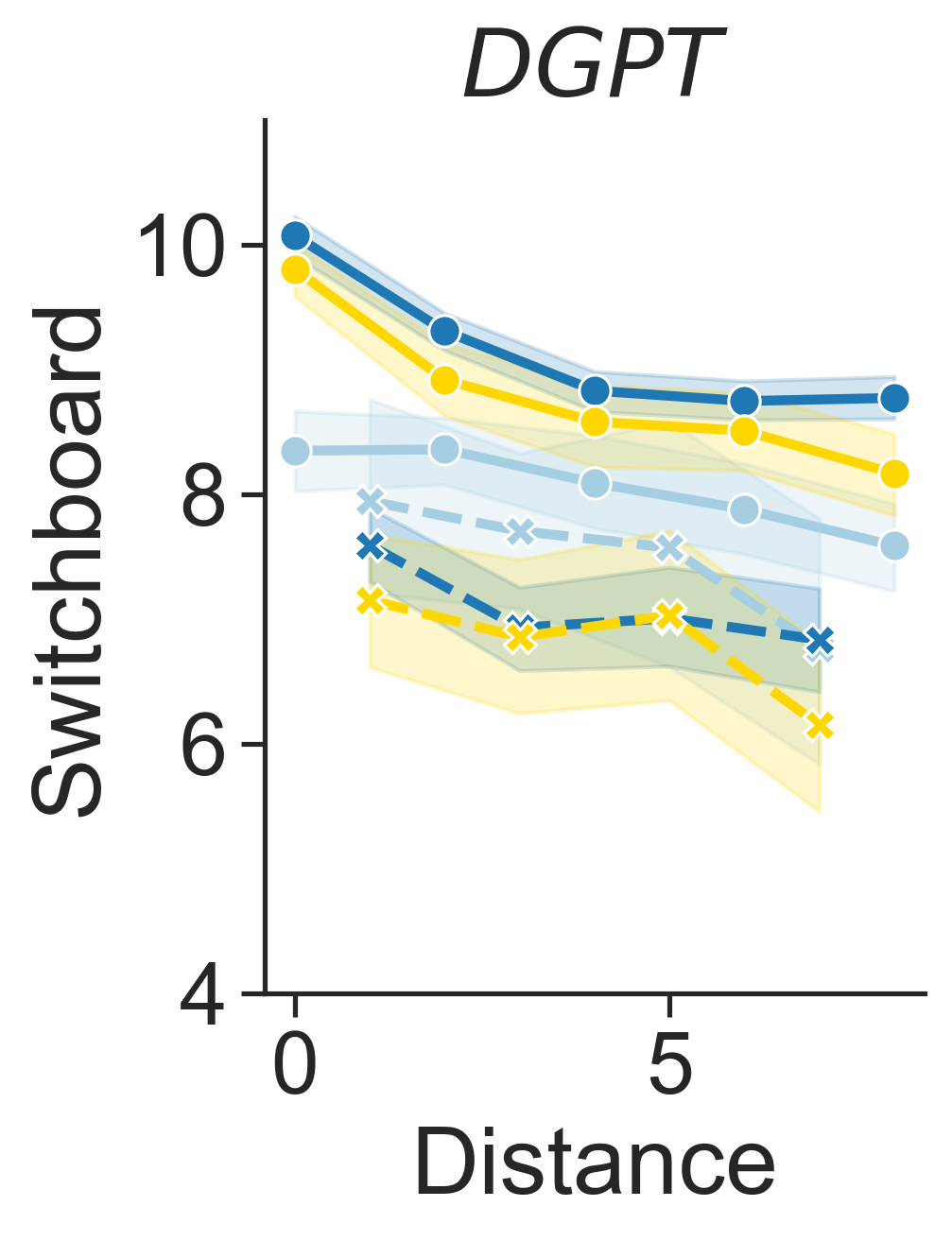}
            \includegraphics[height=2.3cm]{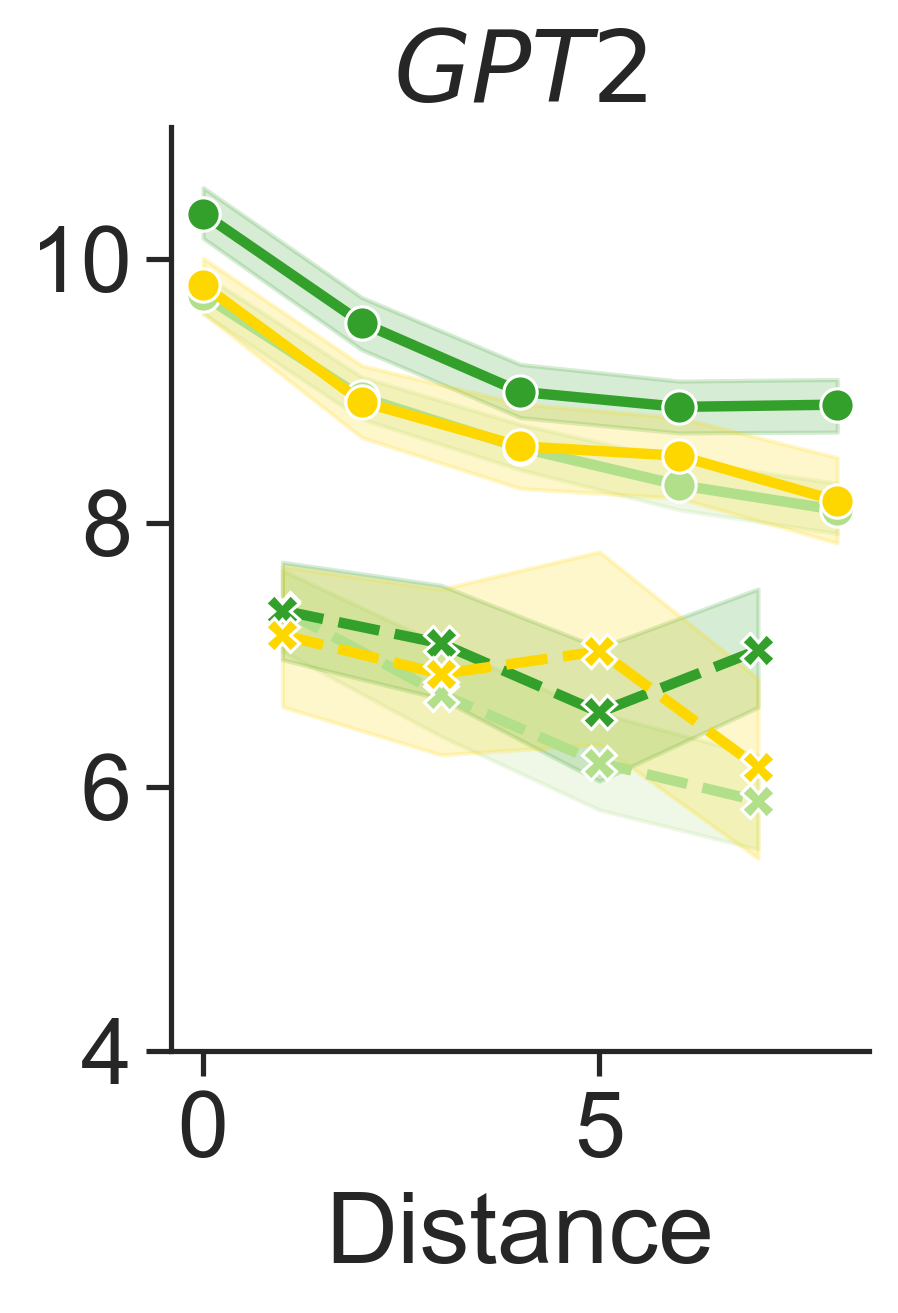}
            \includegraphics[height=2.3cm]{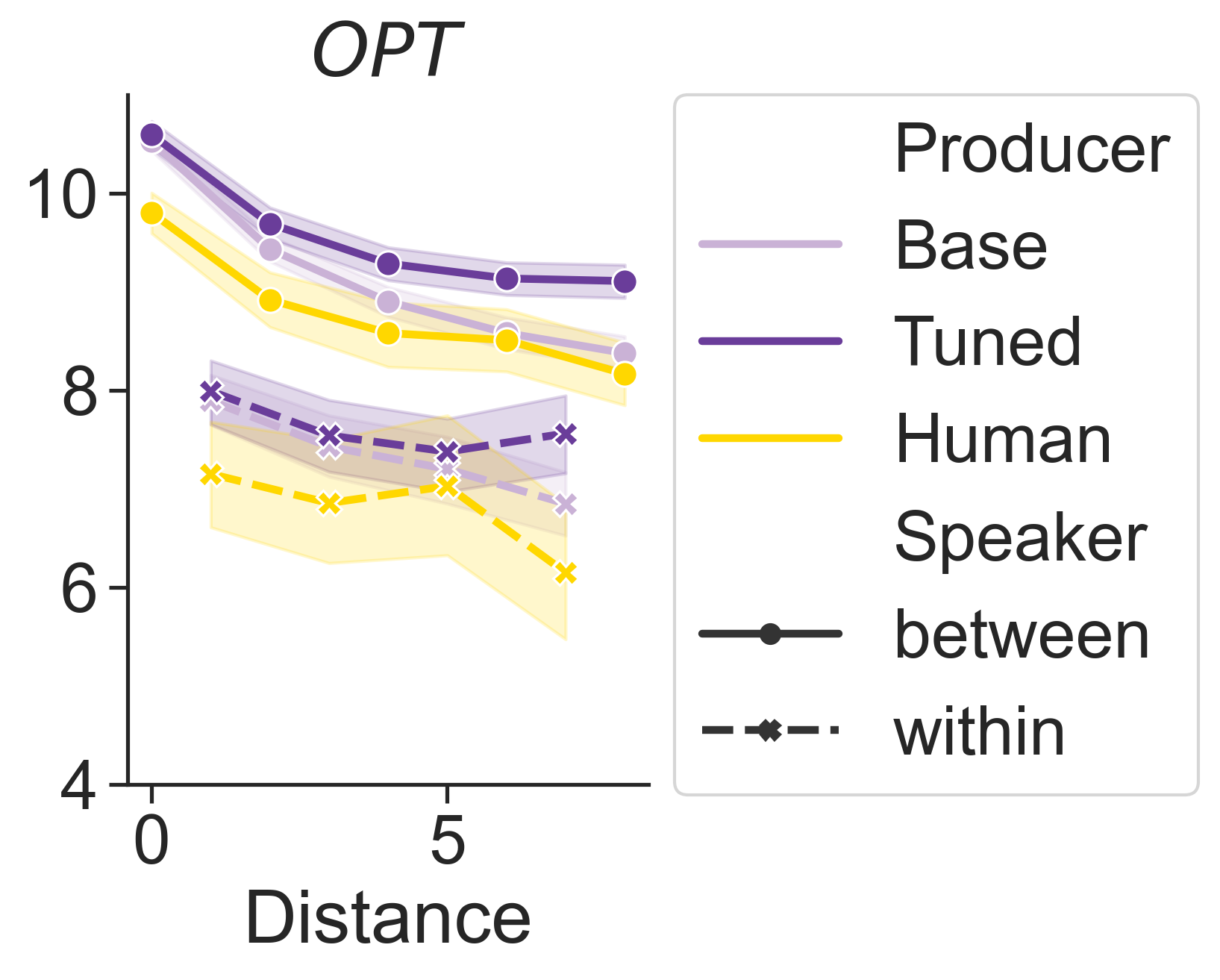}
        \caption{Specificity (\pmi) of repeated constructions. 
        }
        \label{fig:human_model_pmi_decay}
     \end{subfigure}
        \caption{
        Repetition effects for construction overlap \co\ and vocabulary overlap \vo.
        Patterns of human vs.\ model repetition across contexts.
        }
        \label{fig:reptrends}
\end{figure*}

\paragraph{\textit{Humans produce repetitions locally}.}
\label{sec_results_human_repetition}
To evaluate the \textit{local} effects of repetition, we employ linear mixed-effect models, including \textit{dialogue, sample} and \textit{speaker} identifiers as random effects.\footnote{
    Full model output can be found in Appendix~\ref{app:models}. We include dialogue, sample and speaker as random effects, to allow for group-level variability in the linear model.
    }
We confirm that \co\ decays with the distance between a given utterance and those preceding it~($ \beta = -0.001$, $p<0.05$, $95\%\ CI = [-0.001\!:\!-0.001]$); this is not the case for \vo\ 
(Figure~\ref{fig:vocohuman}). 
Decay effects for \co\ are stronger for between-speaker repetition in both corpora. That is, speakers are more likely to repeat their partner's language locally. 
Interestingly, in \sw, decay effect are not observable when looking at the dialogue as a whole~\citep{sinclair2021construction}.
We hypothesise that other, less locally repeated constructions may drive down this effect when analysing the dialogues as a whole, or that some constructions may have multiple short bursts of local repetition over the course of a dialogue~\cite{pierrehumbert2012burstiness}.\looseness-1
%

\paragraph{\textit{Models learn some patterns of local repetition}.} 
\label{sec_results_model_repetition}

We find that fine-tuned models learn 
turn-sensitive patterns of local repetition to some extent. 
Figure~\ref{fig:human_model_voco} demonstrates that models can learn similar patterns of local repetition to those observed in human dialogue. The most dramatic improvement in similarity to human behaviour is for \dgpt.

We find that in \sw, both models and humans show significant \textit{local} repetition effects of \co\, independent of \vo\ effects. 
Investigating \co\ in more detail, while human repetitions are sensitive to the length of the construction (longer constructions predict \co: $ \beta=0.035$, $p<0.05$, $95\%\ CI = [0.025\!:\!0.045]$), this is not the case for models, for which the frequency of the repetition in the sample plays an important role in predicting \co\ (e.g. \gpt\ repetition frequency: ($ \beta=0.01$, $p<0.05$, $95\%\ CI = [0.007\!:\!0.013]$)). 
For \mt, we find that humans repeat highly specific repetitions locally
(\co\: $ \beta=0.006$, $p<0.05$, $95\%\ CI = [0.003\!:\!0.009]$), however this is only true for \gpt\ ($ \beta=0.001$, $p<0.05$, $95\%\ CI = [0.0\!:\!0.002]$).
Full model results in Appendix~\ref{app:repetition_model}.

\paragraph{\textit{Models don't consistently produce speaker-specific repetitions}.}
\label{sec:speaker_repetition_results}
We find that while all models display significant \co\ speaker effects similar to humans, when taking into account other contextual factors, their behaviour with respect to \textit{specificity} varies. While Figure~\ref{fig:human_model_pmi_decay} demonstrates that the \pmi\ of constructions decays with distance, human speakers show no significant independent effect of \pmi~when predicting \co\ 
in either corpus. 
\gpt~exhibits the most similar behaviour to the human data in terms of the effect of distance and speaker on \pmi~in \mt, however learns a significant negative relationship with \pmi~for \sw, not present in the human data. 
Full model results in Appendix~\ref{app:repetition_model}

\begin{table}[ht]\centering \small \scalebox{0.65} { 
\setlength{\tabcolsep}{2pt}
\begin{tabular}{@{}ll r|rrrrrrrr@{}} \toprule 
& & $PPL_{m}$ $\downarrow$ & $PPLg_{ii}$ $\downarrow$ & $PPLg_{id}$ $\downarrow$ & $PPLp_{ii}$ $\downarrow$ & $PPLp_{id}$ $\downarrow$ & BLEU & BertF1 & Mve \\ \midrule
\textit{SW} & & &  & & & & & & \\ 
\hspace{2mm} GPT2 & B & 15.110 & \textbf{3.770} & \textbf{2.870} & 60.879 & 12.985 & 0.009 & 0.710 & 0.035 \\ 
& T & \textbf{12.020} & 3.830 & 2.880 & \textbf{50.608} & \textbf{12.790} & \textbf{0.010} & \textbf{0.730} & \textbf{0.049} \\
\hspace{2mm} OPT & B & 37.540 & \textbf{3.750} & 2.870 & 54.706 & 12.799 & 0.010  & 0.700      & 0.052 \\ 
& T & \textbf{15.130} & 3.830 & 2.870 & \textbf{45.488} & \textbf{12.635} & \textbf{0.014}  & \textbf{0.733}    & \textbf{0.069} \\
\hspace{2mm} DGPT & B & 6935.000 & 7.050 & 2.970 & 1323.338 &  14.064  & 0.000  & 0.656    & 0.006 \\ 
& T & \textbf{10.910} & \textbf{3.570} & \textbf{2.870} &  \textbf{41.700} &  \textbf{12.735}  & \textbf{0.016} & \textbf{0.730}   & \textbf{0.049} \\
\textit{MT} & & & & & & & & & \\ 
\hspace{2mm} GPT2 & B & 16.170 & \textbf{4.920} & \textbf{3.190} & \textbf{136.421} & 18.353 & 0.006  & 0.679   & 0.101 \\ 
& T & \textbf{7.930} & 5.250 & 3.220 &  208.630 & \textbf{18.193}  & \textbf{0.014}  & \textbf{0.702}   & \textbf{0.245} \\
\hspace{2mm} OPT & B & 72.100 & \textbf{5.270} & \textbf{3.210} &  \textbf{199.344} &  \textbf{18.189}  & 0.006  & 0.682   & 0.103 \\ 
& T & \textbf{9.700} & 5.730 & 3.240 & 294.677  & 18.384  & \textbf{0.016}  & \textbf{0.712}  & \textbf{0.339} \\
\hspace{2mm} DGPT & B & 13014.000 & 6.670 & 3.280 & 998.832  & 19.852   & 0.002  & 0.662  & 0.041 \\ 
& T & \textbf{8.050} & \textbf{5.320} & \textbf{3.220} &  \textbf{235.385} &  \textbf{18.007}  & \textbf{0.016} & \textbf{0.699}  & \textbf{0.176} \\ 
\bottomrule \end{tabular}}
\caption{Generation quality results. \textit{SW}: SwitchBoard. \textit{MT}: MapTask. $PPL_{m}$: Perplexity of the models under scrutiny
on the analysis set. 
Perplexity of \gpt~($PPLg_{ix}$) and \textsc{Pythia} ($PPLp_{ix}$) on model-produced utterances ($ii$ independent of, and $id$ dependent on context). 
\textit{B}: base models, \textit{T}: fine-tuned models. \textit{Mve}: MAUVE score. \textbf{Bold} indicates the better value between base and fine-tuned variants.
} 
\label{tab:model_metrics_main} 
\end{table}

\subsubsection{Repetition vs.\ Quality}
Finally, we investigate whether automatic NLG metrics capture human-likeness of repetition. This is an important aspect of naturalness in dialogue which the metrics are not explicitly designed for.
Table~\ref{tab:model_metrics_main} shows the relative generation quality of our base and fine-tuned models. Extended results can be found in Appendix~\ref{app_lm_finetune}. 
All models demonstrate improvement with fine-tuning, although \gpt~base as an evaluator detects less difference than Pythia. 
This is expected, given their training data contains either little dialogue data, or a comparatively very different style of dialogue.

We find that the closer the levels of \co\ and \vo\ are to human-produced language,\footnote{
     We measure this as the absolute value of the difference between human and model values.
} the higher \textit{BertF1}, \textit{BLEU}, and the lower the evaluation model perplexity both dependent and independent of the context.
This correlation is strongest for \textsc{gpt2} with $\rho = -0.395$, $p < 0.05$ for \vo\ and $\rho = -0.258$, $p < 0.05$ for \co. 
This is perhaps to be expected for reference-based metrics, so we additionally inspect whether human-like \co\ levels correlate with MAUVE, a corpus-level metric, finding that more similar \co\ levels between human and model \textit{inversely} correlate with MAUVE quality (above $\rho =0.7$, $p<0.05$ across models).\footnote{
    Table~\ref{tab:mauve} in Appendix~\ref{app:mauve} provides a detailed breakdown of these results.
} This tells us either that better corpus-level metrics need to be defined or, perhaps, that corpus-level evaluation is not really appropriate for dialogue where quality is determined by  local and highly contextually dependent cues. This is in keeping with challenges in evaluating dialogue~\cite{zhang2021d,liu2016not}, and suggests standard NLG evaluation approaches should be complemented by dialogue-specific metrics like the ones we use in our analysis. 

%% file: sections/results_attribution.tex
\section{Interpreting Model Comprehension Behaviour}
\label{sec:results_attribution}
In the previous section, we investigated patterns of repetition in models' production behaviour. Now we turn our attention to their \textit{comprehension} behaviour, making use of interpretability techniques to analyse what properties of the utterances in the context are more salient in determining expectations for a given target utterance. 
We expect models to learn patterns of turn-taking from the structure and contents of the context utterances~\cite{wolf2019transfertransfo,ekstedt2020turngpt,Gu2020speakerawarebert}. We also expect that higher salience will be assigned to 
repetitions with local antecedents, in line with recency effects observed in model priming behaviour~\citep{sinclair2022structural}.

\subsection{Methods}
\label{sec:attribution_method}
\subsubsection{Feature Attribution}
We obtain attributions over the dialogue context for a given target utterance, extracting scores for each token over the entire preceding context.\footnote{
    For creating the attributions we make use of  \texttt{Inseq} \citep{inseq} and \texttt{Captum} \citep{kokhlikyan2020captum}.
} 
We are interested in examining behavioural patterns at the utterance level, in order to investigate the influence of their distance from the target, and design a measure to capture the \textit{relative} boosting effects of the context for a given target utterance.
This approach allows us to inspect attribution patterns 
across the context with respect to properties of the target utterance 
as a whole, allowing us to conduct similar, 
complementary analyses to the previous section. \looseness-1

A wide range of feature attribution methods exist \citep{DBLP:conf/nips/LundbergL17, doi:10.1073/pnas.1900654116}. 
It remains an open question, however, which of these methods are most faithful with respect to the true model behaviour~\citep{DBLP:conf/emnlp/BastingsEZSF22}.

Some methods resolve this through defining theoretical properties that need to be satisfied by the method~\citep{DBLP:conf/icml/SundararajanTY17}.
We focus on one such method, \textit{DeepLift} \citep{Shrikumar2017LearningDifferences}, which, besides its attractive theoretical properties, is also considerably more compute friendly than alternative attribution methods.\looseness-1

\subsubsection{Attribution Aggregation Procedure}

We design a measure that allows us to capture the relative effects that individual utterances in the local context have on models' utterance comprehension.
Our measure aggregates over per-token attributions for a full utterance, returning relative prediction boosting effects of tokens within context utterances, speaker label tokens, and the target itself.\looseness-1

A given sample will consist of \textit{speaker label tokens}, indicative of the change in speaker, e.g. \textit{`A:'} and \textit{`B:'}, the 9 context utterances, 
and the target utterance text. This can look like the following, with the speaker label tokens in \yellow{orange}, context utterances in \green{dark blue}, and the final target utterance of interest in \bluelight{light blue}: \newline
\yellow{
\textit{A: \green{how are you?} B: \green{great, it's sunny} A: \green{about time} B: \green{agreed.} A: \green{I love sun} B: \green{me too} A: \green{makes me think of the beach} B: \green{the beach is great} A: \green{so great} B:{\bluelight{great, we should go to the beach!}}}}\\

Firstly, we create the feature attribution scores of each token in the input $w_i$ with respect to the prediction of each token in the target utterance $w_t$:\looseness-1
\begin{equation}
\small
    \Phi \in \mathbb{R}^{|w_i|\times|w_t|\times n_{emb}}
\end{equation}
Since feature attribution methods provide an importance score on the embedding level, we sum these scores along the embedding dimension $n_{emb}$.\footnote{
    We could opt for the L2 norm as well, but this would hide negative contribution effects \citep{DBLP:conf/emnlp/BastingsEZSF22}.\looseness-1
}
Next, we sum the $\Phi$ matrix along the dimension of the tokens in the target utterance ($w_t$): creating a single score for each input token with respect to the target as a whole.
Then, we create a single importance score for each individual input utterance or turn separator, denoted as a set $T_i$ that contains the indices of the $i^{th}$ utterance:\looseness-1
\begin{equation}
\small
    \Phi'\in\mathbb{R}^{|T|},~~~\Phi'_i = \sum_{j\in T_i}\sum_k\sum_l \Phi_{j,k,l}
\end{equation}
Note that the target utterance itself also yields importance scores of earlier tokens in the target with respect to later predictions.

The scores of $\Phi'$ are still unbounded, and can vary greatly between samples and models. 
We apply two further operations to allow sample and model comparison:
we normalise the scores by the maximum absolute $\Phi'$ score, which maps the scores between -1 and 1, and we then centre the scores around the mean.
This expresses the contribution of each element in the input as its \textit{relative boosting effect} with respect to the other elements in the input\looseness-1
\begin{equation}
    \small
    \Phi'' = \frac{\Phi'}{\max\left(|\Phi'|\right)}
\end{equation}
\begin{equation}
    \small
    \phi = \Phi'' - \textup{mean}(\Phi'')
\end{equation}

\subsection{Analysis}
\label{sec:attribution_results}
We now investigate model attribution patterns over the dialogue context. 
Our goal is to find out whether a model's comprehension behaviour exhibits robust patterns explainable through known psycholinguistic effects thought to influence human language producers, in particular \textit{local, between-speaker} repetition patterns.
While we are currently unable to understand precisely where humans place salience when comprehending, a large body of psycholinguistic research points to patterns of priming and alignment behaviour detectable from brain signals~\citep{hasson2012brain,futrell-etal-2019-neural}, and uses our understanding of the brain to inform analysis of neural language models~\citep{hasson2020direct}.
We will contrast this analysis of model comprehension behaviour to the previous study of their production behaviour. We expect tuned models, the more human-like producers, to comprehend human language in a manner better predicted by factors thought to influence human processes---such as locality and priming effects---than base models. \looseness-1

\newcommand{\tl}{${\textsc{Speaker Label}}$}
\newcommand{\utt}{${\textsc{Utterance}}$}
\subsubsection{Attributions Over Human Utterances}
\begin{figure}[t]
    \centering 
        \includegraphics[height=2.59cm]{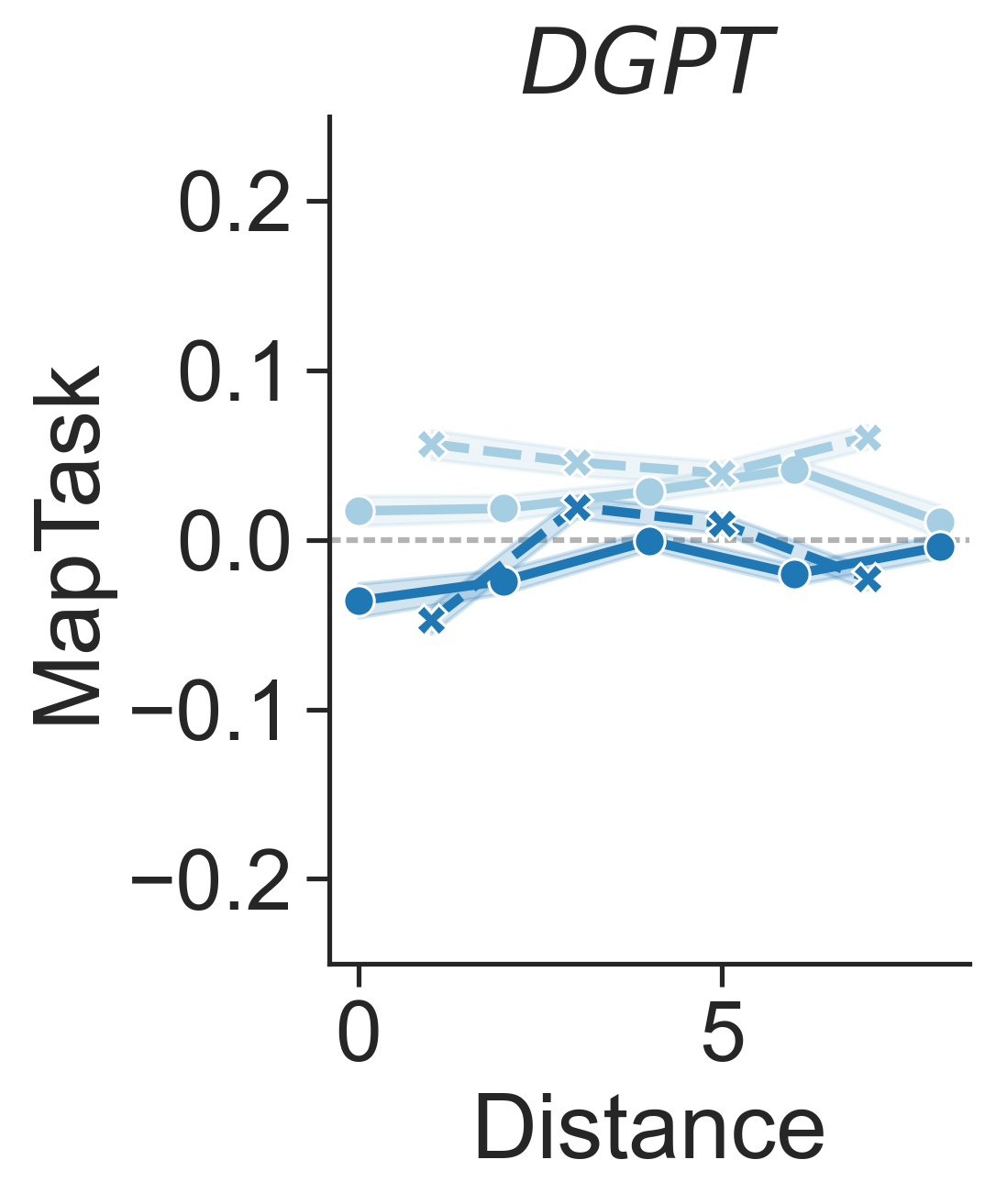}
        \includegraphics[height=2.59cm]{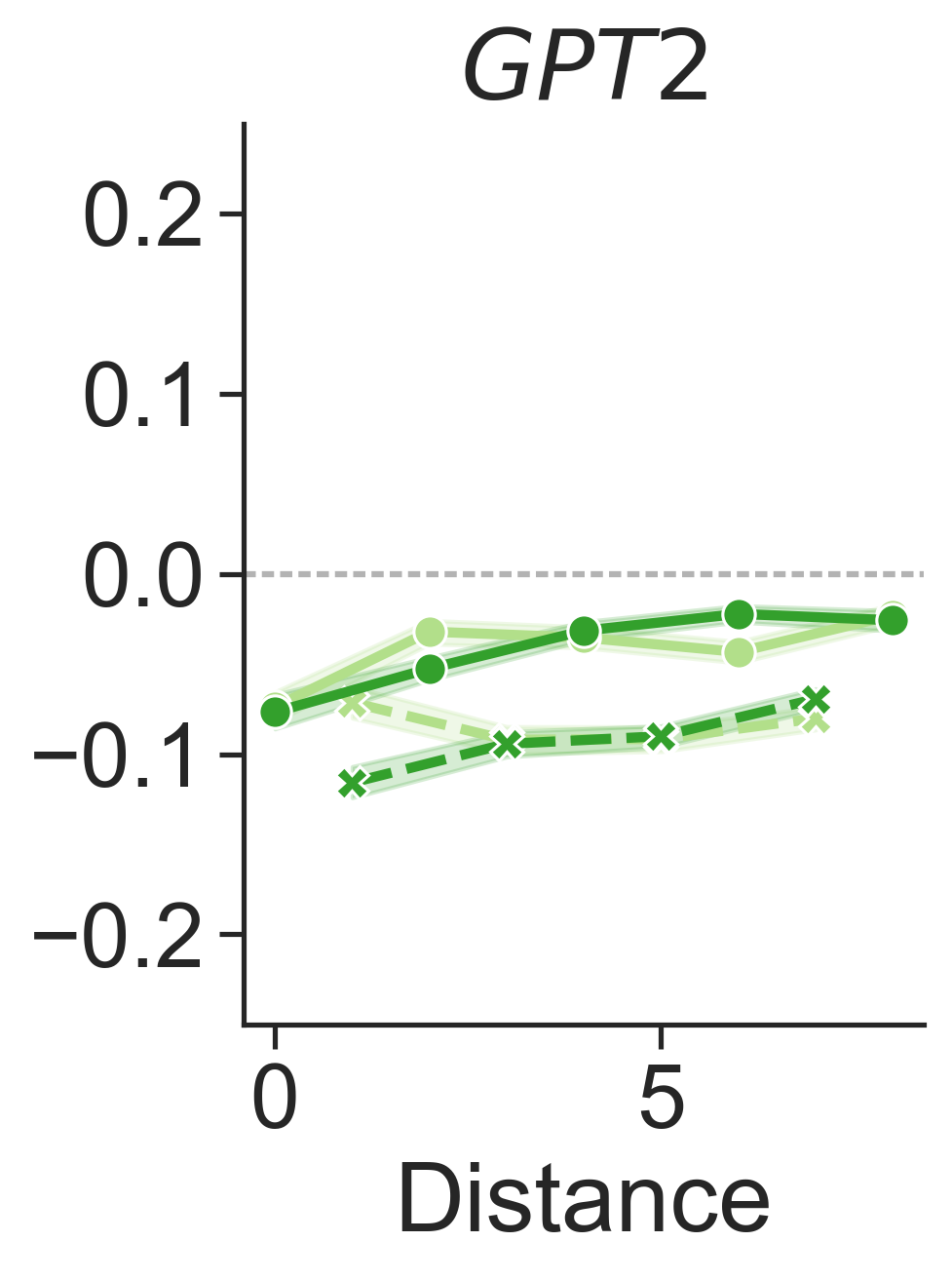}
        \includegraphics[height=2.59cm]{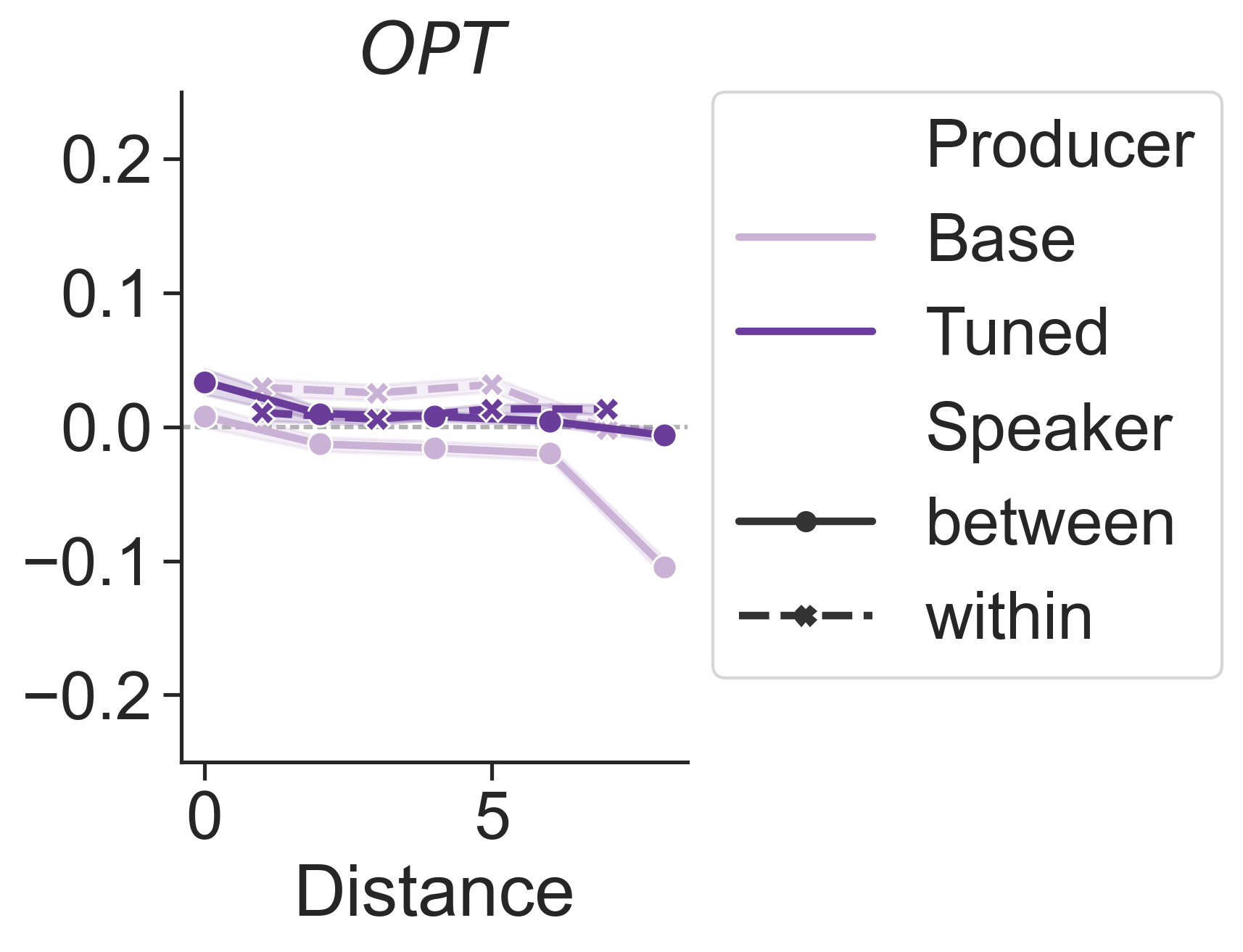}
        
        \includegraphics[height=2.59cm]{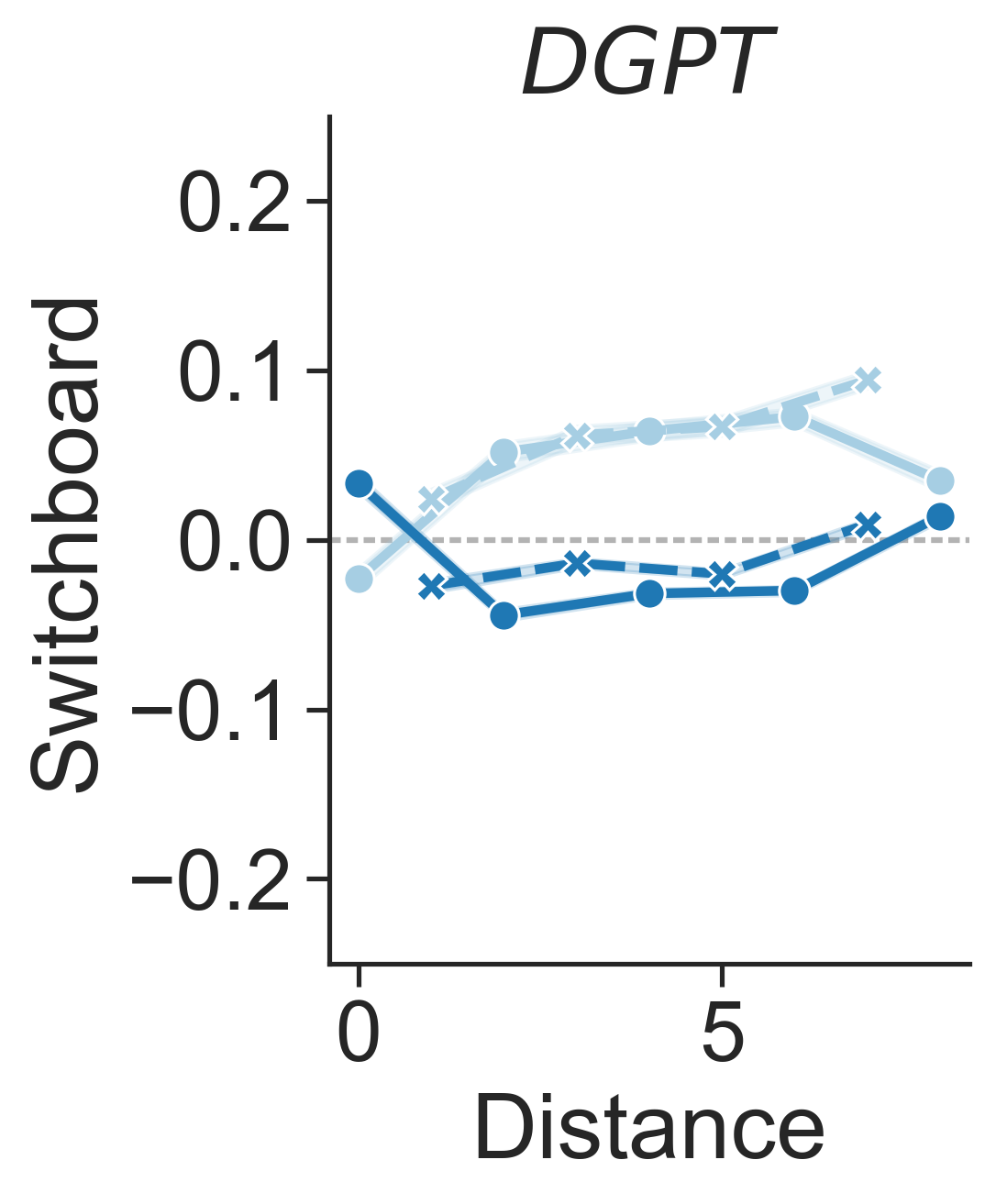}
        \includegraphics[height=2.59cm]{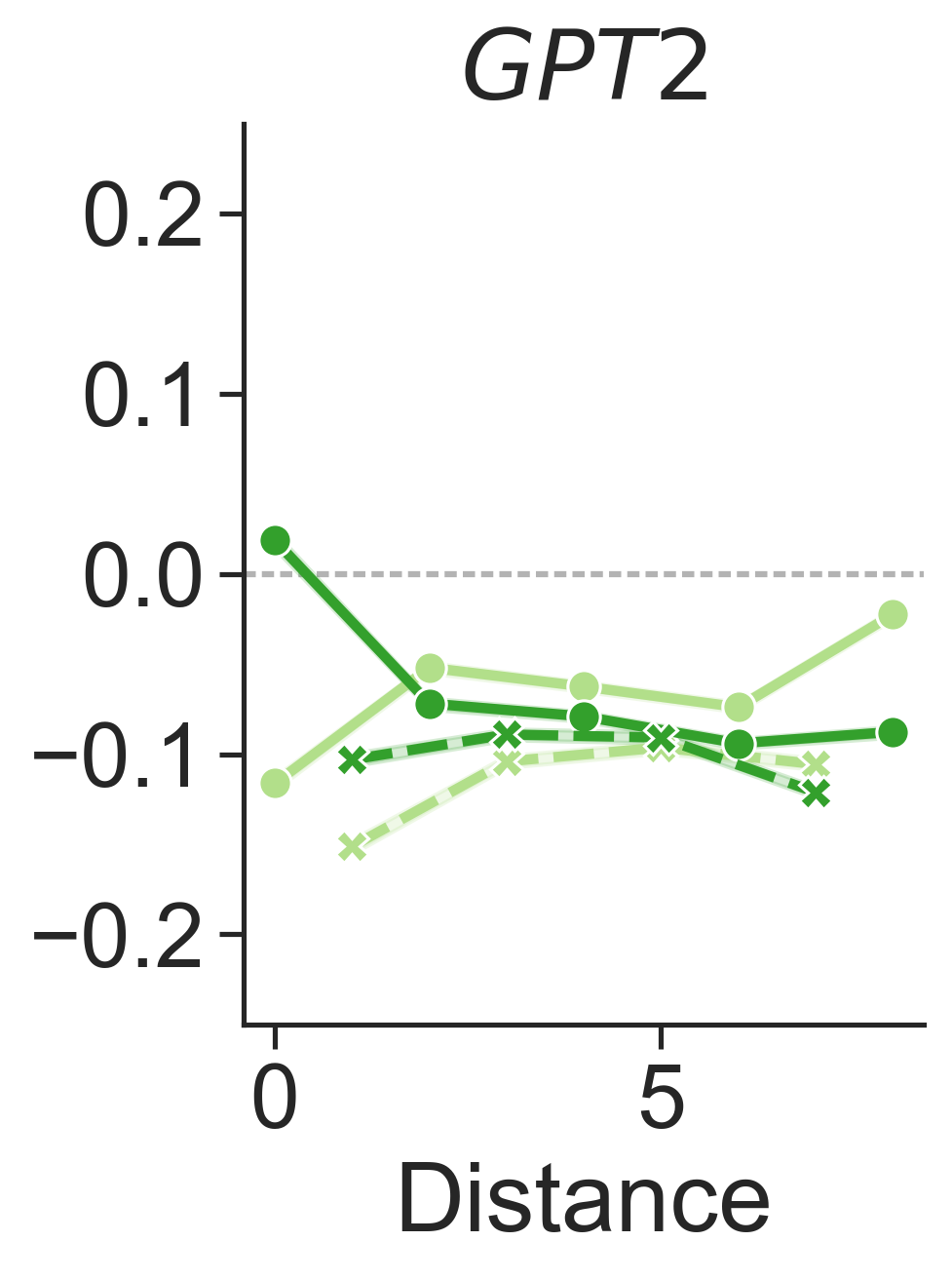}
        \includegraphics[height=2.59cm]{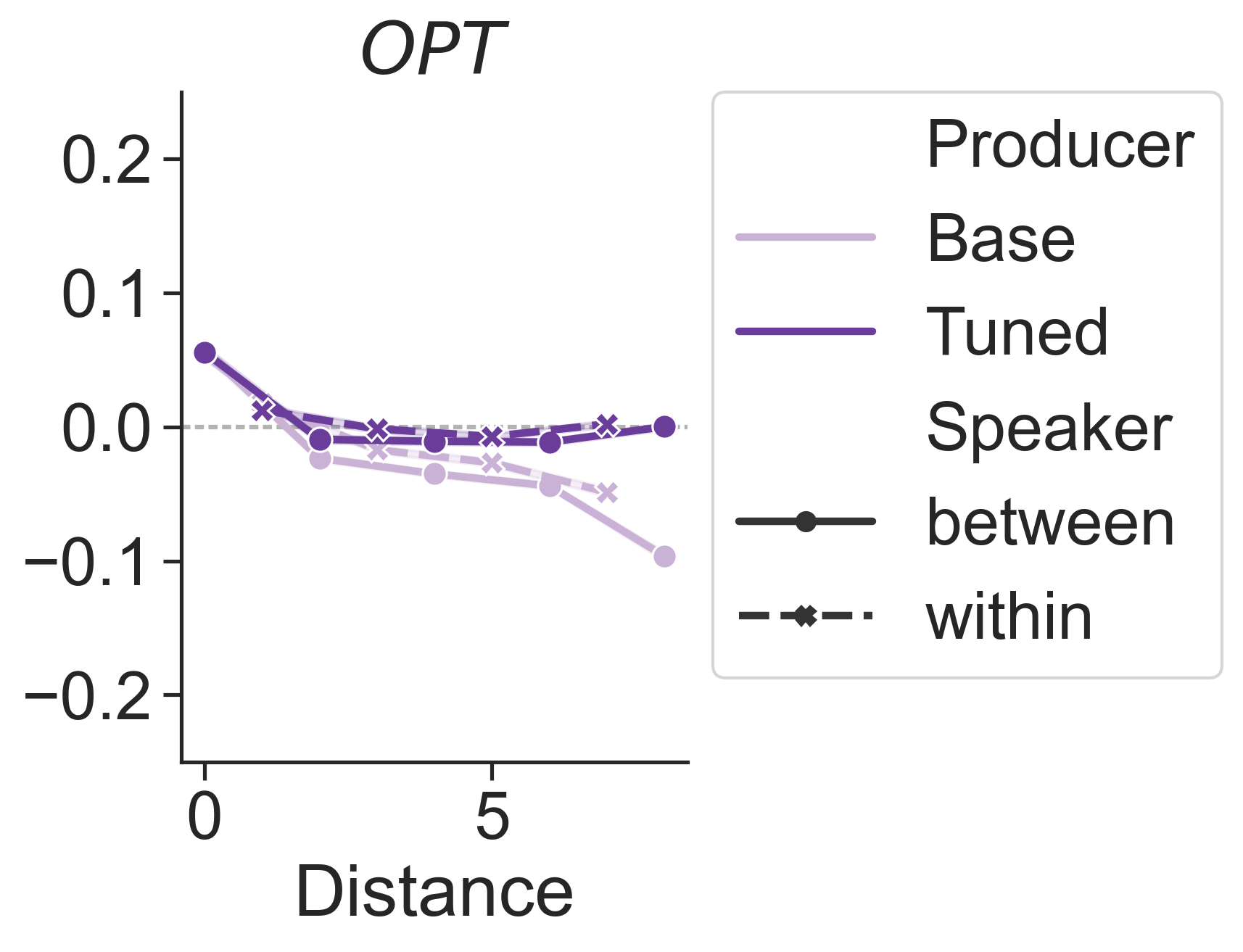}
    \caption{
        Relative attribution properties to human utterances over the dialogue context.
    }
    \label{fig:attribs_dist}
 \end{figure}

Humans and models display priming effects, which can be explained via accounts of residual activation, and they are sensitive to turn-taking~\citep{ten2005temporal,tooley2010syntactic,ekstedt2020turngpt,sinclair2022structural}. 
We thus expect attribution patterns 
to be sensitive to utterance position and speaker shifts within the context. Figure~\ref{fig:attribs_dist} shows how results change with fine-tuning.

\paragraph{\textit{Utterance comprehension is influenced by context locality in open domain dialogue}.}
When comprehending utterances from a given speaker, models fine-tuned on \sw\ learn to attribute more salience to utterances in the nearby context, more strongly so when these are produced by the other speaker. 
This effect is strongest for \gpt~($ \beta=-0.009$, $p<0.05$, $95\%\ CI = [-0.011\!:\!-0.007]$).
For \mt, we do not see such a clear trend, with different behaviours between models. 
Even though evidence for sensitivity to utterance position and speaker shifts in comprehension is only found in one of the two corpora, this is an interesting result when juxtaposed to our analysis of production behaviour. 
It seems to indicate that while models learn to \textit{understand} differences in speakers and in distance within the local context of open-domain dialogue, this does not always translate to human-likeness of \textit{production} behaviour.\looseness-1

 \begin{figure}[ht]
    \centering 
        \includegraphics[height=2.59cm]{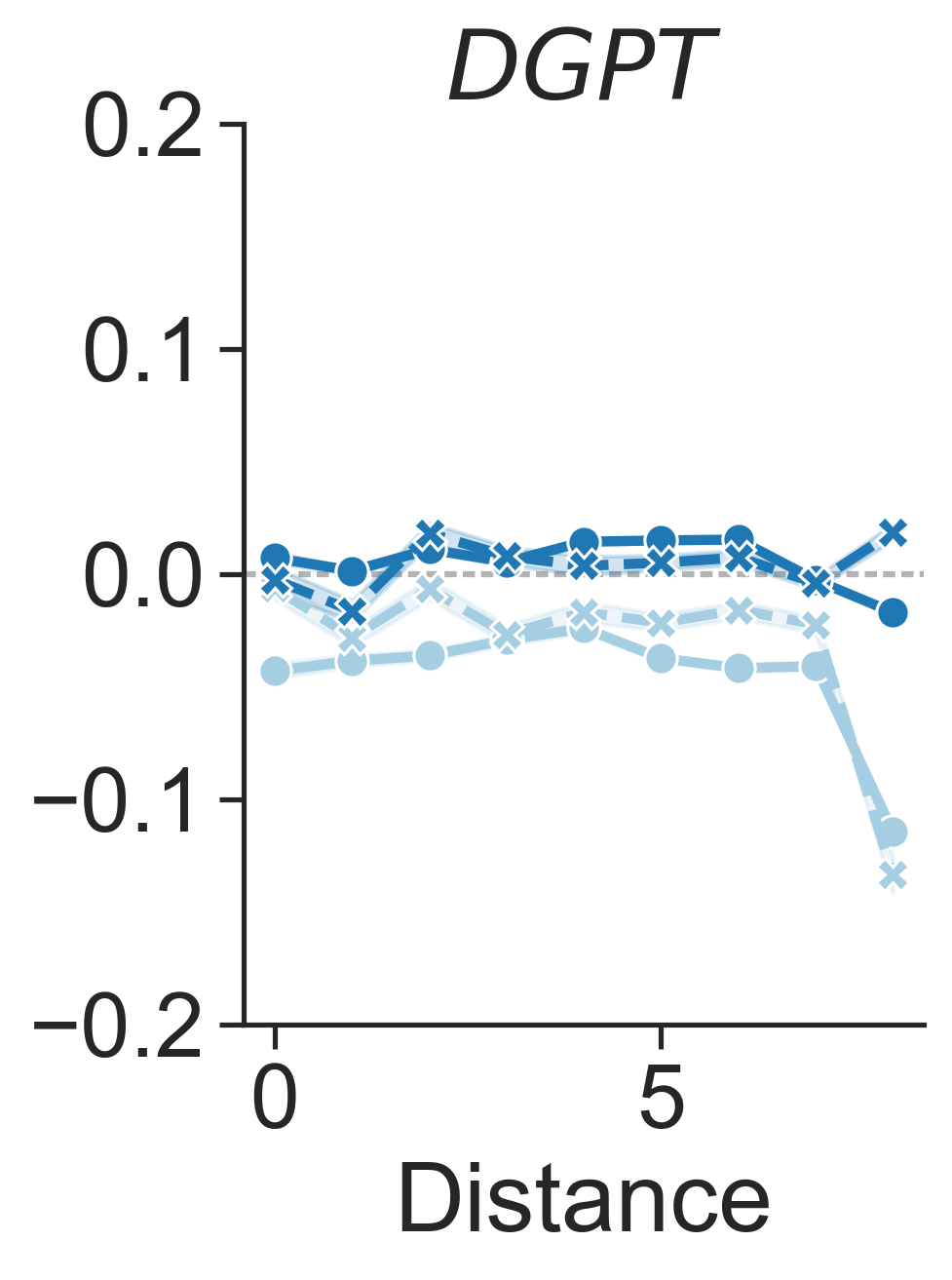}
        \includegraphics[height=2.59cm]{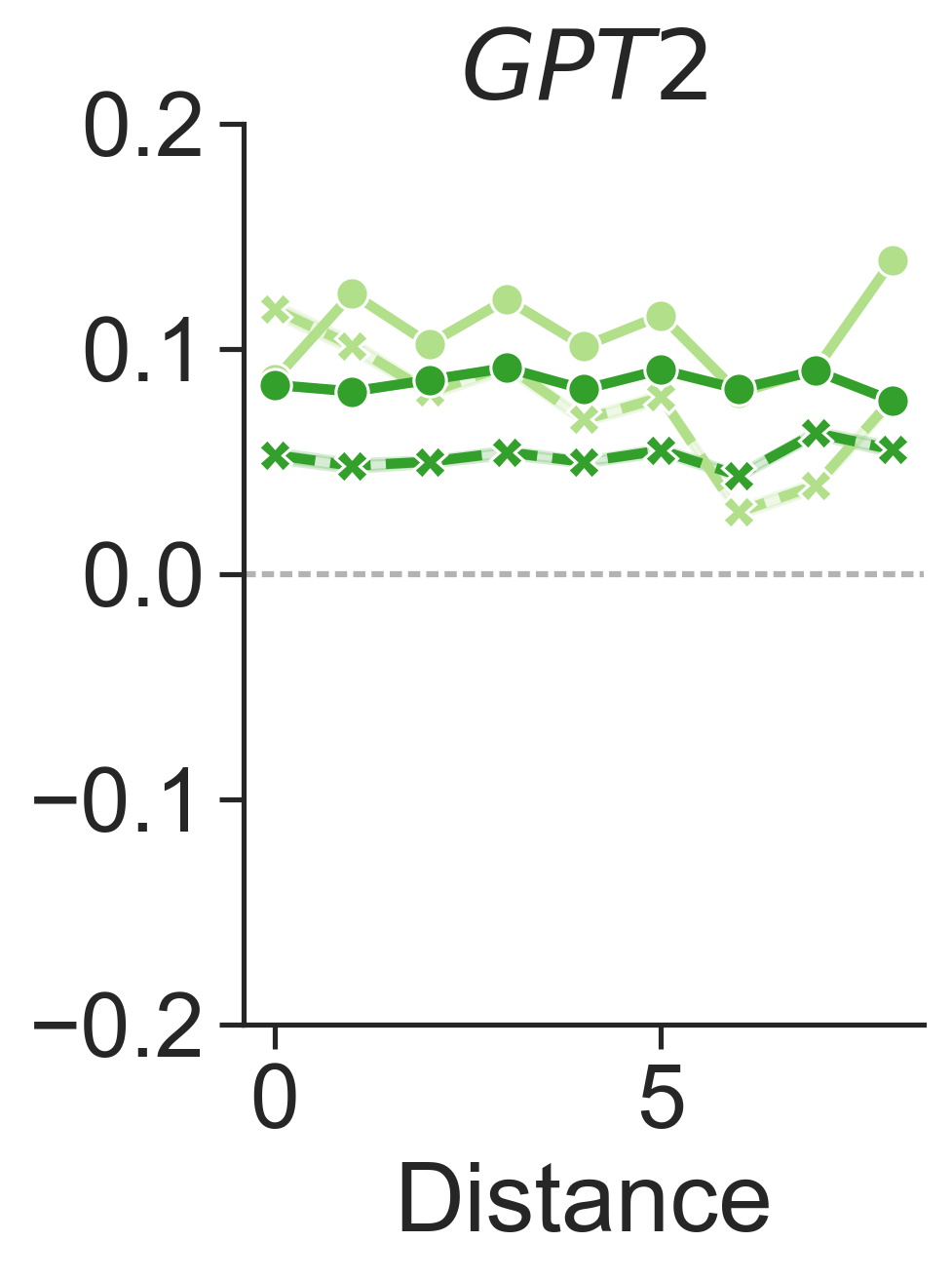}
        \includegraphics[height=2.59cm]{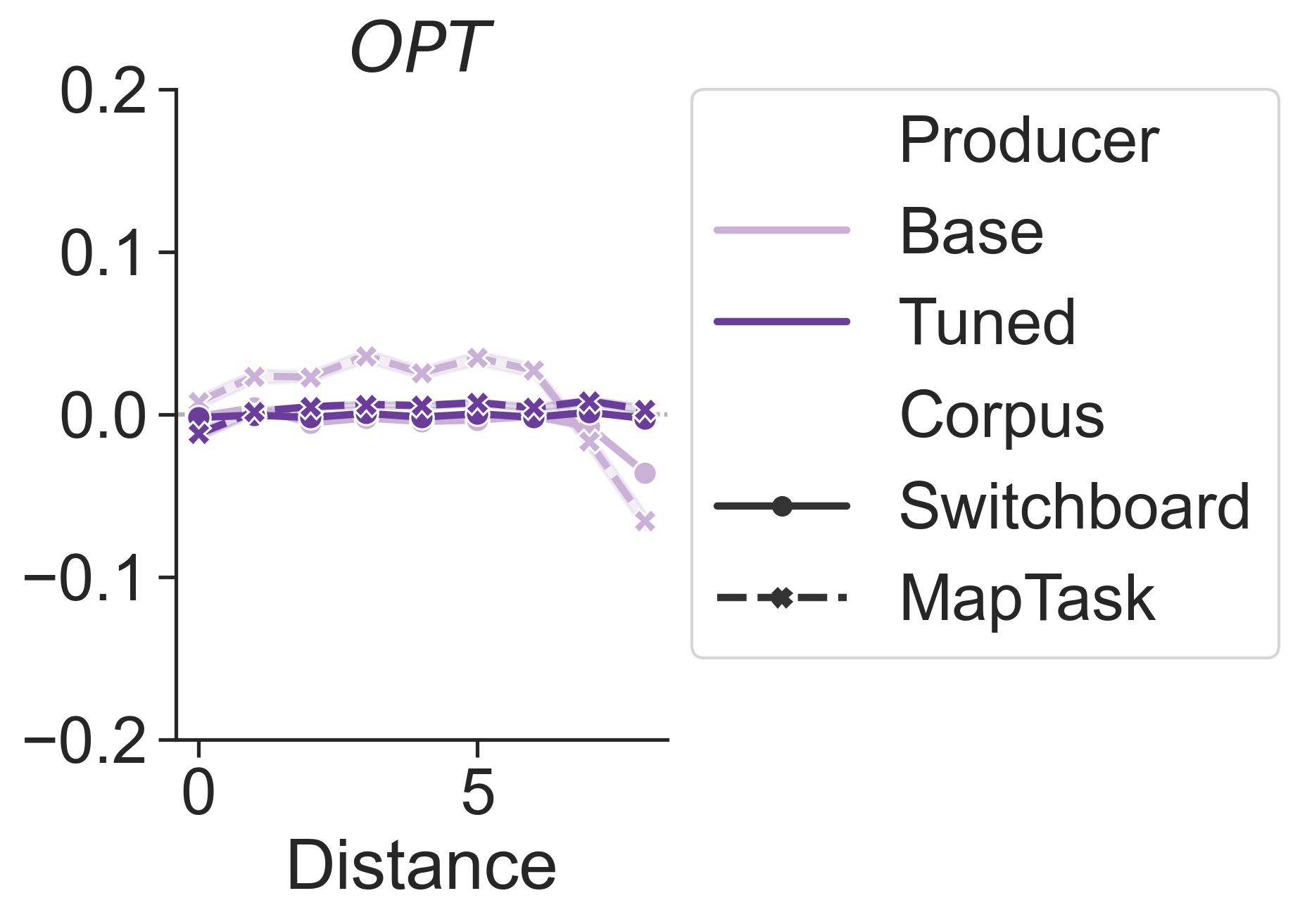}

    \caption{
        Relative attribution importance of speaker labels over the dialogue context.
    }
    \label{fig:s_attribs_dist}
 \end{figure}\looseness-1

\paragraph{\textit{Construction repetition in the local context predicts attribution patterns}.}

High lexical repetition between context and target has been shown to boost priming effects in models~\cite{sinclair2022structural}, however, less is known about how this translates to attribution patterns. In line with priming results, we expect that attribution patterns over context utterances will be predicted by both construction and vocabulary overlap. We see mixed results across models, finding that only for \sw, \gpt~displays significant positive effect of \co\ ~($ \beta=0.277$, $p<0.05$, $95\%\ CI = [0.239\!:\!0.315]$) on attribution strength, independent of \vo\ and distance effects.
Surprisingly, however, the effect of \vo\ on attribution strength is negative ($ \beta=-0.308$, $p<0.05$, $95\%\ CI = [-0.346\!:\!-0.270]$).
More remains to be done to precisely understand the relationship between the repetitions themselves and the local attribution patterns we observe, as well as to identify other factors driving this behaviour.

\subsubsection{Attribution Over Special Tokens}

While we are most interested in models' comprehension behaviour with respect to the utterance text in the context, we also investigate their behaviour over speaker labels.
The effect of structural tokens on the performance and behaviour of LMs is an ongoing area of research~\citep{wolf2019transfertransfo,Gu2020speakerawarebert,ekstedt2020turngpt,wallbridge2023dialogue}.
Speaker labels like \textit{`A:'} and  \textit{`B:'} provide models with important information about the turn-taking dynamics of dialogues. 
Figure~\ref{fig:s_attribs_dist} shows that models learn, through fine-tuning, to attribute salience to speaker labels in a more \textit{uniform} manner (note how the curves of tuned models are flatter).
We find significant differences between base and tuned models in both corpora, with the highest boost in uniformity for \dgpt\ (\sw: $ \beta=0.002$, $p<0.05$, $95\%\ CI = [0.002\!:\!0.002]$, \mt: $ \beta=0.005$, $p<0.05$, $95\%\ CI = [0.005\!:\!0.005]$).\footnote{Full breakdown of results in Appendix~\ref{app:attribution_model}.
} 
Speculatively, this could be taken as an indication that the models have learned to more consistently use these as structural markers of turn-taking. The discrepancy between the uniform attribution patterns over speaker labels and the decaying salience assigned to utterance text is an interesting finding that deserves more attention in future research.

%% file: sections/discussion_conclusion.tex
\section{Discussion \& Conclusion}
Repetition behaviour in dialogue, whether driven by local priming~\cite{bock1986syntactic}, alignment effects~\cite{pickering2004toward}, conceptual pacts~\cite{brennan1996conceptual}, or routinisation~\cite{pickering2005establishing,garrod2007alignment}, is well attested in humans. 
In this study, we investigate the extent to which language models are sensitive to, and display the same \textit{local}, \textit{context-specific}, and \textit{shared} patterns of construction repetition observed in human dialogue.
We conduct an in-depth analysis using two corpora of English task-oriented and open-domain dialogue, and three autoregressive neural language models.

Analysing human interactions, we find that within highly local contexts (we consider dialogue samples consisting of 10 utterances), repetition effects decay with distance from antecedents, particularly when repetitions are between dialogue partners, rather than of a speaker's own language. This contrasts with and complements previous work finding no evidence of locality effects within Switchboard, the same open domain corpus, when considering dialogues as a whole rather than in short excerpts~\cite{sinclair2021construction}, suggesting that some repeated constructions may occur in multiple short bursts~\cite{pierrehumbert2012burstiness} over the course of a dialogue---a phenomenon that is not easily captured by more `global' analyses.

We then evaluate model behaviour under two lenses: \textit{production} behaviour, analysed in terms of the repetition of shared constructions (i.e., word sequences re-used by both dialogue participants) in model generations, and \textit{comprehension} behaviour, measured by models' attribution of salience to contextual units when processing human-produced dialogue.
We find that models learn, via fine-tuning, to generate more human-like patterns of construction re-use, although the degree to which repetitions are local, context-specific, and shared varies by model. 
We also find that while reference-based generation quality metrics correlate with the human-likeness of the repetitions produced, corpus-level metrics like MAUVE fail to capture this important aspect of dialogue quality. This  highlights the need for more refined corpus-level approaches to statistical evaluation which take into account local and highly contextually dependent phenomena, or at least for their integration with instance-level analyses~\citep{deng2022model,giulianelli-etal-2023-what}.
Making use of feature attribution techniques, which provide interpretations of models' comprehension behaviour, we then explore the extent to which models are sensitive to properties of the context thought to influence human propensity to produce \textit{aligned} (i.e., locally repeated and context-specific) language.
We observe that when comprehending utterances, tuned models assign salience to speaker labels in a more uniform manner, 
and that in open-domain dialogue, models learn to assign salience over the context in a more local manner.

We will follow up this study with experiments
where our proposed attribution aggregation procedure is performed specifically over construction tokens in the target utterance. This may allow for more fine-grained interpretation of the relationship between repetitions and the observed local effects, as well as to investigate further psycholinguistic factors which may drive the tight coupling of local context and next utterance generation.
We hope our experimental setup will inspire future work that attempts to create stronger connections between language model behaviour and findings from psycholinguistics. In particular, we look forward to seeing our attribution-based methodology being applied to other dialogue-specific phenomena, 
and the local, dyad-specific repetition measures we investigate applied to the development and evaluation of more adaptive and context-sensitive dialogue response generation systems.

%% file: sections/appendix_details.tex
\section{Contributions}
Conceptualisation: AS. Methodology: AS, JJ. Software: AM. Experiments: AM, AS. Analysis: AM, AS, MG, JJ. Writing - Original Draft: AM, AS. Writing - Review \& Editing: AS, JJ, MG. Supervision \& Project Administration: AS. Order alphabetical. 

\section{Language Model Fine-Tuning}
\label{app_lm_finetune}

We fine-tune GPT-2 \cite{Radford2019LanguageMA}, OPT~\cite{zhang2022opt}, and  DialoGPT \cite{zhang-etal-2020-dialogpt} for 20 epochs, using an early stopping technique to save the best performing model (based on its perplexity). Table~\ref{tab:model_metrics} shows the perplexity of all models, pre-trained and fine-tuned, on the evaluation set. 

Models significantly adapt to the domain in training, given the low fine-tuned perplexities.

\begin{table}[ht]\centering \small \scalebox{0.73} { 
\setlength{\tabcolsep}{1.9pt}
\begin{tabular}{@{}ll r rrr rrr r r@{}} \toprule
& & PPL $\downarrow$ & Prec & Rec & F1 & BLEU & BP $\downarrow$ & LR $\downarrow$ & Mve & L$\pm$Std \\ \midrule
\textit{SW} & & & & & & & & & & \\
GPT2 & B & 15.110 & 0.722 & 0.704 & 0.710 & 0.009 & 0.744 & 0.772 & 0.035 & 11.9 $\pm$ 14.7 \\ 
& T & 12.020 & \textbf{0.745} & 0.720 & 0.730 & 0.010 & \textbf{0.496} & \textbf{0.588} & 0.049 & 8.8 $\pm$ 10.5 \\
OPT & B & 37.540 & 0.703 & 0.702 & 0.700 & 0.010 & 0.859 & 0.868 & 0.052 & 13.0 $\pm$ 13.8 \\ 
& T & 15.130 & 0.737 & \textbf{0.733} & \textbf{0.733} & 0.014 & 0.824 & 0.838 & \textbf{0.069} & 12.6 $\pm$ 12.9 \\
DGPT & B & 6935.000 & 0.667 & 0.648 & 0.656 & 0.000 & 0.148 & 0.343 & 0.006 & 3.3 $\pm$ 3.5 \\ 
& T & \textbf{10.910} & 0.737 & 0.728 & 0.730 & \textbf{0.016} & 0.955 & 0.956 & 0.049 & 14.3 $\pm$ 15.8 \\
\textit{MT} & & & & & & & & & & \\
GPT2 & B & 16.170 & 0.681 & 0.680 & 0.679 & 0.006 & 0.827 & 0.841 & 0.101 & 7.1 $\pm$ 6.2 \\ 
& T & \textbf{7.930} & 0.705 & 0.702 & 0.702 & 0.014 & 0.849 & 0.859 & 0.245 & 7.4 $\pm$ 6.1 \\
OPT & B  & 72.100 & 0.686 & 0.681 & 0.682 & 0.006 & 0.701 & 0.738 & 0.103 & 6.1 $\pm$ 6.4 \\  
& T & 9.700 & \textbf{0.723} & \textbf{0.705} & \textbf{0.712} & 0.016 & \textbf{0.631} & \textbf{0.685} & \textbf{0.339} & 5.7 $\pm$ 5.2 \\
DGPT & B  & 13014.000 & 0.668 & 0.659 & 0.662 & 0.002 & 0.391 & 0.516 & 0.041 & 3.7 $\pm$ 2.8 \\ 
& T & 8.050 & 0.701 & 0.700 & 0.699 & \textbf{0.016} & 0.990 & 0.990 & 0.176 & 8.5 $\pm$ 7.9 \\
\bottomrule
\end{tabular}}
\caption{Post-training metrics of models. \textit{SW}: Switchboard. \textit{MT}: Map Task. Precision (\textit{Prec}), recall (\textit{Rec}) and \textit{F1} are averages over multiple samples and part of BERTScore. \textit{LR}: length ratio (BLEU). \textit{BP}: brevity penalty (BLEU). \textit{PPL}: Perplexity. \textit{B}: base models. \textit{T}: tuned models. \textit{Mve}: MAUVE score. \textit{L}: mean target utterance length (in words). \textbf{Bold} indicates best values across models per corpora per metric.}
\label{tab:model_metrics} 
\end{table}

\section{Language Model Sizes}\label{app:model_sizes}
The considered language models have the following number of parameters. \gpt: 124M, \opt: 125M, \dgpt: 117M, \textsc{Pythia}: 1.4B.

\section{Filled Pauses}
\label{app:filled_pauses}
We define filled pauses using the part-of-speech tags in Map Task and Switchboard. 
\textbf{Map Task:} \textit{uh-huh, er, um, mm-mm, eh, uh, mm, uh-uh, nah, mm-hmm, erm, ehm, huh, hmm, mmhmm}.
\textbf{Switchboard:} \textit{hm, huh, uh, um-hum, huh, huh-uh, uh, uh-huh, um}.

\section{Construction Repetitions} \label{app:repetitionstats}

\subsection{Construction Examples}
\label{app:qual_examples}
Table~\ref{tab:constructions_excerpt} contains two dialogue excerpts with responses generated by a tuned \opt~model. Phrases highlighted bold refer to constructions generated by the model.

Table~\ref{tab:example_constructions} lists the most frequent constructions generated by fine-tuned models, grouped by locality. Local and global constructions are defined as having a repetition distance of $\leq 4$ and $> 4$, respectively. The table contains the top three most frequent produced constructions per model, per dataset, per locality. 

\begin{table}[ht!]
\small
\centering
    \begin{tabular}{p{0.4\textwidth}}
    \toprule
    \textit{Selected local dialogue sample excerpts} \\ \midrule
        A: but . that is a very good point. i am from \textbf{west virginia} so i understand what you are saying there\\
        B: uh-huh.\\
        A: and it's it's a very good point.\\
        B: (\textit{OPT}) uh-huh. yeah, uh, where \textbf{i was}, uh, i went to a church in \textbf{west virginia} and uh, they always wore suits (...) \\ \\

        B: and \textbf{there's a lot of} graft, like people trying to tell
   them, oh, giving them information that was free to them
   anyway if they just knew how to get it.\\
        A: uh, yeah, exactly.\\
        B: yeah.\\
        A: (\textit{OPT}) yeah, \textbf{there's a lot of} corruption,\\ \\

        B: right, so that's that's right, so i'm coming back back
           down \textbf{the paper} again \\
        A: ah, yeah back down \textbf{the paper} \\
        B: uh-huh aye turn right \\
        A: (\textit{DGPT}) and then you go through \textbf{the paper} and past the chapel, to the right of the page \\ \\

        A: \textbf{okay right}, you went down past burnt forest \\
        B: i went underneath burnt forest \\
        A: well, you weren't meant to \\
        B: well you said draw round the cottage \\
        A: \textbf{okay right}, you're meant to come down from the start \\
        B: (\textit{OPT}) \textbf{okay right} \\ \\

        A: oh, yeah, yeah, yeah. \\
        B: \textbf{in the summer} or like in the easter time, like around now? \\
        A: (\textit{HUMAN}) no, usually \textbf{in the summer} time. \\
 
    \bottomrule
    \end{tabular}
\caption{Example local repetitions produced by tuned models. }
\label{tab:constructions_excerpt}
\end{table}\looseness-1

\begin{table*}[ht]
\small
\centering
\scalebox{0.9}{

    \begin{tabular}{l l| l|l|l | l }
    \toprule
     &  \textit{distance}  & \textbf{\textit{Human}}  &  \textbf{\textit{GPT2}}   &  \textbf{\textit{OPT}} &  \textbf{\textit{DGPT}} \\ \midrule
    
    \multirow{6}{*}{MT}  
                         &  \multirow{3}{*}{local}  &  the diamond mine & the trout farm & the diamond mine & the abandoned cottage\\ 
                         &  & the concealed hideout & the diamond mine & the fallen pillars & have you got\\ 
                         &  & the rope bridge & to the left & to the left & the rift valley\\ 
                         
                         & \multirow{3}{*}{global} & the pine forest & of the concealed hideout & edge of the map & outside of the\\ 
                         &  & don't have a & and a half & don't have a graveyard & a saloon bar\\ 
                         &  & the outlaws' hideout & two inches below where & of the walled city & up the map\\ 
    \midrule
    \multirow{6}{*}{SW}  &  
                            \multirow{3}{*}{local} & a lot of & a lot of & a lot of & a lot of\\ 
                         &  & i don't know & i don't know & i don't know & 	i don't know\\ 
                         &  & the peace corps & freedom of speech & one of the & the peace corps\\ 
                        
                         &  \multirow{3}{*}{global} & i used to & it was just & do you think & you're supposed to\\ 
                         &  & would be a & paying sales tax & i think it & i don't know if\\
                         &  & going to be & some of them & because i was & and a lot\\ 
    \bottomrule
    \end{tabular}
}
\caption{Example constructions from tuned models. \textit{MT}: Map Task, \textit{SW}: Switchboard. \textit{Local}: repetition distance $\leq 4$; \textit{global}: repetition distance $> 4$. 
}
\label{tab:example_constructions}
\end{table*}
\looseness-1

\subsection{Repetition Properties}
Tables~\ref{tab:repstat_statsig_sw} and~\ref{tab:repstat_statsig_mt} contain detailed repetition statistics with statistical significance test results.
In both corpora, \dgpt~learns to best approximate human target lengths after fine-tuning (\textit{TH} columns of all models: $-15$, $-92.8$, and $-38.59$ ($t$) for \dgpt, \gpt, and \opt, respectively. $p<0.05$ for all). It generates significantly longer responses ($t=-412.64$, $p<0.05$).
Models robustly generate more dialogue-specific shared constructions after fine-tuned on Switchboard ($t$: $-109.41$, $57.44$, $-19.15$, $p<0.05$). After fine-tuned on Map Task, models learn to generate less dialogue-specific constructions ($t$: $19.83$, $27.43$, $22.85$, $p<0.05$).
Models learn to produce more \textit{distant} shared constructions after trained on both open-ended and task-oriented dialogue data (\textit{SW}: $t$: $-4.34$, $-10.2$, $-20.6$, \textit{MT}: $t$: $-10.76$, $-0.19$ ($p\geq0.05$, exception), $-8.53$, $p<0.05$).
\dgpt~exhibits higher levels of construction overlap ($CO$) after fine-tuned on both Switchboard and Map Task (both between and within speakers), closely approximating human patterns (\textit{SW}: $t$: $-23.09$, $-11.45$, \textit{MT}: $t$: $-29.75$, $-14.75$, $p<0.05$). \gpt~and \opt~generally learn to produce lower $CO$ values, but they already exhibit highly human-like construction overlap scores in their pre-trained states (\textit{SW}: $t$: $6.83$, $2.68$, $16.52$, $3.18$, $p<0.05$, \textit{MT}: $t$: $-1.62$, $-1.4$, $0.75$, $1.05$, $p\geq0.05$).

\begin{table*}[ht]

\centering \small \scalebox{0.75} { 
\setlength{\tabcolsep}{3.5pt}
\begin{tabular}{@{}l r|rrrrr|rrrrr|rrrrr@{}} \toprule
& \textbf{H} & \multicolumn{5}{c}{\textbf{DGPT}} & \multicolumn{5}{c}{\textbf{GPT2}} & \multicolumn{5}{c}{\textbf{OPT}} \\
& & B & T & BH & TH & BT & B & T & BH & TH & BT & B & T & BH & TH & BT \\ \midrule
\textit{SW} & & & & & & & & & & & & & & & & \\
\hspace{2mm} target len. & 15.369 & 3.251 & 14.271 & -174.840 & -15.000 & -412.640 & 11.925 & 8.802 & -47.420 & -92.800 & 108.160 & 13.026 & 12.599 & -32.460 & -38.590 & 14.090 \\
\hspace{2mm} constr. len. & 2.176 & 2.117 & 2.185 & -30.660 & 5.200 & -55.900 & 2.196 & 2.186 & 11.070 & 5.750 & 9.400 & 2.239 & 2.215 & 33.810 & 21.410 & 19.790 \\
\hspace{2mm} PMI & 8.520 & 8.053 & 8.821 & -42.450 & 25.740 & -109.410 & 8.424 & 8.907 & -8.020 & 33.190 & -57.440 & 9.147 & 9.303 & 53.330 & 67.020 & -19.150 \\
\hspace{2mm} freq. & 2.689 & 2.607 & 2.662 & -21.530 & -7.460 & -22.690 & 2.778 & 2.672 & 24.660 & -4.600 & 49.790 & 2.677 & 2.648 & -3.230 & -11.610 & 14.530 \\
\hspace{2mm} rep. dist. & 3.525 & 3.363 & 3.891 & \textcolor{red}{-1.220} & 5.840 & -4.340 & 3.586 & 3.990 & \textcolor{red}{0.980} & 7.040 & -10.200 & 3.104 & 3.774 & -6.870 & 3.950 & -20.600 \\
\hspace{2mm} CO & & & & & & & & & & & & & & & & \\ 
\hspace{4mm} between & 0.006 & 0.002 & 0.006 & -16.910 & \textcolor{red}{-1.270} & -23.090 & 0.008 & 0.005 & 6.830 & -2.520 & 16.070 & 0.011 & 0.007 & 16.520 & 4.340 & 23.460 \\
\hspace{4mm} within & 0.001 & 0.000 & 0.001 & -9.860 & -2.060 & -11.450 & 0.002 & 0.001 & 2.680 & \textcolor{red}{-0.180} & 4.600 & 0.002 & 0.001 & 3.180 & \textcolor{red}{-0.400} & 6.340 \\
\hspace{2mm} VO & & & & & & & & & & & & & & & & \\ 
\hspace{4mm} between & 0.116 & 0.107 & 0.122 & -6.350 & 5.340 & -15.770 & 0.132 & 0.125 & 12.700 & 7.920 & 8.530 & 0.137 & 0.126 & 18.620 & 8.920 & 17.100 \\
\hspace{4mm} within & 0.161 & 0.106 & 0.149 & -34.490 & -7.960 & -38.130 & 0.172 & 0.170 & 6.720 & 5.980 & \textcolor{red}{1.470} & 0.146 & 0.159 & -10.800 & \textcolor{red}{-1.190} & -16.190 \\
\bottomrule \end{tabular}}
\caption{\textbf{Switchboard repetition statistics} with statistical significance tests. \textcolor{red}{Red} values indicate statistical \textit{in}significance ($p\geq.05$). All values not highlighted red are statistically significant. The human (\textit{H}), base model (\textit{B}), and tuned model (\textit{T}) columns contain averages. The base model--human (\textit{BH}), tuned model--human (\textit{TH}), and base model--tuned model (\textit{BT}) comparison columns contain computed t-statistics. \textit{Rep. dist.}: repetition distance. \textit{Target len.}: target utterance length (in words). \textit{Constr. len.}: construction length (in words). \textit{Between/within}: between- and within-speaker. \textit{Freq.}: frequency.}
\label{tab:repstat_statsig_sw}
\end{table*}

\begin{table*}\centering \small \scalebox{0.75} { 
\setlength{\tabcolsep}{3.5pt}
\begin{tabular}{@{}l r|rrrrr|rrrrr|rrrrr@{}} \toprule
& \textbf{H} & \multicolumn{5}{c}{\textbf{DGPT}} & \multicolumn{5}{c}{\textbf{GPT2}} & \multicolumn{5}{c}{\textbf{OPT}} \\
& & B & T & BH & TH & BT & B & T & BH & TH & BT & B & T & BH & TH & BT \\ \midrule
\textit{MT} & & & & & & & & & & & & & & & & \\
\hspace{2mm} target len. & 8.607 & 3.701 & 8.488 & -75.490 & \textcolor{red}{-1.710} & -175.650 & 7.119 & 7.411 & -22.220 & -17.870 & -10.990 & 6.062 & 5.670 & -37.910 & -44.360 & 15.530 \\
\hspace{2mm} constr. len. & 2.373 & 2.272 & 2.240 & -20.790 & -28.610 & 11.740 & 2.321 & 2.287 & -11.000 & -18.390 & 13.830 & 2.427 & 2.403 & 11.210 & 6.270 & 8.260 \\
\hspace{2mm} PMI & 7.063 & 7.339 & 7.113 & 18.580 & 3.220 & 19.830 & 7.652 & 7.341 & 39.130 & 18.180 & 27.430 & 7.956 & 7.722 & 60.480 & 44.730 & 22.850 \\
\hspace{2mm} freq. & 3.249 & 2.980 & 2.999 & -35.100 & -32.780 & -4.180 & 3.214 & 3.180 & -4.590 & -9.000 & 7.310 & 3.230 & 3.105 & -2.470 & -19.060 & 29.250 \\
\hspace{2mm} rep. dist. & 3.281 & 2.736 & 3.554 & -5.830 & 3.950 & -10.760 & 3.439 & 3.447 & 2.270 & 2.390 & \textcolor{red}{-0.190} & 3.245 & 3.625 & \textcolor{red}{-0.530} & 4.840 & -8.520 \\
\hspace{2mm} CO & & & & & & & & & & & & & & & & \\ 
\hspace{4mm} between & 0.028 & 0.010 & 0.028 & -20.600 & \textcolor{red}{-0.480} & -29.750 & 0.027 & 0.026 & \textcolor{red}{-1.620} & \textcolor{red}{-1.860} & \textcolor{red}{0.320} & 0.029 & 0.024 & \textcolor{red}{0.750} & -3.890 & 7.820 \\
\hspace{4mm} within & 0.011 & 0.004 & 0.009 & -14.300 & -4.100 & -14.750 & 0.010 & 0.010 & \textcolor{red}{-1.400} & -2.380 & \textcolor{red}{1.370} & 0.012 & 0.009 & \textcolor{red}{1.050} & -3.650 & 7.540 \\
\hspace{2mm} VO & & & & & & & & & & & & & & & & \\ 
\hspace{4mm} between & 0.118 & 0.121 & 0.130 & \textcolor{red}{1.350} & 5.470 & -6.160 & 0.118 & 0.117 & \textcolor{red}{0.020} & \textcolor{red}{-0.340} & \textcolor{red}{0.660} & 0.139 & 0.137 & 8.570 & 7.260 & \textcolor{red}{1.480} \\
\hspace{4mm} within & 0.164 & 0.124 & 0.158 & -13.920 & -2.190 & -19.590 & 0.149 & 0.162 & -5.630 & \textcolor{red}{-0.380} & -8.910 & 0.157 & 0.180 & -2.370 & 5.050 & -12.890 \\
\bottomrule \end{tabular}}
\caption{\textbf{Map Task repetition statistics} with statistical significance tests. \textcolor{red}{Red} values indicate statistical \textit{in}significance ($p\geq.05$). All values not highlighted red are statistically significant. The human (\textit{H}), base model (\textit{B}), and tuned model (\textit{T}) columns contain averages. The base model--human (\textit{BH}), tuned model--human (\textit{TH}), and base model--tuned model (\textit{BT}) comparison columns contain computed t-statistics. \textit{Rep. dist.}: repetition distance. \textit{Target len.}: target utterance length (in words). \textit{Constr. len.}: construction length (in words). \textit{Between/within}: between- and within-speaker. \textit{Freq.}: frequency.}
\label{tab:repstat_statsig_mt}
\end{table*}

\section{Attributions To Target}
\label{app:attribtargetdetails}
\begin{figure}[ht]
    \centering 
    \includegraphics[height=1.9cm]{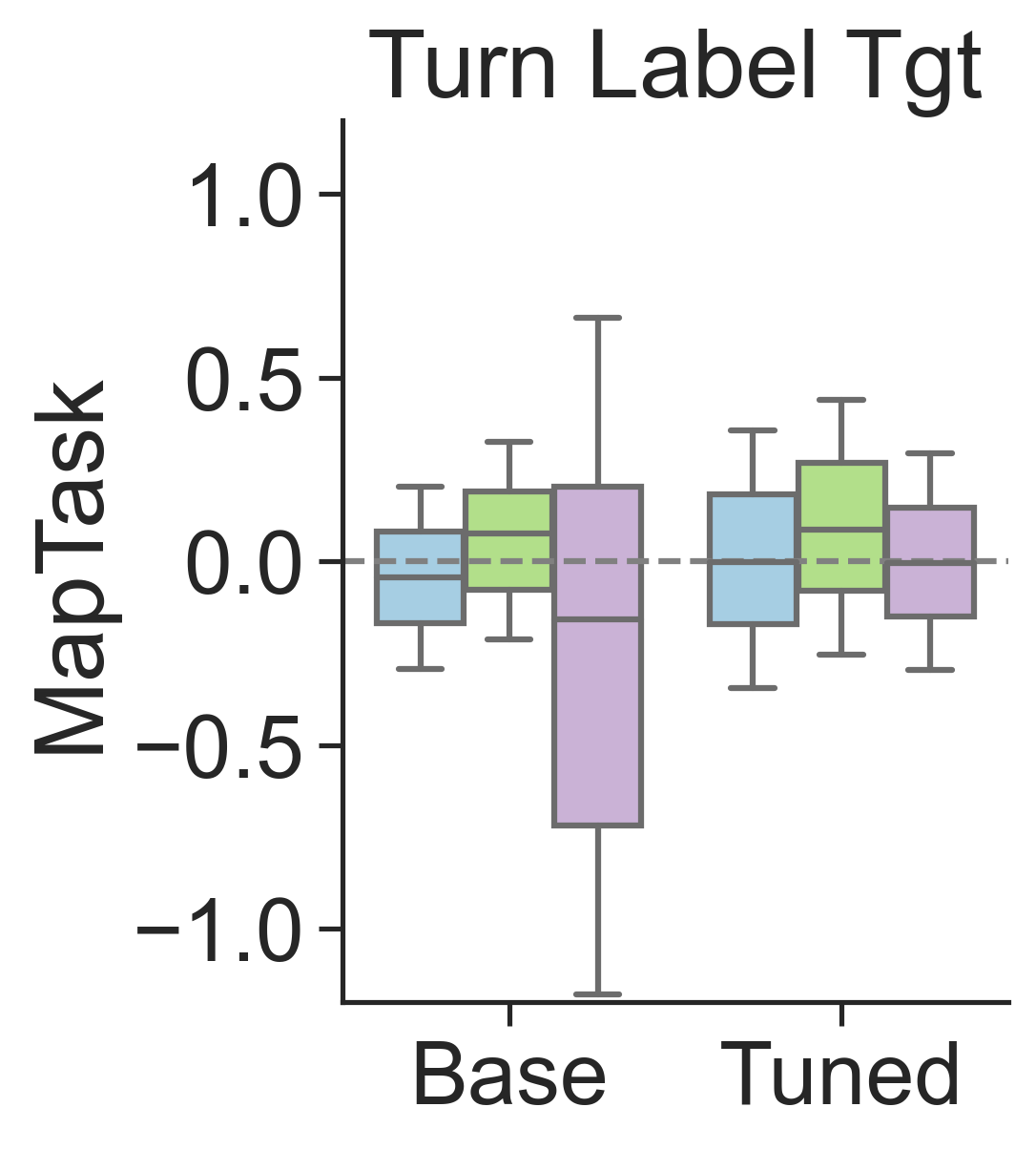}
    \includegraphics[height=1.9cm]{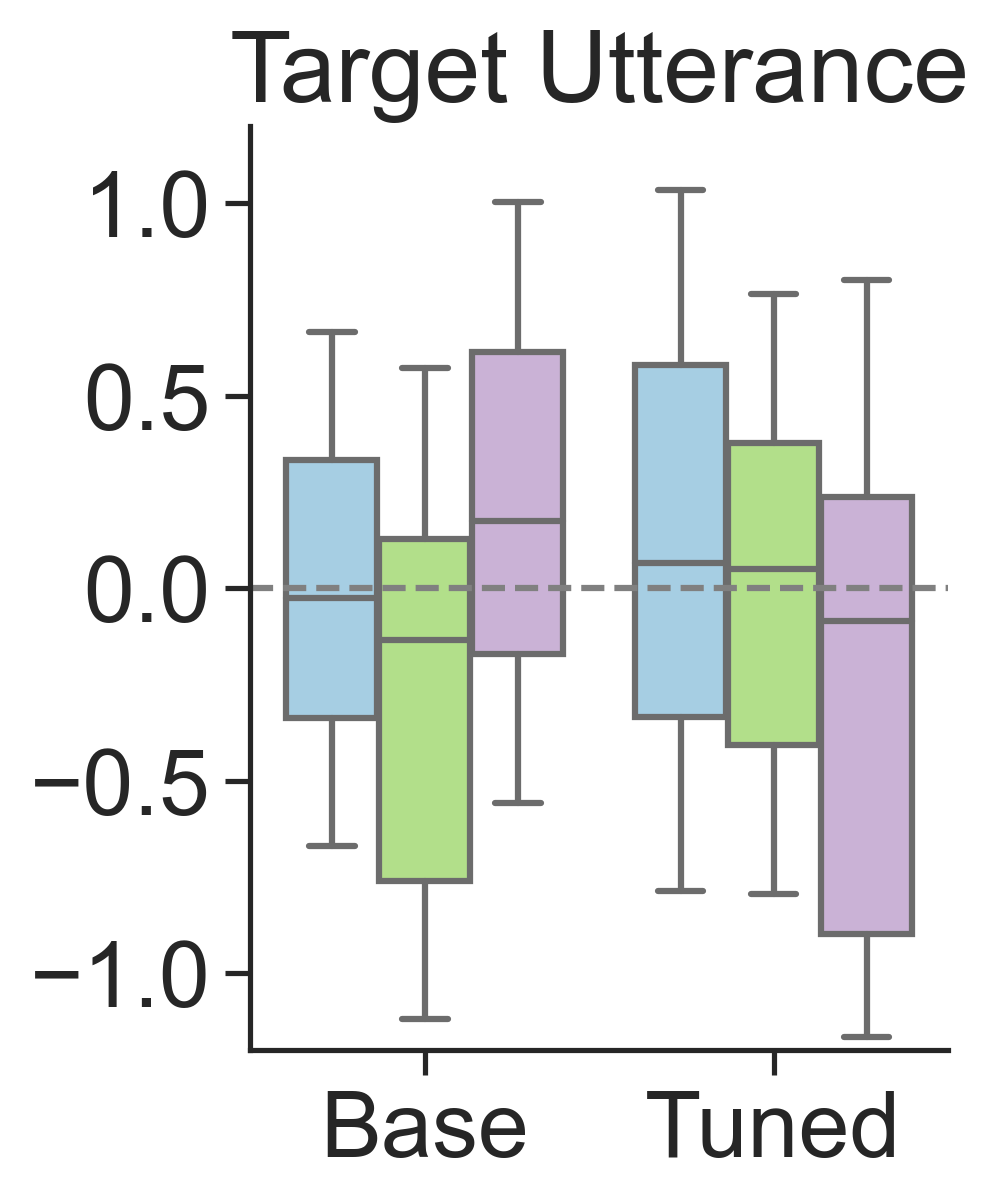}
    \includegraphics[height=1.9cm]{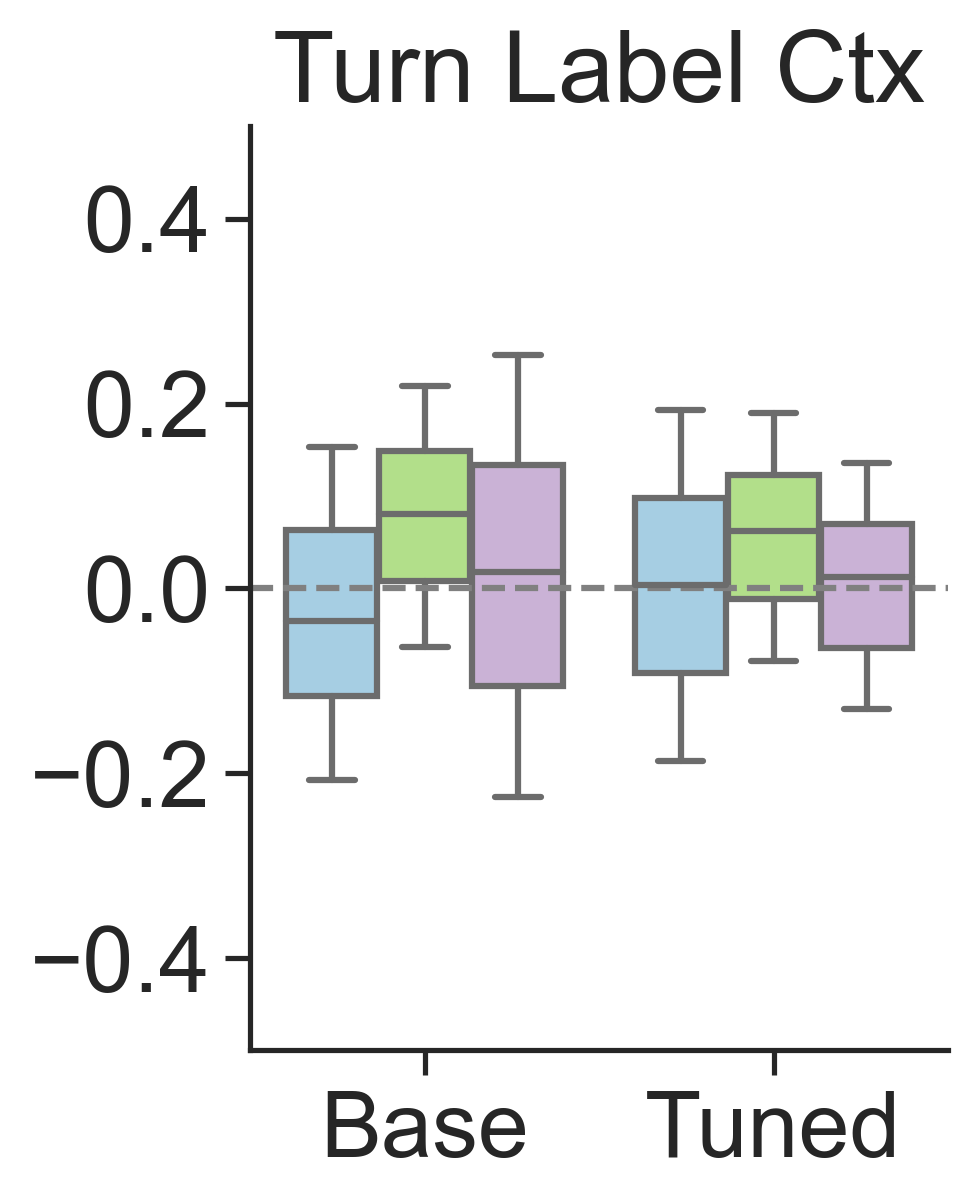}
    \includegraphics[height=1.9cm]{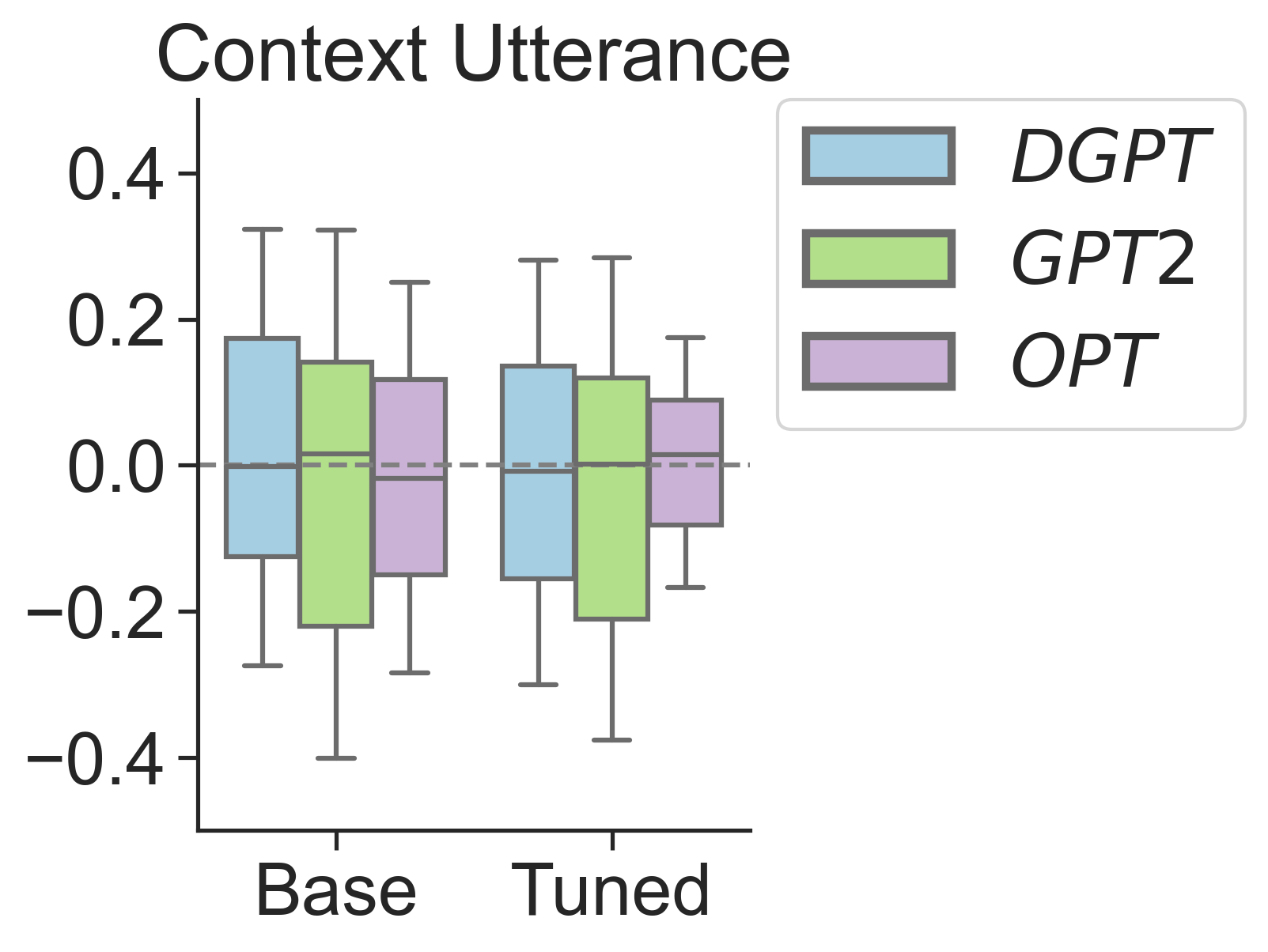}

    \includegraphics[height=1.9cm]{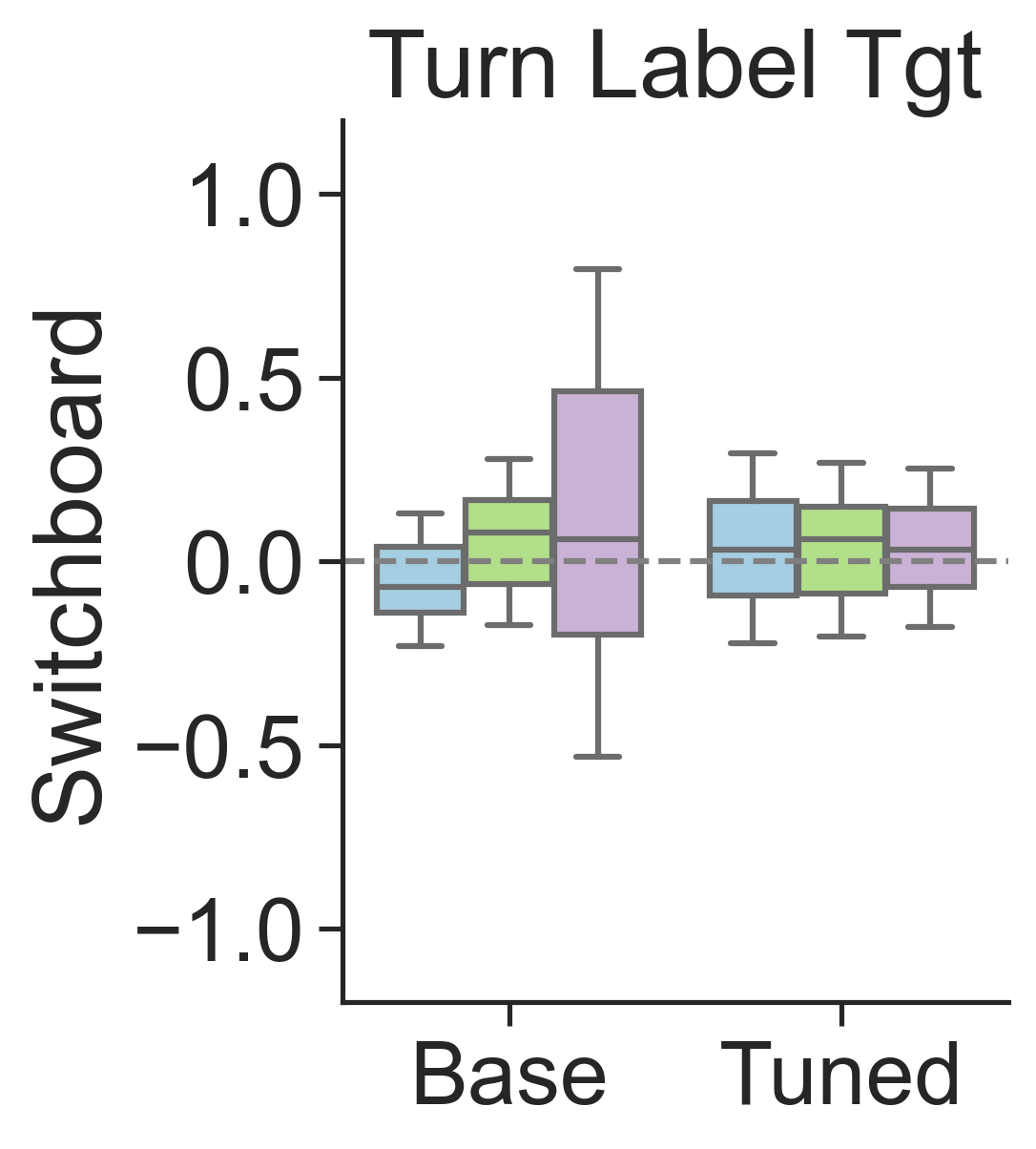}
    \includegraphics[height=1.9cm]{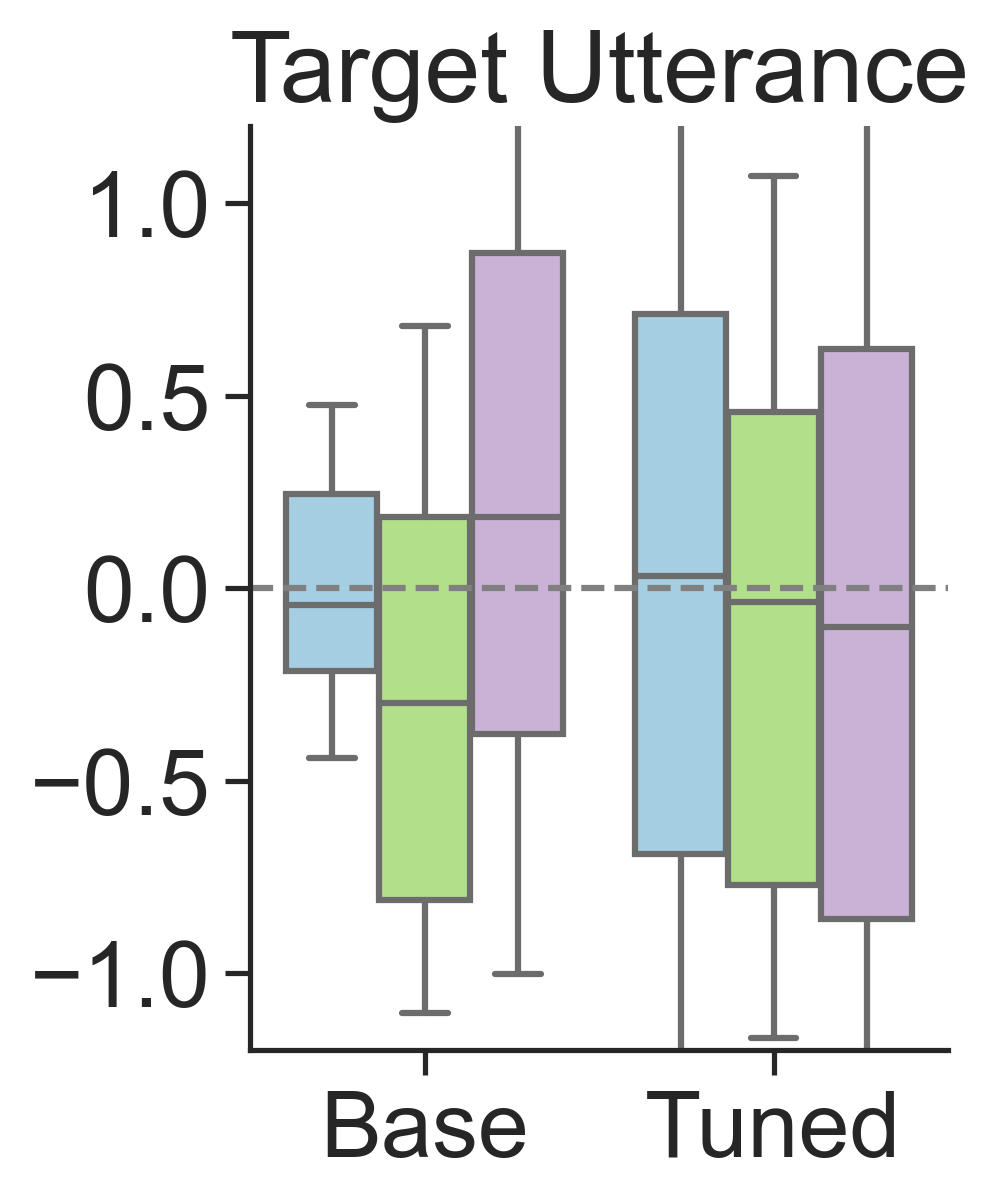}
    \includegraphics[height=1.9cm]{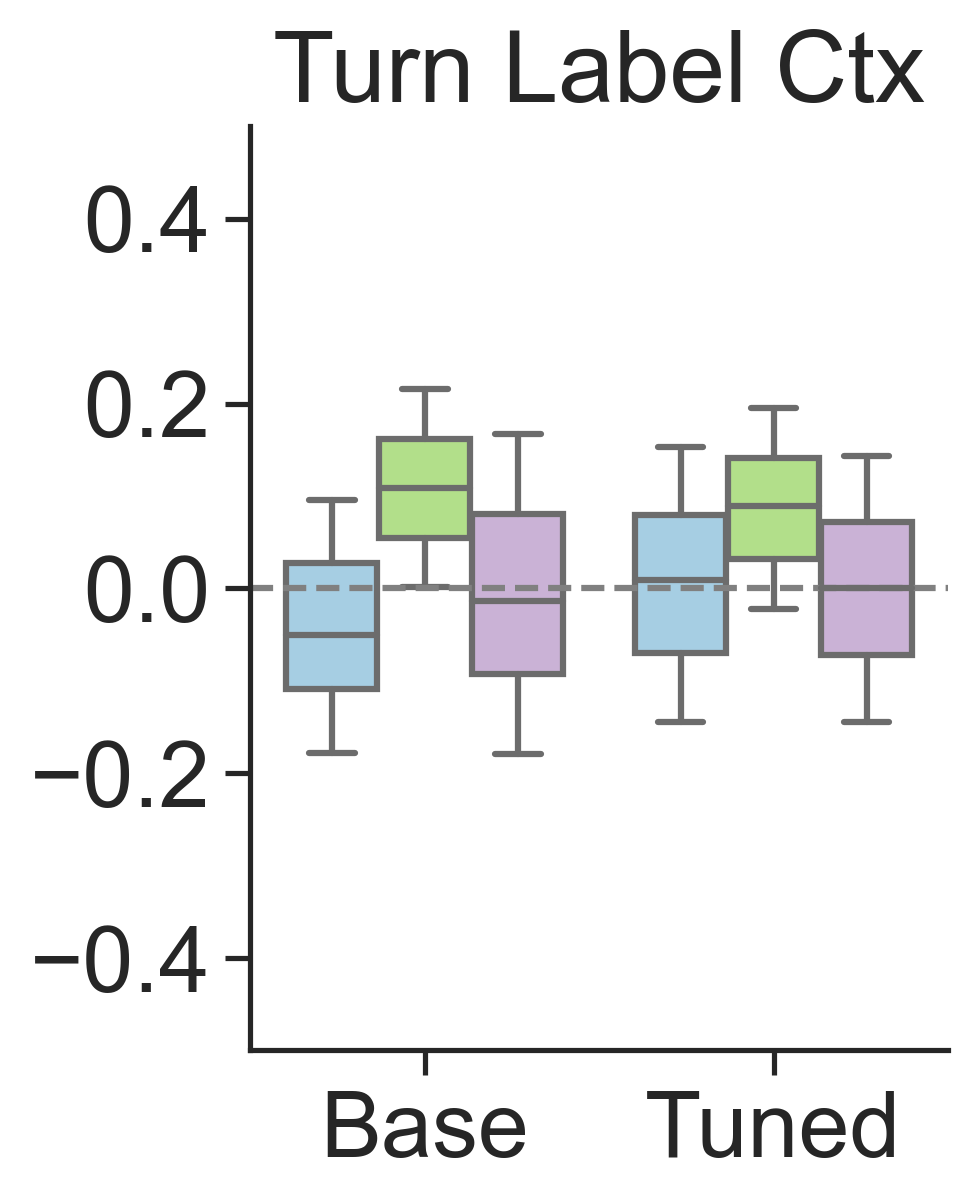}
    \includegraphics[height=1.9cm]{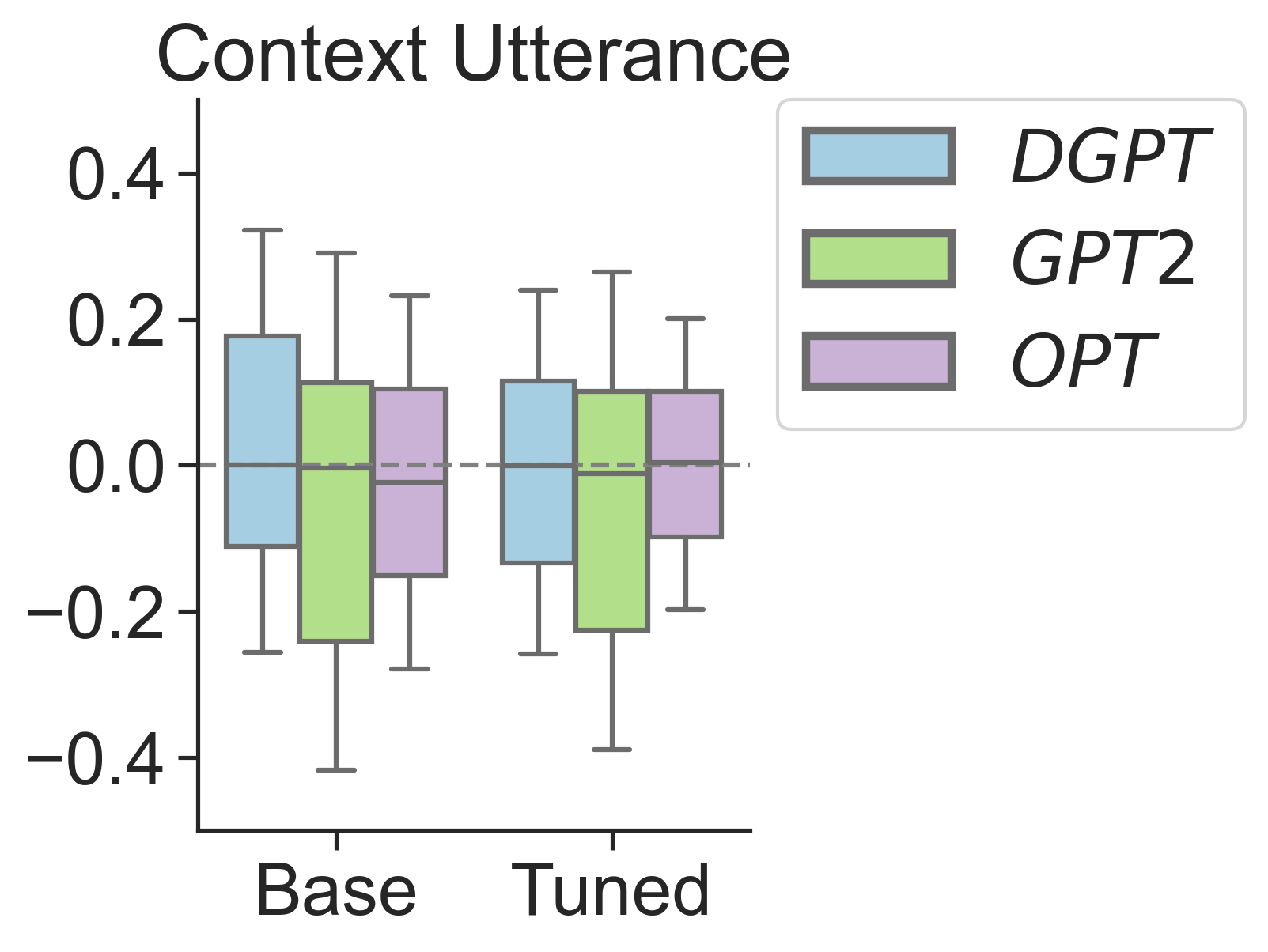}

    \caption{Attribution patterns for \textit{Speaker labels} and \textit{Utterances} in the dialouge Context (\textit{Ctx}) during model comprehension of human Target (\textit{Tgt}) utterances. The y-axis measures the \textit{relative boosting effect}. 
    }
    \label{fig:attribstats}
\end{figure}

We additionally analyse \textit{Target vs. Context vs. Speaker Label} salience patterns.
Regarding the \textit{speaker labels} in the context (i.e., sequences containing non-utterance tokens: \textit{A:}, \textit{<eos>}), the effect of special or structural tokens on the performance and behaviour of LLMs is an ongoing area of research~\citep{wolf2019transfertransfo,Gu2020speakerawarebert,wallbridge2023dialogue,ekstedt2020turngpt}, we expect model attribution behaviour to be more similar between tuned models.

From Figure~\ref{fig:attribstats}, we observe far higher variance in attribution over the target utterance than over the utterances in the context, with a similar relative difference between the speaker label in the target vs. those in the context. 
We observe very few consistent patterns across models in terms of relative boosting effects, except for \textit{speaker label Ctx}, which becomes more relatively uniform (and closer to 0) with tuning.
We observe that \gpt\ learns to attribute relatively higher salience over the text in the context utterances than to that in the target. In other words, they learn to place relatively more importance on the target utterance itself (\sw: $t=-8.01$, $p<0.05$; \mt: $t=-14.42$, $p<0.05$).

\section{Generation Quality}
\label{app:mauve}

To perform a comparable correlation analysis of MAUVE scores and possibly influencing factors, we treat each model generation (we generate five responses to each sample) as a separate corpus. This allows us to compute multiple MAUVE scores for each model (instead of just one score that is based on all the model generations). For best practices, MAUVE requires at least a few thousand examples to run (the original paper uses 5000). Since we have $2,395$ samples in \mt~and $8,705$ samples in \sw, we select the number of samples used for MAUVE score computation to be $3,000$. We make use of all the \mt~samples for computation, and randomly sample model generations when we have more than $3,000$ examples available. We obtain five MAUVE scores for each model (base and fine-tuned), resulting in $30$ scores for each corpus.

Table~\ref{tab:mauve} shows a full breakdown of the most consistent results across models. Since we are interested in general properties which apply to conversational corpora, we combine both \mt and \sw in this analysis. We find a strong $\rho$ correlation across models, weakest for \dgpt. 
\begin{table}
    \small
    \centering
    \begin{tabular}{llllr}
    \toprule
        Metric & Type & Model & $\rho$ & $p$\\ 
        \hline
        Construction Overlap & B & \dgpt & 0.914 & 0\\ 
        Construction Overlap & B & \gpt & 0.933 & 0\\ 
        Construction Overlap & B & \opt & 0.888 & 0.001\\ 
        Construction Overlap & T & \dgpt & 0.698 & 0.025\\ 
        Construction Overlap & T & \gpt & 0.808 & 0.005\\ 
        Construction Overlap & T & \opt & 0.976 & 0\\ 
        Prop. Repetition & B & \dgpt & 0.905 & 0\\ 
        Prop. Repetition & B & \gpt & 0.91 & 0\\ 
        Prop. Repetition & B & \opt & 0.944 & 0\\ 
        Prop. Repetition & T & \dgpt & 0.637 & 0.047\\ 
        Prop. Repetition & T & \gpt & 0.747 & 0.013\\ 
        Prop. Repetition & T & \opt & 0.98 & 0 \\
        \bottomrule
    \end{tabular}
    \caption{MAUVE $\rho$ correlation results. Metrics are the absolute value of the \textit{difference} between model and human levels of \co~and repetition, thus a positive correlation indicates an inverse correlation of the two metrics of human-likeness}
    \label{tab:mauve}
\end{table}

%% file: sections/appendix_models.tex
\section{Linear Mixed Effects Regression Results}
\label{app:models}
To evaluate \textit{local} effects, specifically the relationship between utterances in the context and the target utterance, we employ linear mixed-effect models, including \textit{dialogue and sample} identifiers as random effects.

\subsection{Production: Repetition Effects}
\label{app:repetition_model}

To measure repetition effects we fit separate models for construction overlap \co, and vocabulary overlap \vo, making these the dependent variables. We include dialogue and sample as random effects to allow for group-level variability in the linear model. 
We firstly investigate the effects of speaker, and distance. To measure repetition in the human data, we include speaker, and distance given speaker as fixed effects.
To measure repetition in models, we follow the same process as for the human data, but adding model type (base or tuned) and their interaction with distance as additional fixed effects. Results for \vo\ can be found in Table~\ref{tab:lmer_vo_model}, and \co\ in Table~\ref{tab:lmer_co_model}.

We then conduct a second analysis, this time to investigate the impact of different properties of constructions on the \co\ effects. We include speaker, distance, construction length, specificity (PMI) and frequency as independent fixed effects. Results can be found in Table~\ref{tab:lmer_props}.

\subsection{Comprehension: Attribution Effects}
\label{app:attribution_model}

To measure Attribution strengths over the context utterances during model comprehension of human-produced target utterances, we made attribution the dependent variable.

\subsection{Attribution Over Human Utterances}
To investigate the effect of local context repetition on model attribution strengths to context utterance text during target utterance comprehension, we include speaker, distance, construction overlap, vocabulary overlap, average construction PMI, and construction frequency as fixed effects. Results can be found in Table~\ref{tab:lmer_attrib}.

\subsection{Attribution Over Special Tokens}
To investigate the effect of distance on model attribution to speaker labels within the context during target utterance comprehension, we include distance, model type (base or tuned) and their interaction as fixed effects. Results can be found in Table~\ref{tab:lmer_attrib_label}.


\begin{table*}\centering \small \scalebox{0.8} { 
\setlength{\tabcolsep}{3.5pt}
\begin{tabular}{@{}l|rrrrrr|rrrrrr@{}} \toprule
& \multicolumn{6}{c}{\textit{Switchboard}} & \multicolumn{6}{c}{\textit{Map Task}} \\
& Coef. & Std. & $z$ & $P>|z|$ & $[0.025$ & $0.975]$ & Coef. & Std. & $z$ & $P>|z|$ & $[0.025$ & $0.975]$ \\
\midrule

\textit{Human} & & & & & & & & & & & & \\
\hspace{2mm}Intercept & 0.119 & 0.002 & 58.807 & 0.000 & 0.115 & 0.122 & 0.137 & 0.004& 33.787 &0.000 &0.129 &0.145 \\
\hspace{2mm}S[T.same] & 0.064 & 0.003 & 19.889 & 0.000 & 0.058 & 0.071 & 0.033 & 0.007 &5.013 &0.000 &0.020 &0.045 \\
\hspace{2mm}dist:S[diff] & -0.001& 0.000 & -1.868 & 0.062 & -0.001 & 0.000 & -0.005& 0.001& -6.592& 0.000& -0.006& -0.003 \\
\hspace{2mm}dist:S[same] & -0.005 & 0.001 & -10.705 & 0.000 & -0.006 & -0.004 &  -0.002 &0.001 &-1.488 &0.137 &-0.004& 0.000 \\

\midrule
\textit{\gpt} & & & & & & & & & & & & \\
\hspace{2mm}Intercept & 0.129 & 0.001 & 110.696 & 0.000 & 0.127 & 0.132 & 0.129 & 0.002 & 67.475 & 0.000 & 0.125 & 0.133 \\
\hspace{2mm}S[T.same] & 0.076 & 0.002 & 48.199 & 0.000 & 0.073 & 0.080 & 0.050 & 0.003 & 19.480 & 0.000 & 0.045 & 0.056 \\
\hspace{2mm}type[T.tuned] & -0.011 & 0.001 & -10.672 & 0.000 & -0.013 & -0.009 & -0.002 & 0.002 & -1.357 & 0.175 & -0.006 & 0.001 \\
\hspace{2mm}dist:S[diff]:type[base] & 0.000 & 0.000 & 2.142 & 0.032 & 0.000 & 0.001 & -0.003 & 0.000 & -9.877 & 0.000 & -0.003 & -0.002 \\
\hspace{2mm}dist:S[same]:type[base] & -0.008 & 0.000 & -36.207 & 0.000 & -0.009 & -0.008 & -0.008 & 0.000 & -20.167 & 0.000 & -0.008 & -0.007 \\
\hspace{2mm}dist:S[diff]:type[tuned] & 0.002 & 0.000 & 11.460 & 0.000 & 0.002 & 0.002 & -0.002 & 0.000 & -8.011 & 0.000 & -0.003 & -0.002 \\
\hspace{2mm}dist:S[same]:type[tuned] & -0.006 & 0.000 & -28.161 & 0.000 & -0.007 & -0.006 & -0.004 & 0.000 & -10.058 & 0.000 & -0.005 & -0.003 \\

\midrule
\textit{\opt} & & & & & & & &  & & & & \\
\hspace{2mm}Intercept & 0.147 & 0.001 & 147.422 & 0.000 & 0.145 & 0.149 & 0.158 & 0.002 & 69.367 & 0.000 & 0.153 & 0.162 \\
\hspace{2mm}S[T.same] & 0.034 & 0.001 & 25.623 & 0.000 & 0.032 & 0.037 & 0.034 & 0.003 & 11.096 & 0.000 & 0.028 & 0.040 \\
\hspace{2mm}type[T.tuned] & -0.015 & 0.001 & -16.526 & 0.000 & -0.017 & -0.013 & -0.010 & 0.002 & -5.213 & 0.000 & -0.014 & -0.007 \\
\hspace{2mm}dist:S[diff]:type[base] & -0.003 & 0.000 & -19.647 & 0.000 & -0.003 & -0.003 & -0.005 & 0.000 & -14.935 & 0.000 & -0.006 & -0.004 \\
\hspace{2mm}dist:S[same]:type[base] & -0.008 & 0.000 & -38.836 & 0.000 & -0.008 & -0.007 & -0.009 & 0.000 & -19.171 & 0.000 & -0.009 & -0.008 \\
\hspace{2mm}dist:S[diff]:type[tuned] & -0.001 & 0.000 & -5.039 & 0.000 & -0.001 & -0.000 & -0.002 & 0.000 & -7.227 & 0.000 & -0.003 & -0.002 \\
\hspace{2mm}dist:S[same]:type[tuned] & -0.003 & 0.000 & -12.382 & 0.000 & -0.003 & -0.002 & -0.001 & 0.000 & -2.042 & 0.041 & -0.002 & -0.000 \\

\midrule
\textit{\dgpt} & & & & & & & & & & & & \\
\hspace{2mm}Intercept & 0.104 & 0.001 & 69.536 & 0.000 & 0.101 & 0.107 & 0.142 & 0.002 & 65.090 & 0.000 & 0.138 & 0.146 \\
\hspace{2mm}S[T.same] & 0.047 & 0.002 & 27.535 & 0.000 & 0.043 & 0.050 & 0.027 &0.003  & 9.267 &  0.000  & 0.021  & 0.032 \\
\hspace{2mm}type[T.tuned] & 0.018 & 0.001 & 13.055 & 0.000 & 0.015 & 0.020 & 0.001 & 0.002  & 0.427 &  0.669  & -0.003  & 0.005 \\
\hspace{2mm}dist:S[diff]:type[base] & 0.001 & 0.000 & 3.648 & 0.000 & 0.000 & 0.001 & -0.004  & 0.000 & -11.628 &  0.000 &  -0.005 & -0.003 \\
\hspace{2mm}dist:S[same]:type[base] & -0.007&  0.000&  -23.073&  0.000 & -0.008&  -0.007 & -0.010  & 0.000  & -22.139 &  0.000 &  -0.011  & -0.009 \\
\hspace{2mm}dist:S[diff]:type[tuned] & 0.001&  0.000&  3.920&  0.000&  0.000&  0.001 & -0.004  & 0.000  & -11.219  & 0.000 &  -0.004  & -0.003 \\
\hspace{2mm}dist:S[same]:type[tuned] & -0.005&  0.000&  -22.278 & 0.000&  -0.006&  -0.005 & -0.004  & 0.000  & -9.171 &  0.000  & -0.005  & -0.003 \\

\bottomrule \end{tabular}}
\caption{Repetition effects for Vocabulary Overlap \vo. \textit{S} indicates speaker, \textit{type} indicates model type (base or fine-tuned), \textit{diff} indicates whether the two utterances come from different speakers, or between-speaker repetition, and \textit{same} indicates whether the two utterances come from the same speakers, or within-speaker repetition.}
\label{tab:lmer_vo_model}
\end{table*}

\begin{table*}\centering \small \scalebox{0.8} { 
\setlength{\tabcolsep}{3.5pt}
\begin{tabular}{@{}l|rrrrrr|rrrrrr@{}} \toprule
& \multicolumn{6}{c}{\textit{Switchboard}} & \multicolumn{6}{c}{\textit{Map Task}} \\
& Coef. & Std. & $z$ & $P>|z|$ & $[0.025$ & $0.975]$ & Coef. & Std. & $z$ & $P>|z|$ & $[0.025$ & $0.975]$ \\

\midrule
\textit{Human} & & & & & & & & & & & & \\
\hspace{2mm}Intercept & 0.009 & 0.000 & 31.878 & 0.000 & 0.009 & 0.010 & 0.047 & 0.002 & 29.468 & 0.000 & 0.043 & 0.050 \\
\hspace{2mm}S[T.same] & -0.007 & 0.000 & -14.930 & 0.000 & -0.008 & -0.006 & -0.033 & 0.003 & -12.807 & 0.000 & -0.038 & -0.028 \\
\hspace{2mm}dist:S[diff] & -0.001 & 0.000 & -15.367 & 0.000 & -0.001 & -0.001 & -0.005 & 0.000 & -15.659 & 0.000 & -0.005 & -0.004 \\
\hspace{2mm}dist:S[same] & -0.000 & 0.000 & -2.386 & 0.017 & -0.000 & -0.000 & -0.001 & 0.000 & -1.471 & 0.141 & -0.001 & 0.000 \\

\midrule
\textit{\gpt} & & & & & & & &  & & & & \\
\hspace{2mm}Intercept & 0.010 & 0.000 & 63.140 & 0.000 & 0.009 & 0.010 & 0.037 & 0.001 & 54.133 & 0.000 & 0.036 & 0.038 \\
\hspace{2mm}S[T.same] & -0.006 & 0.000 & -27.845 & 0.000 & -0.006 & -0.005 & -0.023 & 0.001 & -25.390 & 0.000 & -0.025 & -0.021 \\
\hspace{2mm}type[T.tuned] & -0.003 & 0.000 & -19.413 & 0.000 & -0.004 & -0.003 & -0.000 & 0.001 & -0.624 & 0.533 & -0.002 & 0.001 \\
\hspace{2mm}dist:S[diff]:type[base] & -0.001 & 0.000 & -19.494 & 0.000 & -0.001 & -0.000 & -0.003 & 0.000 & -21.228 & 0.000 & -0.003 & -0.002 \\
\hspace{2mm}dist:S[same]:type[base] & -0.000 & 0.000 & -12.555 & 0.000 & -0.001 & -0.000 & -0.001 & 0.000 & -5.939 & 0.000 & -0.001 & -0.001 \\
\hspace{2mm}dist:S[diff]:type[tuned] & -0.000 & 0.000 & -7.264 & 0.000 & -0.000 & -0.000 & -0.003 & 0.000 & -21.669 & 0.000 & -0.003 & -0.002 \\
\hspace{2mm}dist:S[same]:type[tuned] & 0.000 & 0.000 & 2.012 & 0.044 & 0.000 & 0.000 & -0.001 & 0.000 & -5.276 & 0.000 & -0.001 & -0.001 \\

\midrule
\textit{\opt} & & & & & & & &  & & & & \\
\hspace{2mm}Intercept & 0.016 & 0.000 & 103.178 & 0.000 & 0.015 & 0.016 & 0.043 & 0.001 & 58.941 & 0.000 & 0.042 & 0.045 \\
\hspace{2mm}S[T.same] & -0.011 & 0.000 & -52.886 & 0.000 & -0.011 & -0.010 & -0.024 & 0.001 & -24.048 & 0.000 & -0.025 & -0.022 \\
\hspace{2mm}type[T.tuned] & -0.006 & 0.000 & -32.546 & 0.000 & -0.006 & -0.005 & -0.010 & 0.001 & -13.559 & 0.000 & -0.012 & -0.009 \\
\hspace{2mm}dist:S[diff]:type[base] & -0.001 & 0.000 & -49.486 & 0.000 & -0.001 & -0.001 & -0.004 & 0.000 & -26.986 & 0.000 & -0.004 & -0.003 \\
\hspace{2mm}dist:S[same]:type[base] & -0.001 & 0.000 & -17.805 & 0.000 & -0.001 & -0.001 & -0.002 & 0.000 & -10.631 & 0.000 & -0.002 & -0.001 \\
\hspace{2mm}dist:S[diff]:type[tuned] & -0.001 & 0.000 & -25.315 & 0.000 & -0.001 & -0.001 & -0.002 & 0.000 & -16.731 & 0.000 & -0.002 & -0.002 \\
\hspace{2mm}dist:S[same]:type[tuned] & 0.000 & 0.000 & 8.118 & 0.000 & 0.000 & 0.000 & -0.000 & 0.000 & -0.706 & 0.480 & -0.000 & 0.000 \\

\midrule
\textit{\dgpt} & & & & & & & & & & & & \\
\hspace{2mm}Intercept & 0.004 & 0.000 & 21.791 & 0.000 & 0.003 & 0.004 & 0.022 & 0.001 & 33.796 & 0.000 & 0.020 & 0.023 \\
\hspace{2mm}S[T.same] & -0.004 & 0.000 & -24.266 & 0.000 & -0.004 & -0.004 & -0.019 & 0.001 & -23.392 & 0.000 & -0.021 & -0.018 \\
\hspace{2mm}type[T.tuned] & 0.003 & 0.000 & 16.913 & 0.000 & 0.003 & 0.003 & 0.013 & 0.001 & 19.424 & 0.000 & 0.012 & 0.014 \\
\hspace{2mm}dist:S[diff]:type[base] & -0.000 & 0.000 & -10.319 & 0.000 & -0.000 & -0.000 & -0.002 & 0.000 & -19.909 & 0.000 & -0.003 & -0.002 \\
\hspace{2mm}dist:S[same]:type[base] & 0.000 & 0.000 & 3.740 & 0.000 & 0.000 & 0.000 & 0.000 & 0.000 & 0.303 & 0.762 & -0.000 & 0.000 \\
\hspace{2mm}dist:S[diff]:type[tuned] & -0.000 & 0.000 & -10.197 & 0.000 & -0.000 & -0.000 & -0.002 & 0.000 & -17.875 & 0.000 & -0.002 & -0.002 \\
\hspace{2mm}dist:S[same]:type[tuned] & -0.000 & 0.000 & -8.171 & 0.000 & -0.000 & -0.000 & -0.001 & 0.000 & -9.446 & 0.000 & -0.002 & -0.001 \\

\bottomrule \end{tabular}}
\caption{Repetition effects for Construction Overlap \co. \textit{S} indicates speaker, \textit{type} indicates model type (base or fine-tuned), \textit{diff} indicates whether the two utterances come from different speakers, or between-speaker repetition, and \textit{same} indicates whether the two utterances come from the same speakers, or within-speaker repetition.}
\label{tab:lmer_co_model}
\end{table*}

\begin{table*}\centering \small \scalebox{0.8} { 
\setlength{\tabcolsep}{3.5pt}
\begin{tabular}{@{}l|rrrrrr|rrrrrr@{}} \toprule
& \multicolumn{6}{c}{\textit{Switchboard}} & \multicolumn{6}{c}{\textit{Map Task}} \\
& Coef. & Std. & $z$ & $P>|z|$ & $[0.025$ & $0.975]$ & Coef. & Std. & $z$ & $P>|z|$ & $[0.025$ & $0.975]$ \\

\midrule
\textit{Human} & & & & & & & &  & & & & \\
\hspace{2mm}Intercept & 0.074 & 0.021 & 3.505 & 0.000 & 0.033 & 0.116 & 0.099 & 0.028 & 3.554 & 0.000 & 0.045 & 0.154 \\
\hspace{2mm}S[T.same] & -0.006 & 0.011 & -0.533 & 0.594 & -0.029 & 0.016 & -0.031 & 0.015 & -2.061 & 0.039 & -0.060 & -0.002 \\
\hspace{2mm}dist & -0.003 & 0.001 & -4.506 & 0.000 & -0.005 & -0.002 & -0.004 & 0.001 & -3.330 & 0.001 & -0.006 & -0.001 \\
\hspace{2mm}avg\_constr\_len & 0.057 & 0.006 & 10.155 & 0.000 & 0.046 & 0.068 & 0.133 & 0.007 & 18.607 & 0.000 & 0.119 & 0.146 \\
\hspace{2mm}pmi\_avg & 0.001 & 0.001 & 0.865 & 0.387 & -0.001 & 0.003 & 0.003 & 0.002 & 1.427 & 0.154 & -0.001 & 0.008 \\
\hspace{2mm}freq\_constr & -0.014 & 0.004 & -3.392 & 0.001 & -0.023 & -0.006 & -0.035 & 0.005 & -7.074 & 0.000 & -0.045 & -0.025 \\

\midrule
\textit{BASE} & & & & & & & & & & & & \\
\midrule

\textit{\gpt} & & & & & & & & & & & & \\
\hspace{2mm}Intercept & 0.048 & 0.010 & 4.629 & 0.000 & 0.028 & 0.068 & 0.109 & 0.014 & 7.533 & 0.000 & 0.081 & 0.137 \\
\hspace{2mm}S[T.same] & -0.026 & 0.006 & -4.395 & 0.000 & -0.037 & -0.014 & -0.017 & 0.008 & -2.138 & 0.032 & -0.033 & -0.001 \\
\hspace{2mm}dist & -0.004 & 0.001 & -8.614 & 0.000 & -0.006 & -0.003 & -0.005 & 0.001 & -5.689 & 0.000 & -0.006 & -0.003 \\
\hspace{2mm}avg\_constr\_len & 0.058 & 0.003 & 19.832 & 0.000 & 0.052 & 0.064 & 0.127 & 0.004 & 29.966 & 0.000 & 0.119 & 0.135 \\
\hspace{2mm}pmi\_avg & 0.002 & 0.000 & 3.454 & 0.001 & 0.001 & 0.002 & 0.004 & 0.001 & 3.865 & 0.000 & 0.002 & 0.006 \\
\hspace{2mm}freq\_constr & 0.005 & 0.002 & 2.150 & 0.032 & 0.000 & 0.009 & -0.016 & 0.003 & -6.018 & 0.000 & -0.022 & -0.011 \\

\midrule
\textit{\opt} & & & & & & & & & & & & \\
\hspace{2mm}Intercept & 0.022 & 0.007 & 3.110 & 0.002 & 0.008 & 0.036 & 0.088 & 0.016 & 5.516 & 0.000 & 0.057 & 0.119 \\
\hspace{2mm}S[T.same] & -0.025 & 0.005 & -5.151 & 0.000 & -0.034 & -0.015 & -0.030 & 0.010 & -3.134 & 0.002 & -0.049 & -0.011 \\
\hspace{2mm}dist & -0.004 & 0.000 & -9.875 & 0.000 & -0.004 & -0.003 & -0.007 & 0.001 & -8.165 & 0.000 & -0.008 & -0.005 \\
\hspace{2mm}avg\_constr\_len & 0.077 & 0.002 & 41.700 & 0.000 & 0.073 & 0.081 & 0.134 & 0.004 & 37.148 & 0.000 & 0.127 & 0.141 \\
\hspace{2mm}pmi\_avg & 0.001 & 0.000 & 3.862 & 0.000 & 0.001 & 0.002 & 0.004 & 0.001 & 3.105 & 0.002 & 0.001 & 0.006 \\
\hspace{2mm}freq\_constr & -0.000 & 0.002 & -0.232 & 0.816 & -0.004 & 0.003 & -0.003 & 0.003 & -1.162 & 0.245 & -0.009 & 0.002 \\

\midrule
\textit{\dgpt} & & & & & & & &  & & & & \\
\hspace{2mm}Intercept & 0.314 & 0.084 & 3.759 & 0.000 & 0.150 & 0.478 & 0.162 & 0.035 & 4.594 & 0.000 & 0.093 & 0.231 \\
\hspace{2mm}S[T.same] & -0.041 & 0.039 & -1.059 & 0.290 & -0.117 & 0.035 & -0.011 & 0.017 & -0.623 & 0.533 & -0.044 & 0.023 \\
\hspace{2mm}dist & -0.010 & 0.004 & -2.844 & 0.004 & -0.017 & -0.003 & -0.006 & 0.002 & -3.210 & 0.001 & -0.010 & -0.002 \\
\hspace{2mm}avg\_constr\_len & 0.083 & 0.027 & 3.099 & 0.002 & 0.030 & 0.135 & 0.115 & 0.009 & 12.720 & 0.000 & 0.097 & 0.132 \\
\hspace{2mm}pmi\_avg & 0.000 & 0.007 & 0.059 & 0.953 & -0.013 & 0.014 & 0.008 & 0.003 & 2.914 & 0.004 & 0.003 & 0.014 \\
\hspace{2mm}freq\_constr & -0.019 & 0.009 & -2.059 & 0.039 & -0.037 & -0.001 & -0.002 & 0.007 & -0.237 & 0.812 & -0.015 & 0.012 \\

\midrule
\textit{TUNED} & & & & & & & & & & & & \\
\midrule

\textit{\gpt} & & & & & & & & & & & & \\
\hspace{2mm}Intercept & 0.202 & 0.020 & 10.227 & 0.000 & 0.163 & 0.241 & 0.059 & 0.014 & 4.282 & 0.000 & 0.032 & 0.087 \\
\hspace{2mm}S[T.same] & -0.030 & 0.010 & -2.920 & 0.004 & -0.051 & -0.010 & -0.031 & 0.007 & -4.447 & 0.000 & -0.044 & -0.017 \\
\hspace{2mm}dist & -0.005 & 0.001 & -5.801 & 0.000 & -0.007 & -0.004 & -0.006 & 0.001 & -7.508 & 0.000 & -0.007 & -0.004 \\
\hspace{2mm}avg\_constr\_len & 0.067 & 0.006 & 11.523 & 0.000 & 0.055 & 0.078 & 0.128 & 0.004 & 28.787 & 0.000 & 0.119 & 0.137 \\
\hspace{2mm}pmi\_avg & -0.010 & 0.001 & -11.189 & 0.000 & -0.012 & -0.008 & 0.004 & 0.001 & 4.017 & 0.000 & 0.002 & 0.005 \\
\hspace{2mm}freq\_constr & 0.004 & 0.004 & 1.032 & 0.302 & -0.004 & 0.013 & -0.011 & 0.003 & -4.175 & 0.000 & -0.016 & -0.006 \\

\midrule
\textit{\opt} & & & & & & & & & & & & \\
\hspace{2mm}Intercept & 0.056 & 0.010 & 5.793 & 0.000 & 0.037 & 0.075 & 0.192 & 0.018 & 10.965 & 0.000 & 0.158 & 0.227 \\
\hspace{2mm}S[T.same] & -0.025 & 0.006 & -4.117 & 0.000 & -0.038 & -0.013 & -0.057 & 0.010 & -5.581 & 0.000 & -0.077 & -0.037 \\
\hspace{2mm}dist & -0.003 & 0.000 & -6.406 & 0.000 & -0.004 & -0.002 & -0.006 & 0.001 & -6.700 & 0.000 & -0.008 & -0.004 \\
\hspace{2mm}avg\_constr\_len & 0.064 & 0.003 & 24.984 & 0.000 & 0.059 & 0.069 & 0.123 & 0.004 & 28.582 & 0.000 & 0.114 & 0.131 \\
\hspace{2mm}pmi\_avg & 0.001 & 0.000 & 3.123 & 0.002 & 0.001 & 0.002 & -0.001 & 0.001 & -1.085 & 0.278 & -0.004 & 0.001 \\
\hspace{2mm}freq\_constr & -0.004 & 0.002 & -2.011 & 0.044 & -0.009 & -0.000 & -0.022 & 0.003 & -6.438 & 0.000 & -0.029 & -0.016 \\

\midrule
\textit{\dgpt} & & & & & & & & & & & & \\
\hspace{2mm}Intercept & 0.023 & 0.009 & 2.429 & 0.015 & 0.004 & 0.041 & 0.124 & 0.015 & 8.252 & 0.000 & 0.094 & 0.153 \\
\hspace{2mm}S[T.same] & -0.015 & 0.005 & -3.130 & 0.002 & -0.024 & -0.006 & -0.026 & 0.007 & -3.524 & 0.000 & -0.040 & -0.011 \\
\hspace{2mm}dist & -0.005 & 0.000 & -10.320 & 0.000 & -0.006 & -0.004 & -0.005 & 0.001 & -5.817 & 0.000 & -0.006 & -0.003 \\
\hspace{2mm}avg\_constr\_len & 0.054 & 0.003 & 18.517 & 0.000 & 0.048 & 0.059 & 0.110 & 0.005 & 22.849 & 0.000 & 0.100 & 0.119 \\
\hspace{2mm}pmi\_avg & 0.001 & 0.000 & 2.872 & 0.004 & 0.000 & 0.002 & -0.002 & 0.001 & -2.332 & 0.020 & -0.004 & -0.000 \\
\hspace{2mm}freq\_constr & 0.003 & 0.002 & 1.717 & 0.086 & -0.000 & 0.007 & -0.013 & 0.003 & -4.412 & 0.000 & -0.019 & -0.007 \\

\bottomrule \end{tabular}}
\caption{
Repetition details for \co\ taking into account length, specificity (PMI) and construction frequency (freq). \textit{S} indicates speaker, \textit{type} indicates model type (base or fine-tuned), \textit{diff} indicates whether the two utterances come from different speakers, or between-speaker repetition, and \textit{same} indicates whether the two utterances come from the same speakers, or within-speaker repetition.
} 
\label{tab:lmer_props}
\end{table*}

\begin{table*}\centering \small \scalebox{0.7} { 
\setlength{\tabcolsep}{3.5pt}
\begin{tabular}{@{}l|rrrrrr|rrrrrr@{}} \toprule
& \multicolumn{6}{c}{\textit{Switchboard}} & \multicolumn{6}{c}{\textit{Map Task}} \\
& Coef. & Std. & $z$ & $P>|z|$ & $[0.025$ & $0.975]$ & Coef. & Std. & $z$ & $P>|z|$ & $[0.025$ & $0.975]$ \\

\midrule
\textit{BASE} & & & & & & & & & & & & \\
\midrule

\textit{\gpt} & & & & & & & & & & & & \\
\hspace{2mm}Intercept & 0.399 & 0.010 & 39.506 & 0.000 & 0.380 & 0.419 & 0.457 & 0.016 & 28.858 & 0.000 & 0.426 & 0.488 \\
\hspace{2mm}S[T.same] & 0.003 & 0.006 & 0.493 & 0.622 & -0.009 & 0.014 & -0.015 & 0.008 & -1.752 & 0.080 & -0.031 & 0.002 \\
\hspace{2mm}dist\_from\_prev\_turn & 0.002 & 0.001 & 3.559 & 0.000 & 0.001 & 0.003 & -0.000 & 0.001 & -0.199 & 0.842 & -0.002 & 0.002 \\
\hspace{2mm}constr\_overlap & 0.323 & 0.015 & 22.127 & 0.000 & 0.294 & 0.351 & 0.190 & 0.024 & 7.797 & 0.000 & 0.142 & 0.237 \\
\hspace{2mm}vocab\_overlap & -0.383 & 0.013 & -30.143 & 0.000 & -0.408 & -0.358 & -0.198 & 0.023 & -8.626 & 0.000 & -0.243 & -0.153 \\
\hspace{2mm}pmi\_avg & 0.003 & 0.001 & 5.488 & 0.000 & 0.002 & 0.004 & -0.001 & 0.001 & -1.038 & 0.299 & -0.004 & 0.001 \\
\hspace{2mm}freq\_constr & 0.008 & 0.002 & 3.090 & 0.002 & 0.003 & 0.012 & 0.002 & 0.003 & 0.725 & 0.469 & -0.004 & 0.008 \\

\midrule
\textit{\opt} & & & & & & & & & & & & \\
\hspace{2mm}Intercept & 0.534 & 0.012 & 46.370 & 0.000 & 0.511 & 0.556 & 0.516 & 0.018 & 29.281 & 0.000 & 0.481 & 0.551 \\
\hspace{2mm}S[T.same] & -0.002 & 0.007 & -0.338 & 0.736 & -0.016 & 0.011 & 0.039 & 0.008 & 4.822 & 0.000 & 0.023 & 0.055 \\
\hspace{2mm}dist\_from\_prev\_turn & -0.014 & 0.001 & -22.485 & 0.000 & -0.016 & -0.013 & -0.008 & 0.001 & -7.799 & 0.000 & -0.010 & -0.006 \\
\hspace{2mm}constr\_overlap & 0.039 & 0.017 & 2.258 & 0.024 & 0.005 & 0.072 & 0.035 & 0.021 & 1.716 & 0.086 & -0.005 & 0.076 \\
\hspace{2mm}vocab\_overlap & -0.041 & 0.014 & -2.928 & 0.003 & -0.068 & -0.013 & -0.034 & 0.020 & -1.704 & 0.088 & -0.073 & 0.005 \\
\hspace{2mm}pmi\_avg & 0.000 & 0.001 & 0.065 & 0.949 & -0.001 & 0.001 & -0.000 & 0.001 & -0.217 & 0.828 & -0.003 & 0.002 \\
\hspace{2mm}freq\_constr & 0.001 & 0.003 & 0.341 & 0.733 & -0.005 & 0.006 & -0.000 & 0.003 & -0.119 & 0.905 & -0.007 & 0.006 \\

\midrule
\textit{\dgpt} & & & & & & & &  & & & & \\
\hspace{2mm}Intercept & 0.524 & 0.071 & 7.365 & 0.000 & 0.384 & 0.663 & 0.482 & 0.041 & 11.645 & 0.000 & 0.401 & 0.563 \\
\hspace{2mm}S[T.same] & -0.024 & 0.036 & -0.647 & 0.518 & -0.095 & 0.048 & 0.061 & 0.020 & 3.071 & 0.002 & 0.022 & 0.100 \\
\hspace{2mm}dist\_from\_prev\_turn & 0.012 & 0.004 & 2.871 & 0.004 & 0.004 & 0.020 & 0.007 & 0.003 & 2.704 & 0.007 & 0.002 & 0.012 \\
\hspace{2mm}constr\_overlap & 0.018 & 0.083 & 0.215 & 0.829 & -0.145 & 0.181 & -0.086 & 0.052 & -1.656 & 0.098 & -0.187 & 0.016 \\
\hspace{2mm}vocab\_overlap & -0.023 & 0.085 & -0.275 & 0.784 & -0.191 & 0.144 & 0.095 & 0.047 & 2.018 & 0.044 & 0.003 & 0.188 \\
\hspace{2mm}pmi\_avg & 0.001 & 0.007 & 0.174 & 0.861 & -0.013 & 0.016 & 0.007 & 0.003 & 2.116 & 0.034 & 0.001 & 0.014 \\
\hspace{2mm}freq\_constr & -0.011 & 0.009 & -1.218 & 0.223 & -0.028 & 0.007 & -0.017 & 0.008 & -2.032 & 0.042 & -0.033 & -0.001 \\

\midrule
\textit{TUNED} & & & & & & & & & & & & \\
\midrule

\textit{\gpt} & & & & & & & & & & & & \\
\hspace{2mm}Intercept & 0.463 & 0.017 & 26.730 & 0.000 & 0.429 & 0.497 & 0.436 & 0.015 & 29.226 & 0.000 & 0.406 & 0.465 \\
\hspace{2mm}S[T.same] & -0.033 & 0.009 & -3.510 & 0.000 & -0.051 & -0.014 & -0.013 & 0.008 & -1.590 & 0.112 & -0.030 & 0.003 \\
\hspace{2mm}dist\_from\_prev\_turn & -0.009 & 0.001 & -9.436 & 0.000 & -0.011 & -0.007 & 0.001 & 0.001 & 1.416 & 0.157 & -0.001 & 0.003 \\
\hspace{2mm}constr\_overlap & 0.277 & 0.020 & 14.149 & 0.000 & 0.239 & 0.315 & 0.183 & 0.024 & 7.511 & 0.000 & 0.135 & 0.230 \\
\hspace{2mm}vocab\_overlap & -0.308 & 0.019 & -15.922 & 0.000 & -0.346 & -0.270 & -0.202 & 0.022 & -9.113 & 0.000 & -0.245 & -0.159 \\
\hspace{2mm}pmi\_avg & 0.001 & 0.001 & 1.018 & 0.309 & -0.001 & 0.003 & -0.001 & 0.001 & -0.753 & 0.451 & -0.003 & 0.001 \\
\hspace{2mm}freq\_constr & 0.007 & 0.004 & 1.729 & 0.084 & -0.001 & 0.015 & 0.006 & 0.003 & 1.963 & 0.050 & 0.000 & 0.013 \\

\midrule
\textit{\opt} & & & & & & & & & & & & \\
\hspace{2mm}Intercept & 0.528 & 0.013 & 39.783 & 0.000 & 0.502 & 0.554 & 0.494 & 0.017 & 29.608 & 0.000 & 0.461 & 0.526 \\
\hspace{2mm}S[T.same] & -0.004 & 0.008 & -0.499 & 0.618 & -0.020 & 0.012 & 0.002 & 0.009 & 0.234 & 0.815 & -0.015 & 0.019 \\
\hspace{2mm}dist\_from\_prev\_turn & -0.004 & 0.001 & -5.376 & 0.000 & -0.005 & -0.002 & 0.001 & 0.001 & 1.536 & 0.124 & -0.000 & 0.003 \\
\hspace{2mm}constr\_overlap & 0.021 & 0.019 & 1.129 & 0.259 & -0.016 & 0.058 & -0.022 & 0.021 & -1.026 & 0.305 & -0.063 & 0.020 \\
\hspace{2mm}vocab\_overlap & -0.039 & 0.016 & -2.508 & 0.012 & -0.070 & -0.009 & 0.012 & 0.021 & 0.575 & 0.566 & -0.029 & 0.052 \\
\hspace{2mm}pmi\_avg & -0.001 & 0.001 & -1.377 & 0.168 & -0.002 & 0.000 & -0.001 & 0.001 & -0.568 & 0.570 & -0.003 & 0.002 \\
\hspace{2mm}freq\_constr & 0.001 & 0.003 & 0.195 & 0.845 & -0.006 & 0.007 & 0.004 & 0.003 & 1.108 & 0.268 & -0.003 & 0.011 \\

\midrule
\textit{\dgpt} & & & & & & & & & & & & \\
\hspace{2mm}Intercept & 0.472 & 0.013 & 35.438 & 0.000 & 0.446 & 0.498 & 0.445 & 0.017 & 25.447 & 0.000 & 0.411 & 0.479 \\
\hspace{2mm}S[T.same] & 0.003 & 0.008 & 0.401 & 0.689 & -0.012 & 0.019 & -0.006 & 0.010 & -0.637 & 0.524 & -0.026 & 0.013 \\
\hspace{2mm}dist\_from\_prev\_turn & 0.001 & 0.001 & 1.285 & 0.199 & -0.001 & 0.003 & 0.005 & 0.001 & 4.126 & 0.000 & 0.002 & 0.007 \\
\hspace{2mm}constr\_overlap & 0.022 & 0.021 & 1.039 & 0.299 & -0.019 & 0.063 & 0.064 & 0.028 & 2.305 & 0.021 & 0.010 & 0.118 \\
\hspace{2mm}vocab\_overlap & -0.046 & 0.017 & -2.748 & 0.006 & -0.079 & -0.013 & -0.055 & 0.025 & -2.225 & 0.026 & -0.104 & -0.007 \\
\hspace{2mm}pmi\_avg & 0.001 & 0.001 & 1.169 & 0.242 & -0.001 & 0.002 & -0.002 & 0.001 & -1.264 & 0.206 & -0.004 & 0.001 \\
\hspace{2mm}freq\_constr & 0.001 & 0.003 & 0.360 & 0.719 & -0.005 & 0.008 & 0.011 & 0.004 & 2.716 & 0.007 & 0.003 & 0.019 \\

\bottomrule \end{tabular}}
\caption{Attribution effects over human utterances. \textit{S} indicates speaker, \textit{type} indicates model type (base or fine-tuned), \textit{diff} indicates whether the two utterances come from different speakers, or between-speaker repetition, and \textit{same} indicates whether the two utterances come from the same speakers, or within-speaker repetition. constr\_overlap indicates \co, vocab\_overlap indicates \vo, PMI indicates specificity, and freq, frequency of shared constructions.
} 
\label{tab:lmer_attrib}
\end{table*}

\begin{table*}\centering \small \scalebox{0.7} { 
\setlength{\tabcolsep}{3.5pt}
\begin{tabular}{@{}l|rrrrrr|rrrrrr@{}} \toprule
& \multicolumn{6}{c}{\textit{Switchboard}} & \multicolumn{6}{c}{\textit{Map Task}} \\
& Coef. & Std. & $z$ & $P>|z|$ & $[0.025$ & $0.975]$ & Coef. & Std. & $z$ & $P>|z|$ & $[0.025$ & $0.975]$ \\
\midrule
\textit{\gpt} & & & & & & & & & & & & \\
\hspace{2mm}Intercept & 0.552 & 0.000 & 2122.312 & 0.000 & 0.551 & 0.552 & 0.554 & 0.001 & 878.909 & 0.000 & 0.552 & 0.555 \\
\hspace{2mm}m\_type[T.tuned] & -0.009 & 0.000 & -42.336 & 0.000 & -0.009 & -0.008 & -0.029 & 0.001 & -53.563 & 0.000 & -0.030 & -0.028 \\
\hspace{2mm}dist & 0.000 & 0.000 & 16.487 & 0.000 & 0.000 & 0.001 & -0.004 & 0.000 & -48.544 & 0.000 & -0.004 & -0.004 \\
\hspace{2mm}dist:m\_type[T.tuned] & -0.001 & 0.000 & -13.645 & 0.000 & -0.001 & -0.000 & 0.004 & 0.000 & 37.490 & 0.000 & 0.004 & 0.004 \\
\midrule
\textit{\opt} & & & & & & & & & & & & \\
\hspace{2mm}Intercept & 0.502 & 0.000 & 1599.293 & 0.000 & 0.502 & 0.503 & 0.519 & 0.001 & 730.825 & 0.000 & 0.518 & 0.520 \\
\hspace{2mm}m\_type[T.tuned] & -0.003 & 0.000 & -11.565 & 0.000 & -0.003 & -0.002 & -0.020 & 0.001 & -26.957 & 0.000 & -0.021 & -0.018 \\
\hspace{2mm}dist & -0.001 & 0.000 & -37.286 & 0.000 & -0.001 & -0.001 & -0.003 & 0.000 & -31.255 & 0.000 & -0.004 & -0.003 \\
\hspace{2mm}dist:m\_type[T.tuned] & 0.001 & 0.000 & 26.777 & 0.000 & 0.001 & 0.002 & 0.004 & 0.000 & 26.279 & 0.000 & 0.004 & 0.004 \\
\midrule
\textit{\dgpt} & & & & & & & &  & & & & \\
\hspace{2mm}Intercept & 0.488 & 0.000 & 1079.600 & 0.000 & 0.488 & 0.489 & 0.501 & 0.001 & 550.576 & 0.000 & 0.499 & 0.503 \\
\hspace{2mm}m\_type[T.tuned] & 0.017 & 0.000 & 42.653 & 0.000 & 0.017 & 0.018 & -0.003 & 0.001 & -2.734 & 0.006 & -0.005 & -0.001 \\
\hspace{2mm}dist & -0.003 & 0.000 & -37.818 & 0.000 & -0.003 & -0.002 & -0.004 & 0.000 & -29.147 & 0.000 & -0.004 & -0.004 \\
\hspace{2mm}dist:m\_type[T.tuned] & 0.002 & 0.000 & 22.719 & 0.000 & 0.002 & 0.002 & 0.005 & 0.000 & 25.426 & 0.000 & 0.005 & 0.005 \\
\bottomrule \end{tabular}}
\caption{Attribution effects over speaker labels. \textit{m\_type} indicates model: either base or tuned. \textit{dist} indicates distance between context and target utterances.} 
\label{tab:lmer_attrib_label}
\end{table*}